\documentclass[11pt]{article}
\usepackage{multirow}
\usepackage{array}
\usepackage{extarrows}
\usepackage{graphicx}
\usepackage[section]{placeins}
\usepackage{relsize}
\usepackage{comment}
\usepackage[titletoc,title]{appendix}
\usepackage{amsthm}
\usepackage{enumerate}
\usepackage[ruled,vlined,linesnumbered,noresetcount]{algorithm2e}
\usepackage[ruled]{algorithm2e} 
\usepackage{natbib} 
\usepackage{endnotes}
\let\footnote=\endnote

\usepackage{tikz}
\usetikzlibrary{shapes,fit,positioning}
\usepackage{accents}
\newcommand{\ub}[1]{\underaccent{\bar}{#1}}
\usepackage{graphicx}
\usepackage{multirow}
\usepackage{hhline}

\usepackage{color}
\definecolor{strcolor}{rgb}{0.6, 0.2, 0.6}
\definecolor{commentcolor}{rgb}{0.3125, 0.5, 0.3125}
\definecolor{keycol}{rgb}{0, 0, 1}

\usepackage{subcaption}

\usepackage{comment}

\usepackage{enumerate}

\usepackage{placeins}

\usepackage{bm}

\DeclareMathOperator*{\argmax}{arg\,max}
\DeclareMathOperator*{\argmin}{arg\,min}

\newcommand\maxmin{\text{maxmin}}
\newcommand\opt{\text{opt}}

\def\Halmos{\mbox{\quad$\square$}}

\usepackage{appendix}
\usepackage{amssymb,amsmath}
\usepackage{latexsym}

\numberwithin{equation}{section}

\newcommand{\be}{\begin{equation}}
\newcommand{\ee}{\end{equation}}
\newcommand{\beaa}{\begin{eqnarray*}}
\newcommand{\eeaa}{\end{eqnarray*}}
\newcommand{\bea}{\begin{eqnarray}}
\newcommand{\eea}{\end{eqnarray}}

\newcommand{\bei}{\begin{itemize}}
\newcommand{\eei}{\end{itemize}}

\newtheorem{theorem}{ \noindent T{\footnotesize HEOREM}}
\newtheorem{proposition}{ \noindent P{\footnotesize ROPOSITION}}[section]
\newtheorem{lemma}{ \noindent L{\footnotesize EMMA}}[section]
\newtheorem{corollary}{ \noindent C{\footnotesize OROLLARY}}

\newtheorem{assumption}{Assumption}

\begin{document}

	\title{Active Learning for Fair and Stable Online Allocations}
\author{Riddhiman Bhattacharya, Thanh Nguyen, Will Wei Sun, Mohit Tawarmalani
	\\
	Purdue University}

\date{}
\maketitle

\footnotetext[1]{Daniel School of Business, Purdue University, USA, bhatta76@purdue.edu. 
.}

\begin{abstract}
\noindent We explore an active learning approach for dynamic fair resource allocation problems. Unlike previous work that assumes full feedback from all agents on their allocations, we consider feedback from a select subset of agents at each epoch of the online resource allocation process. Despite this restriction, our proposed algorithms provide regret bounds that are sub-linear in number of time-periods for various measures that include fairness metrics commonly used in resource allocation problems and  stability considerations in matching mechanisms. The key insight of our algorithms lies in adaptively identifying the most informative feedback using \emph{dueling upper and lower confidence bounds}. With this strategy, we show that efficient decision-making does not require extensive feedback and produces efficient outcomes for a variety of problem classes.

\end{abstract}
\section{Introduction}
Ensuring fair and stable allocation of scarce resources is a fundamental challenge in a wide range of applications. Traditional literature assumes that information regarding agents' preferences, whether available centrally to the designer or held privately by the agents, is known before the allocation process (the mechanism). However, this assumption hinders application in practical settings where agents typically evaluate resources only after receiving or consuming them. Furthermore, such preference information is often noisy and expensive for the central designer to gather from all agents, thus complicating the implementation of traditional mechanisms.

Examples of domains where these challenges manifest include applications where geographical and  time constraints impede information collection, such as distributing resources to food banks and providing humanitarian aid to disaster areas and war zones \citep{aleksandrov2015online, aleksandrov2020online}. 
Even in online marketplaces devoid of physical constraints, such as dating services and job matching, evaluating information and collecting data presents a formidable challenge. Participants in these systems often assess compatibility only after the job commences or partnership begins, revealing the limitations of relying on pre-established preferences. Additionally, platforms themselves must exert significant effort to gather feedback through surveys and other mechanisms.

Recent literature bridges this gap partially by learning noisy preferences as allocation decisions are made. This approach makes allocation processes more adaptable and efficient when the information is incomplete or dynamically changing. However, the current research typically assumes that input from \emph{all} participants is available at each time-epoch of the allocation process~\citep{bistritz2020my, yamada2023learning, leshem2024fair, liu2020competing, cen2022regret}. Since gathering information is costly and often practical considerations make it infeasible, assuming its availability overlooks the possibility of designing efficient algorithms that operate with limited feedback and the accompanying analysis fails to illuminate which feedback is crucial for efficient design.

\begin{figure}[h!]
	\centering
	\includegraphics[width=10cm]{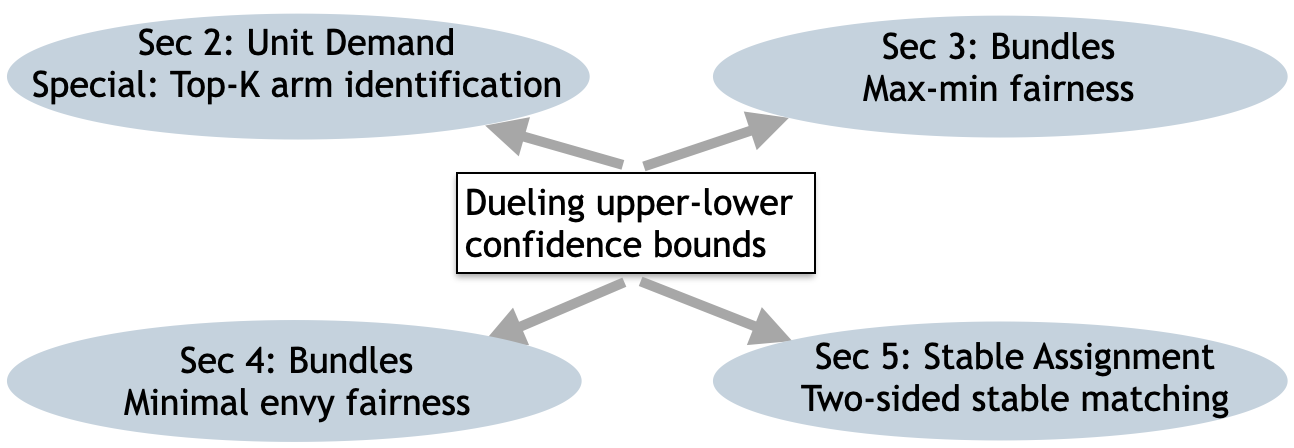}
	\caption{Outline of the four interconnected problems addressed in our paper.}
	\label{fig:outline}
\end{figure}

Our paper contributes on three fronts. First, we introduce a deliberate constraint on feedback, restricting it to a single agent or a limited number of agents per period instead of allowing input from all agents. Second, our paper makes a methodological contribution by developing a versatile framework that applies to both max-min/min-max envy scenarios and stable matching problems. The versatility of our approach underscores its adaptive and comprehensive nature, demonstrating that it is effective across diverse problem domains. Third, a key theoretical contribution of the paper is that, despite restricted feedback, our algorithms do not sacrifice regret significantly while addressing fairness and stability concerns. Our approach hinges on an active-learning procedure that carefully selects the agent from whom to gather feedback, ensuring its effectiveness in the allocation process. In  the following, we describe a series of problems, with increasing degrees of complexity, and briefly describe our solutions. Figure \ref{fig:outline} provides an outline of our paper. 
The simplest version we treat involves unit-demand agents seeking to consume single, indivisible items from a diverse set of resources. Minimax fairness focuses on the lowest reward any agent receives in the allocation, and our algorithm aims to maximize this reward. The online variant assumes that the reward matrix is unknown, which must be learnt during the allocation process. We additionally impose that feedback should be collected from one agent at each time-period. As specified, the problem adds to the growing literature on multi-armed bandit problems. A special case of interest is where rewards are agent-independent, and, in this setting, the problem reduces to the classical problem of finding the top K arms. Even for this special case, which is well-studied \citep{NIPS2015_ab233b68,locatelli2016optimal, JMLR:v18:16-206, pmlr-v151-garcelon22b,zhang2021quantile}, limited feedback is new since earlier studies uniformly assume that feedback from all $K$ arms is available at each time.  We introduce Algorithm~\ref{algo:ucblcb} based on the innovative concept of dueling upper-lower confidence bounds (dueling-ULCB). In this approach, we select the allocation based on the upper  confidence bound (UCB) but choose the feedback based on the lower  confidence bound (LCB). Out of various natural ideas for selecting feedback, this method proves to be the most effective, resulting in an algorithm with a sub-linear regret. Furthermore, the intuition behind it facilitates extensions to more complex problems.

We extend the original problem setting in three directions. First, Section \ref{sec:CMAB} explores a bundle setting where each agent is allocated a set of items rather than just one item. This corresponds to extending the classical MAB setting to the combinatorial multi-armed-bandit (CMAB) setting~\citep{cesa2012combinatorial, chen2013combinatorial} where the agents at each epoch pull a set of arms (super arms) instead of one arm. The technical challenge this setting poses is that the decision and sample space may be exponentially sized in the number of arms. To avoid regret depending on this large sample space, our method utilizes LCB and UCB bounds for each individual good rather than treating super-arms individually. Consistent with limited feedback, throughout the paper, we are interested in the amount of feedback collected, and solve decision problems in each time-period using oracles which may have exponential complexity depending on the problem complexity.

The second extension relates to the objective. Instead of pursuing the max-min objective, Section \ref{sec:Envy} aims to find an assignment that minimizes the maximum envy between any pair of agents. The envy of agent $i$ towards agent $i'$ is defined as the gain in reward if agent $i$ were to receive the bundle allocated to agent $i'$. The main challenge lies in the non-monotonic nature of the envy measure, requiring caution in the application of dueling ULCB. Instead, we use the the main insight from UCB and LCB to construct upper and lower estimates of the true envy and leverage these estimates in a dueling fashion. Our main results show that this approach allows us to identify allocations with optimal envy incurring sub-linear regret.

The third extension concerns stable matching, shifting the emphasis from maximizing an objective to the pursuit of a stable solution or determining its nonexistence. The key observation is that in many setting stability constraints are ``local'' constraints that involve a small number of agents. Therefore, in each period, the algorithm selects a constraint that violates stability the most to collect feedback. We furnish an algorithm that allows us to identify stable matchings in all but $O(\log T)$ epochs on average. 


In each section, we assume computational oracles exist to solve certain subproblems. Our focus will remain on sampling complexity rather than computational efficiency. Even if the assumed oracles for computing optimal decision allocations in each period are precise, challenges still arise when determining which feedback to select. Our analysis can be readily extended to include approximation oracles. In the examples discussed in this work, we can regard these oracles (exact or approximate) as linear or integer programs that solve the static problem.

\subsection{Related Works}
Our paper contributes to the existing literature on online allocations by integrating two unique aspects. Firstly, we incorporate learning with noisy (bandit) feedback, which enhances the adaptability of our approach. Secondly, we impose strict constraints on the number of feedback instances.  We elaborate on these differences in comparison to the three existing lines of work.

\textit{Online fair allocation with noiseless feedback:} The online fair allocation, where items arrive dynamically and must be allocated to agents without revocation, has received extensive attention in the literature. Research has explored allocations that adhere to principles of fairness and efficiency \citep{aleksandrov2015online, aleksandrov2017expected, aleksandrov2017most, aleksandrov2017pure, aleksandrov2019strategy, benade2023fair}. Recent efforts have considered online max-min fair allocations in adversarial settings \citep{kawase2022online,fikioris2023online}, and have developed approaches aimed at maximizing welfare or minimizing envy with full or partial information \citep{markakis2011worst, benade2018make, banerjee2022proportionally, benade2022dynamic, banerjee2023online}. All the existing body of work assumes noiseless utility, where the true utility is observed in each allocation instance. In such cases, no learning mechanisms are involved, and the focus remains on efficiently achieving online fair allocation objectives. However, in many real examples, the precise observation of utility is not always possible, necessitating the handling of noisy feedback regarding the utility of the item received by the agent.

\textit{Online fair allocation with bandit learning:} Recent developments in online fair allocation have increasingly emphasized the utilization of bandit learning \citep{bistritz2020my, yamada2023learning, leshem2024fair}. These approaches are designed to tackle the challenge when the central planner does not have precise knowledge of agents' utilities. Diverging from traditional online algorithms, these approaches rely on noisy, estimated utilities obtained after item allocations. Moreover, they integrate the concept of UCB from multi-armed bandit problems to enhance the efficiency and fairness of online allocation processes. However, much of the current research assumes access to input from all participants at each allocation time-epoch, which may not be practical due to limitations on information gathering or the high cost associated with collecting feedback. To address it, we propose active learning strategies aimed at gathering the most informative feedback from a single agent (or few agents) per step. This strategic adaptation is necessary due to the limited feedback available at each time instance, rendering existing algorithms based solely on UCB techniques inadequate. 

\textit{Online stable matching with bandit learning:} Recent research has applied bandit learning techniques to the domain of online stable matching, effectively framing the two-sided competing matching problem within a sequential decision-making framework \citep{liu2020competing, cen2022regret, min2022learn, jagadeesan2023learning, li2023double, li2023rate, muthirayan2023competing}. 
For instance, \cite{liu2020competing} tackle the centralized multi-agent multi-armed competing bandit problem, where arms' preferences over agents are known, while agents' preferences over arms need to be learned from historical data. This work marks one of the pioneering efforts in online stable matching, considering the scenario where agents learn their preferences through bandit techniques. Subsequent studies have explored various aspects of bandit learning in online stable matching, such as handling unknown true preferences from both sides \citep{cen2022regret}, episodic reinforcement learning settings \citep{min2022learn}, incorporating contextual information \citep{li2023rate}, and time-varying matching \citep{muthirayan2023competing}. However, existing works typically assume observable noisy feedback from all matched pairs at each time instance, enabling the application of UCB-type or simple ETC-type algorithms. In our setting, feedback collection is costly and only one feedback is observed at each time. This motivates us to devise new dueling-type algorithm to incorporate both UCB and LCB of the estimated utility to address these challenges.

\section{Max-min Fairness for Unit Demand Agents}\label{sec:MAB}
In this section, we explore the proposed fair allocation framework tailored to unit demand agents, wherein a central platform distributes indivisible resources among agents who each consumes a single unit of a good. Following sections will expand this framework to encompass more diverse scenarios.

In the online setting, at each epoch $t=1,2,\cdots,T$, the platform allocates $K$ out of the $N$ goods (represented by $\mathcal{N}=\{1,2,3,\cdots,N\}$) among the $K$ agents (represented by $\mathcal{K}=\{1, 2, \cdots, K\}$).
Upon receiving a good $i$, the agent $j$ receives a noisy reward that is given as $X_{ij}(t)=\mu^j_i+\epsilon_{ij}(t)$, where $\mu^j_i$ is the unknown true reward of agent $j \in \mathcal{K}$ being assigned an item $i \in \mathcal{N}$. The $\epsilon_{ij}(t)$ are subgaussian distributions with mean $0$ and variance $\sigma^2$ for some $\sigma >0$, and are independent and identically distributed across $t$. The goal is to assign at most one good to each agent in order to achieve max-min fairness, ensuring that the reward of the lowest-rewarded agent is maximized. Specifically, let $\mathcal{M}$ denote the set of all possible allocations of agents to goods, represented by $\phi: \mathcal{K}\to \mathcal{N}$. 
When the true reward $\mu^j_i$ is known, the optimal max-min objective is
\begin{equation}\begin{aligned}\label{max:min:unit:demand}
\mu_{\maxmin}=\max_{\phi \in \mathcal{M}} \;\;\min_{j \in \mathcal{K}} \mu^{j}_{\phi(j)}.  
\end{aligned}\end{equation} 


To evaluate the performance of a given allocation policy that allocates a good $\phi_t(j)$ to the agent $j$ at time $t$, we employ the expected cumulative regret. This metric measures the accumulated difference between the objective from the proposed policy and the optimal max-min objective. Specifically, the expected cumulative regret over time horizon $T$ is 
\begin{equation}\begin{aligned}\label{MAB:rgret:cum}
R_T=\mathbb{E}\left[\sum_{t=1}^{T}\left( \mu_{\maxmin}-\min_{j \in \mathcal{K}}\mu^{j}_{\phi_t(j)}\right)\right].
\end{aligned}\end{equation}
The goal is to design an online allocation policy to minimize this expected cumulative regret. The difficulty lies in the combination of an unknown true reward that needs to be learned gradually over time and the constraint of a single feedback in each period. These challenges underscore the need for developing a new method for fair online allocation.


\subsection{A Special Example: Top $K$-Arm Identification}\label{sec:top-K}
To better understand the challenge and the  intuition behind our approach, we start with a special case of this problem when the rewards are agent-independent, i.e., $\mu^{j}_i=\mu_i$ for all $j \in \mathcal{K}$.  In this case, we can conceptualize the problem in a multi-arm bandit (MAB) setting. Each good $i$ corresponds to an arm with an unknown reward $\mu_i$, and the max-min allocation problem reduces to identifying the set of the best $K$ arms.
When $K=1$, the problem reduces to the classical MAB problem, where a single arm is selected at each instance to identify the arm with the highest reward. One of the most widely employed algorithms in this context is the Upper Confidence Bound (UCB) algorithm, which operates on the principle of optimism in the face of uncertainty, pulling the arm that has the largest UCB~\citep{lattimore2020bandit}. Specifically, let's define $T_i(t)$
as the number of times an arm $i$ has been pulled till time $t$, and additionally let \(\hat{\mu}_i(t)=\sum_{l=1}^{T_{i}(t)} X_i(l)/T_{i}(t)\) denote the average of the collected noisy rewards obtained by pulling arm $i$.
At time $t$, the UCB for arm $i$ is defined as 
\begin{center}
	\begin{equation}\label{UCBdef}
	\begin{aligned}
	\bar{\nu}_i(t)=\hat{\mu}_i(t)+\sqrt{\frac{1}{T_{i}(t-1)}\,2\sigma^2 \log t^{\alpha}}.
	\end{aligned}
	\end{equation}
\end{center}
Here, $\sqrt{\frac{1}{T_{i}(t-1)}2\sigma^2 \log t^{\alpha}}$ serves as a bonus term, ensuring that the algorithm explores the arms optimistically in the face of uncertainty and $\alpha$ is a constant larger than $2$~\citep{lattimore2020bandit}. We implicitly assume $\alpha>2$ and $T>N$ in all our theoretical analysis and fix $\alpha=3$ in all experiments. 
The UCB algorithm, at each time instance $t$, selects the $i$ with the highest $\bar{\nu}_i(t)$ value.  
This algorithm has been analyzed extensively and it is known that each sub-optimal arm is pulled $O(\log T)$ times which gives an $O(N \log T)$ cumulative regret over total time horizon $T$ \citep{lattimore2020bandit}.



For a general value of $K>1$, previous work has explored the selection of the $K$ best arms \citep{JMLR:v18:16-206, kalyanakrishnan2012pac,zhou2022approximate}. However, these studies necessitate pulling all $K$ arms in each period. We constrain the feedback so that only one arm can be pulled. This introduces a new challenge, and to our knowledge, prior work does not specifically address this constraint. To demonstrate the difficulty, consider a simple MAB setting with $3$ arms and the objective is to find the second best arm. If we use an analogue of the UCB algorithm here, i.e., find all the UCB estimates of the arms and choose the second highest UCB, then the algorithm fails to find the second best arm. This occurs because the error added to UCB increases if that arm is not explored. Consider the true arm ordering as $\mu_1>\mu_2>\mu_3$. It's possible that at a certain point $t$, the order of UCB does not align with the order of the real rewards, leading to a scenario such as $\bar{\nu}_2(t) > \bar{\nu}_1(t) > \bar{\nu}_3(t)$. If the algorithm continues exploring arm 1, $\bar{\nu}_3(t)$ and $\bar{\nu}_2(t)$ will keep increasing. This shall result in the ordering $\bar{\nu}_2(t) > \bar{\nu}_3(t)>\bar{\nu}_1(t).$ The algorithm then starts exploring arm 3, and keeps continuously exploring arms 1 and 3, while arm 2 remains unexplored. As a result, the algorithm fails to converge to the correct solution. Figure \ref{fig:toy:example} in the simulation section shows that the cumulative regret of this ``Second Best UCB" algorithm is linear in a MAB simulation with $3$ arms. 


To fix this, our proposal is to incorporate the Lower Confidence Bound (LCB) given as 
\begin{equation}\begin{aligned}\label{LCBdef}
\underline{\nu}_i(t)=\hat{\mu}_i(t)-\sqrt{\frac{1}{T_{i}(t-1)}\,2\sigma^2 \log t^{\alpha}}.
\end{aligned}\end{equation}
One notes that the LCB estimate has to be used in a correct fashion for the idea to work. An example of an incorrect usage of LCB is to select the arm with the $K$-th highest LCB estimate. This approach also fails. Consider, again, the arm orderings as $\mu_1>\mu_2>\mu_3$. Then it is possible that at a certain point we shall have $\bar{\nu}_2(t) > \bar{\nu}_1(t) > \bar{\nu}_3(t)$ and $\underline{\nu}_2(t) > \underline{\nu}_3(t) > \underline{\nu}_1(t)$. In this case the algorithm explores arms 1 and 3, while never exploring arm 2. Therefore this approach also fails.

The key idea to remedy such issues is a new procedure called Dueling ULCB. It first identifies the arms with the top $K$ UCB estimates and then selects the arm with lowest LCB estimate among the selected $K$ arms. Our novel method effectively mitigates the challenges encountered in both of the aforementioned scenarios.  Assume that we only select arms 1 and 3 for exploration. Then, for sufficiently large $\bar{t}$, we will have $\bar{\nu}_1(\bar{t}) > \bar{\nu}_3(\bar{t})$ and $\underline{\nu}_1(\bar{t}) > \underline{\nu}_3(\bar{t})$ as the UCB and LCB values under continuous exploration shrink to the true estimates. Moreover, $\underline{\nu}_1(\bar{t}) > \underline{\nu}_2(\bar{t})$ since $\underline{\nu}_2(\cdot)$ is a decreasing function while arm $2$ is not pulled. Also, with sufficient pulls of arm $3$, $\bar{\nu}_2(\bar{t}) > \bar{\nu}_3(\bar{t})$. This implies that arms $1$ and $2$ are identified by UCB, and among them arm $2$ has a lower LCB which leads it to be being pulled. Among many other natural considerations, we show that this  idea is highly effective, enabling us to address a wide range of problems. As shown in Figure \ref{fig:toy:example}, our approach demonstrates a substantial improvement over the ``Second Best UCB" algorithm. 




\subsection{Our  Algorithm and Regret Analysis}
Now, we revisit the general fair online allocation problem, where we can represent each agent-good pair as an arm with $\mu^j_i$  for $j=1,2,\cdots,K$ and $i=1,2,\cdots,N$ as the true reward.
The UCB and LCB estimates for the reward agent $j$ receives from good $i$ are given as 
\begin{equation*}
\begin{aligned}
&\bar{\nu}^j_i(t)=\hat{\mu}^j_i(t)+\sqrt{\frac{1}{T^j_{i}(t-1)}\,2\sigma^2 \log t^{\alpha}}\\ &\underline{\nu}^j_i(t)=\hat{\mu}^j_i(t)-\sqrt{\frac{1}{T^j_{i}(t-1)}\,2\sigma^2 \log t^{\alpha}},
\end{aligned}
\end{equation*}
where $T^j_{i}(t)$ is the number of times that good-agent pair $(i,j)$ has been chosen up to period $t$, and  
\(\hat{\mu}^j_i(t)=\sum_{K=1}^{T^j_{i}(t)} X_{ij}(K)/T^j_{i}(t)\) is the estimate of the true reward.

Using the idea of dueling, we exploit the UCB and the LCB estimates to present an online algorithm which identifies the max-min allocation in most iterations. To achieve this goal, we need to balance between the need to acquire more knowledge about the reward distributions of each of the arms (exploration) and the need to estimate the max-min reward based on its current knowledge (exploitation). Before presenting our general Dueling ULCB algorithm in Algorithm \ref{algo:ucblcb}, we introduce two oracles $\tilde{O}_1$ and $\tilde{O}_2$ that solve the corresponding static problems. Specifically, given any matrix $\mathbf{X} \in \mathbb{R}^{K\times N}$, $\tilde{\mathcal{O}}_1$ returns 
$\phi^*\in \arg \max_{\phi \in \mathcal{M}}\min_{j \in \mathcal{K}} x_{j\phi(j)}$,
which is the assignment that gives the max-min allocation when $x_{ji}=\mu^j_i$. This problem can be solved via linear programming~\citep{golovin2005max}. 
The second oracle $\tilde{\mathcal{O}}_2$ returns the minimum of a given a set of numbers. 
In our algorithm, we solve the max-min problem via $\tilde{\mathcal{O}}_1$ using UCB values of the arms and identify the minimum via $\tilde{\mathcal{O}}_2$ using the LCB values of the arms.
In this work, we do not consider the algorithmic complexities of the oracles, but rather focus on the learning algorithm assuming the existence of such oracles. This is justified in our setting as we consider feedback to be costly and do not concern ourselves with the computational work required to solve the problems at each epoch. Although we do not detail here, approximation algorithms can be used to substitute our oracles with a corresponding loss of efficiency in regret. For each $j \in \mathcal{K}$, denote by $\bar{\bm{\nu}}^j(t)=(\bar{\nu}^j_1(t),\bar{\nu}^j_2(t),\bar{\nu}^j_3(t),\cdots, \bar{\nu}^j_N(t))$ as the vector of all UCB estimates of the goods for agent $j$. Algorithm \ref{algo:ucblcb} starts by pulling each arm once. Following this, the algorithm sequentially estimates the UCB and LCB values for each arm. At each epoch, the algorithm uses the UCB estimates of the arms and the first oracle to compute the max-min allocation. Finally, the algorithm explores (seeks feedback from) the arm in the max-min allocation with the lowest LCB estimate. 

\begin{algorithm}[htb]
	\textbf{Input} $K, \alpha, \sigma^2$.\\
	\textbf{for} $t=1,2,\cdots,N\times K$\\
	Pull each arm once.\\
	\textbf{Update}:\\
	$T^j_i(t)$, $\hat{\mu}^j_i(t)$, $\bar{\nu}^j_i(t)$ and $\underline{\nu}^j_i(t)$.\\
	\textbf{for} $t=N+1,\cdots,T$ \textbf{do}:\\
	UCB: 
	$\bar{\nu}^j_i(t)=\hat{\mu}^j_i(t)+\sqrt{\frac{1}{T^j_{i}(t-1)}\,2\sigma^2 \log t^{\alpha}}$\\
	and \\
	LCB: $
	\underline{\nu}^j_i(t)=\hat{\mu}^j_i(t)-\sqrt{\frac{1}{T^j_{i}(t-1)}\,2\sigma^2 \log t^{\alpha}}$\\
	Identify max-min allocation, $\phi_t$ using UCB values, i.e.,  $\phi_t=\tilde{\mathcal{O}}_1(\bar{\bm{\nu}}^1(t),\bar{\bm{\nu}}^2(t),\cdots,\bar{\bm{\nu}}^N(t))$.\\
	Select the arm with the lowest LCB from the selected allocation as $I_t=\tilde{\mathcal{O}}_2(\phi_t, \underline{\bm{\nu}}(t))$.\\
	\textbf{Pull} the arm $I_t$ and \textbf{Output} $(\phi_t,I_t)$.
	\caption{Dueling ULCB Algorithm}
	\label{algo:ucblcb}
\end{algorithm}

Next we study the theoretical properties of the proposed Dueling ULCB algorithm. We show that it indeed chooses the correct max-min allocation in most cases. The regret defined in \eqref{MAB:rgret:cum} thus reduces to 
$
R_T=\mathbb{E}\big[\sum_{t=1}^{T} \mu^{j^*}_{\phi^*(j^*)}-\min_{j \in \mathcal{K}} \mu^{j}_{\phi_t(j)}\big]
$
where $\phi^*$ and $j^*$ are a max-min allocation and an agent who receives the max-min allocation respectively and $\phi_t$ is the allocation chosen using Algorithm~\ref{algo:ucblcb}.  Note that the regret is not defined with respect to the revealed arm, but with respect to the allocation $\phi_t$. Observe that the regret is non-negative for each $t$ since $\min_{j \in \mathcal{K}}\mu^j_{\phi_t(j)} \le \mu^{j^*}_{\phi(j^*)}$,  as at least one of the chosen agents receives a suboptimal allocation compared to $j^*$ in the optimal solution. The main rationale for this choice of regret is that we wish to select the correct set by revealing only one arm. To establish our main results, we need to state an assumption on the true rewards. Define $$\Phi^*=\left\{\phi \in \mathcal{M}: \, \min_{j \in \mathcal{K}} \mu^{j}_{\phi(j)}=\mu_{\maxmin} \right\}$$
as the set of allocations having the same true minimal allocation which is the max-min objective.
\begin{assumption}\label{assm:gap:MAB}
	For any $\phi \in \mathcal{M}\backslash \Phi^*$, there exists a gap $\Delta_{\min}>0$ such that 
	\[\mu_{\maxmin}-\min_{j \in \mathcal{K}} \mu^{j}_{\phi(j)}>\Delta_{\min}.\] 
	Furthermore, for any $i_1,i_2 \in \mathcal{N}$ and $j_1, j_2 \in \mathcal{K}$, one has 
	\[\left|\mu^{j_1}_{i_1}-\mu^{j_2}_{i_2}\right|\le \Delta_{\max}.\]
\end{assumption}
Assumption~\ref{assm:gap:MAB} serves as a measure of identifiability for the max-min allocation. It's worth noting that this assumption is a generalization of the gap assumption in classical MAB setting \citep{lattimore2020learning}, which assumes a gap between the top arm and all other arms. In our case, we posit a gap between our desired set of allocations (which may not be unique) and the remaining allocations. Moreover, Assumption~\ref{assm:gap:MAB} also requires that the differences between true rewards are upper bounded. This is a minor assumption and can be satisfied when the true reward is bounded.


\begin{theorem}\label{thm:unit:demand}
	Under Assumption~\ref{assm:gap:MAB}, the expected cumulative regret of Algorithm~\ref{algo:ucblcb} satisfies 
	\[R_T \le 3\,\Delta_{\max} \, N\, K\left[ \frac{\left(\sqrt{2\alpha}+2\right)^2 \, \sigma^2}{\Delta^2_{\min}}\,  \log T+2\,\frac{\alpha-1}{\alpha-2}+2\right].\]
\end{theorem}
Next we discuss the regret bound with respect to a few key terms. First, the regret bound inflates with a decrease in $\Delta_{\min}$ as it becomes harder to identify the correct allocations. Further note that as $\Delta_{\max}$ increases, the regret increases as we penalize more when an incorrect allocation is chosen. With respect to the time horizon, the regret is sub-linear with rate $O(\log T)$. This implies that our algorithm is able to select the correct max-min allocation in most iterations. It's worth noting that Algorithm~\ref{algo:ucblcb} incorporates the LCB step, enabling us not only to identify the true allocation but also the arm that achieves the true max-min objective.
Finally, it's useful to note that when there is only one agent and the platform's task is to assign the agent the best item, the problem simplifies to the classical MAB setting where the objective is to obtain the best reward. By substituting $K=1$ in the regret bound in Theorem~\ref{thm:unit:demand}, we get the same regret $O(N\, \log T)$ as that for the classic MAB setting. 

In contrast to the proof in the classic MAB, our setting with a general $K>1$ presents two key differences: the regret is based on a set rather than a single arm, and there are constraints on how we can explore the arms, specifically in terms of allocations. These factors make the problem considerably more challenging, and we leverage the properties of UCB and LCB to address them. In our approach, the regret bound is established using the ranking, where we demonstrate that the rankings of the UCB, LCB, and the true values should align after a sufficient number of arm pulls. 

The regret bound in Theorem \ref{thm:unit:demand} relies on the gap assumption and is known as the instance-dependent bound. Next we provide a regret bound that is instance-independent. To ease the presentation, we consider the case where the rewards are identical for all agents, i.e., $\mu^{j}_i=\mu_i$ for all $j \in \mathcal{K}$. Note that in this setting, the max-min allocation is the set of the top $K$ arms, and the $K$-th reward is the max-min objective. We refer to the true top $K$ set as $G^*$. In this case, the regret defined in \eqref{MAB:rgret:cum} reduces to 
$R_T=\mathbb{E}\left[\sum_{t=1}^{T}\left(\min_{i \in G^*} \mu_i - \min_{j \in G_t} \mu_j\right)\right],$
where $G_t$ is the top $K$ arms identified at time $t$ using Algorithm~\ref{algo:ucblcb}. In this setting, the oracle $\tilde{\mathcal{O}}_1$ simply outputs the order-K statistic and we only need one oracle in this case. Define $\Delta_i(K)=|\mu_{(K)}-\mu_{i}|$ and
\[G(\delta)=\left\{i \in \mathcal{N}: \, \mu_{(K)}-\mu_i> \delta\right\}.\]

\begin{proposition}\label{MAB:prop:gap:ind}
	In the setting $\mu^{j}_i=\mu_i$ for all $j \in \mathcal{K}$, the regret of Algorithm~\ref{algo:ucblcb} over horizon $T$ satisfies
	
	\begin{equation*}
	\begin{aligned}
	R_T & \le 2\, \sqrt{\left((N-K+1)^2-1\right)\, K\, 8 \, \sigma^2\alpha\, T \,\log T} \\
	&\quad +\sum_{i \in G(0)} \left(\frac{ 2\, K\,\Delta_i(K)\, \alpha}{\alpha-2}+\frac{ (N-K)\,K\, \Delta_i(K)\, \alpha}{\alpha-2}\right).
	\end{aligned}
	\end{equation*}
\end{proposition}

Since the second term in Proposition~\ref{MAB:prop:gap:ind} is independent of the time horizon $T$, the overall rate of the regret bound in Proposition~\ref{MAB:prop:gap:ind} with respect to the time horizon is of the order $O( \sqrt{ T\, \log T})$. 
Note that the regret is defined in terms of the set selected and not the arm pulled. Therefore, in the case the correct minimal arm is not pulled, the regret is inflated by a factor of $\sqrt{N}$ to uniformly account for all the cases when the explored arm is not in the optimal set. Further note that in the case $K=1$, such inflation of $\sqrt{N}$ does not appear and hence our rate shall match that in the MAB \citep{lattimore2020bandit}. Thus the inflation of regret occurs due to definition of the regret and the uniform bounding of it over all cases. Note that when $K=N$, the regret is zero. This is expected as the set $G(0)=\{i: \, \mu_i < \mu_{\maxmin}\}$ becomes smaller when $K$ increases and eventually becomes empty when $K=N$.

The primary challenge in this problem lies in addressing the additional errors resulting from regret being calculated with respect to the set rather than the explored arm in the case where there is a significant difference between the arms. To address this, we divide the problem into two cases- where the pulled arm is the true minimum of the considered set, and the other where it is not.

\section{Allocation of Bundles}\label{sec:CMAB}
In this section, we expand upon the fair allocation framework introduced for unit demand agents in Section \ref{sec:MAB} to encompass scenarios involving bundles of goods.
\subsection{Max-Min Fairness}\label{subsec:Max-Min:Fairness}
We consider a scenario with a set of agents, labeled $\mathcal{K}=\{1,2,\cdots,K\}$ and a set of goods, labeled $\mathcal{N}=\{1,2,\cdots,N\}$. 
A \emph{bundle} is defined as a subset of goods. Each agent is allocated a bundle in such a way that for any two agents, their bundles are disjoint. The list of bundles assigned to the agents collectively is referred to as an \emph{allocation}.

Our main objective is to identify an allocation that optimizes fairness by maximizing the least reward attained by any agent. This goal extends the principle of max-min fairness to the case of bundles, rather than individual goods outlined in Section~\ref{sec:MAB}. Similar to our previous setting, we assume active feedback framework and require sampling the reward of only one agent. 

\FloatBarrier
\begin{figure}[!ht]
\begin{center}
\begin{tikzpicture}
    \node at (1,2) {Item:};
    \node[circle, draw] (top1) at (3,2) {1};
    \node[circle, draw] (top2) at (4,2) {2};
    \node at (5,2) {\dots};
    \node[circle, draw] (topi) at (6.5,2) {\(i\)};
    \node[circle, draw] (topi1) at (7.5,2) {$\phantom{i}$};
    \node[circle, draw] (topi2) at (8.5,2) {$\phantom{i}$};
    \node at (10,2) {\dots};
    \node[circle, draw] (topN) at (11,2) {N};
    \node[ellipse, dashed, red, label=above: $\phi(j)$, draw, inner sep=0pt, fit=(topi)(topi1)(topi2)] {};
    
    \node at (1,0) {Agent:};
    \node[draw] (bottom1) at (3,0) {1};
    \node[draw] (bottom2) at (4,0) {2};
    \node at (5,0) {\dots};
    \node[draw] (bottomj) at (6,0) {\(j\)};
    \node at (7,0) {\dots};
    \node[draw] (bottomK) at (8,0) {K};
    
    \draw[->] (topi) -- (bottomj) node[midway, left]{$\mu_i$};
    \draw (topi1) -- (bottomj);
    \draw (topi2) -- (bottomj);
       
\end{tikzpicture}
\end{center}
\caption{Illustration of an allocation with bundle $\phi(j)$ being allocated to agent $j$.}
\label{fig:allocation:diagram}
\end{figure}
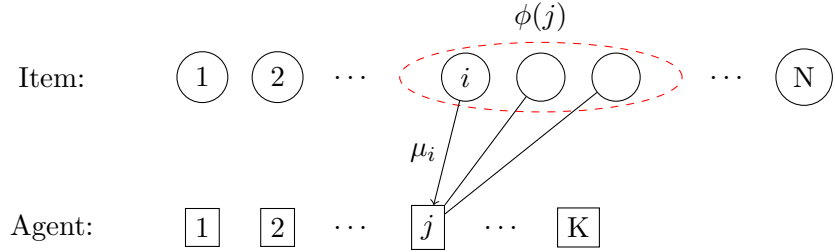

We assume, each agent $j \in \mathcal{K}$ has access to certain sets of bundles of  goods which is denoted as $\mathcal{A}_j \subseteq 2^{\mathcal{N}}$.  We call $\mathcal{A}_j$ as the set of feasible bundles for agent $j$. We assume $\emptyset\in \mathcal{A}_j$.  The individual set $\mathcal{A}_j$ captures diverse geographical and technological constraints that vary among agents. We assume each good has a common true ``quality'' $\mu_i$. Our result extends to the case where these qualities are agent-specific as well.    For agent $j \in \mathcal{K}$,  the reward for agent $j$, is given by a function 
$$r^j:\mathbb{R}^N\times \mathcal{A}_j\to \mathbb{R}$$   
In particular, given the true ``quality'' $\bm{\mu}=(\mu_1,..,\mu_N)$ and a feasible bundle
$S$ in $\mathcal{A}_j$, the true reward  of agent $j$ for receiving $S$ is  $r^{j}(\bm{\mu}; S)$.  Note that in this context, we assume that the reward for a set $S$ solely depends on the quality of items in $S$. However, to keep our notation simpler, we specify all qualities as arguments to the reward function $r^{j}(\bm{\mu}; S)$.

An allocation assigns a feasible bundle to each agent, ensuring that no item is assigned to more than one agent. In particular, 
\[
\mathcal{M}=\left\{ \phi:\mathcal{K} \to \overset{K}{\underset{j=1}{\cup}}\mathcal{A}_j :\phi(j) \in \mathcal{A}_j \text{ for all} \, j \in \mathcal{K}, \text{ and } \forall i\ne j, \phi(i)\cap \phi(j)=\emptyset  \right\}
\]
denote the set of all allocations. For every allocation $\phi\in \mathcal{M}$,
$\phi(j)$ is the bundle assigned to $j$.\footnote{Note that, for simplicity, we require $\phi(i)\cap \phi(j)=\emptyset$, meaning each good has a capacity of 1. Our model extends to the general case where each good has a finite capacity.}


In this setting we define the max-min problem as 
\begin{equation}\begin{aligned}\label{max-min:obj:Combinatorial}
\argmax_{\phi \in \mathcal{M}}\min_{1\le j \le K} r^j(\bm{\mu};\phi(j)),
\end{aligned}\end{equation}
and denote $\opt^{\maxmin}$ as its oracle max-min objective when the true reward values are observed, i.e. $\opt^{\maxmin}=\max_{\phi \in \mathcal{M}}\min_{1\le j \le K} r^j(\bm{\mu};\phi(j))$.
Note that there might be multiple solutions to the max-min objective. Thus, with some minor abuse of notation, as in Section~\ref{sec:MAB}, define $\Phi^*=\{\phi: \, \min_{j \in \mathcal{K}} r^{j}(\bm{\mu}; \phi(j))=\opt^{\maxmin}\}$ as the optimal set of allocations. However, we posit that presenting any of these solutions is sufficient, and there is no imperative need to differentiate between them, as they all lead to the same reward outcome.

At every period $t$, if we choose to collect feedback of agent $j$ for bundle $S$, we receive a noisy signal on the quality of good $i\in S$,  $X_i(t)=\mu_i+ \epsilon_i(t)$, and the corresponding noisy reward $r^{j}(\mathbf{X}(t); S)$. To evaluate the performance of any allocation policy, we compare its decision against an optimal benchmark that assumes full knowledge of the base reward vector and reward function. Thus, for any allocation policy that allocates $\phi_t(\cdot)$ at time $t$, its overall regret over time horizon $T$ is quantified by
\begin{equation}\label{regret:def:Combinatorial}
\begin{aligned}
R_T=\mathbb{E}\left[\sum_{t=1}^{T} \left(\opt^{\maxmin}-\min_{j \in \mathcal{K}} r^j(\bm{\mu}; \phi_t(j))\right)\right].
\end{aligned}
\end{equation}

Considering that we're dealing with combinations of items, this scenario fits within the framework of the Combinatorial Multi-Armed Bandit (CMAB), which extends the conventional MAB model in Section~\ref{sec:MAB}. In CMAB, at each time step a player selects a combination of arms, termed as a super-arm, instead of a single arm. The reward from pulling a super-arm is determined by the  rewards of its constituent individual arms. In the typical CMAB scenario, the objective is to identify the super-arm that yields the highest reward. In this case, algorithms leveraging UCB estimates of the base arms have been developed to achieve sub-linear regret~\citep{chen2013combinatorial}. However, our max-min fair allocation problem diverges from the standard CMAB framework: we are restricted to selecting super-arms that constitute a partition of $\mathcal{N}$, and our allocations consist of sets of super-arms rather than individual base arms. 
In our specific context, relying solely on algorithms utilizing UCB estimates for the base arms proves ineffective. This issue is explored in Section~\ref{sec:MAB}, where we present examples demonstrating the limitations of solely adjusting the UCB algorithm. To attain the desired results, it is imperative to also incorporate LCB estimates of the arms. 

\subsection{Our  Algorithm and Regret Analysis}
We present the algorithm to solve the online max-min allocation problem. In order to do this, we define the two oracles that are necessary for our algorithm. Define $\mathcal{O}_1$ as the oracle which takes a vector of $N$ entries, the set of all super-arms, and the number of agents $K$ and returns a max-min allocation. Namely, given any $\textbf{x} \in \mathbb{R}^N$ and reward function $r$, $\mathcal{O}_1$ solves \eqref{max-min:obj:Combinatorial} where $\bm{\mu}=\textbf{x}$. Additionally, we define another oracle, denoted as $\mathcal{O}_2$, which accepts a vector comprising $N$ entries, which is a reward estimate for each item,  along with an allocation, and outputs the agent with the lowest reward. Essentially, this oracle operates similar to a quicksort algorithm. Again, as in Section~\ref{sec:MAB}, note that since we do not know the true rewards, i.e., $\bm{\mu}$, we must learn it through agent feedback. Denote the UCB vector of the base arms as $\bar{\bm{\nu}}(t)=(\bar{\nu}_1(t),\bar{\nu}_2(t),\cdots,\bar{\nu}_N(t))$ and the LCB vector of the base arms as $\underline{\bm{\nu}}(t)=(\underline{\nu}_1(t),\underline{\nu}_2(t),\cdots,\underline{\nu}_N(t))$. Our Dueling Max-Min ULCB is shown in Algorithm \ref{algo1}.

\begin{algorithm}
	\textbf{Input}: $K, \sigma, \alpha$.\\
	\textbf{for}: $t=1,2,\cdots, N$;\\
	Pull each arm.\\
	\textbf{Update}: $T_t(i), \hat{\mu}_i(t),\bar{\nu}_i(t), \underline{\nu}_i(t)$.\\
	\textbf{for} $t=N+1,\cdots,T$ \textbf{do}:\\
	UCB: $\bar{\nu}_i(t)=\hat{\mu}_i(t)+\sqrt{\frac{1}{T_{i}(t-1)}\,2\sigma^2 \log t^{\alpha}}$.\\
	and \\
	LCB: $
	\underline{\nu}_i(t)=\hat{\mu}_i(t)-\sqrt{\frac{1}{T_{i}(t-1)}\,2\sigma^2 \log t^{\alpha}}$.\\
	$(\phi_t,j_t)=\mathcal{O}_2(\underline{\bm{\nu}}(t),\mathcal{O}_1(\bar{\bm{\nu}}(t),\overset{K}{\underset{i=1}{\cup}} \mathcal{A}_j,K),K)$.\\
	\textbf{Output and Pull} $(\phi_t,j_t)$ for $t=N+1,2,\cdots,T$.
	\caption{Dueling Max-Min ULCB Algorithm}
	\label{algo1}
\end{algorithm} 
The algorithm operates as follows: initially, all base arms are pulled at least once. Subsequently, at each time step, the UCB and LCB estimates of the arms are updated. Based on these, the oracle $\mathcal{O}_1$ is used to obtain the max-min allocation and the oracle $\mathcal{O}_2$ is used to decide the arm to pull for collecting feedback. 
Our algorithm progressively transitions from exploration to exploitation. Initially, there may be incorrect allocations due to inadequate exploration. However, as time passes, the number of mistakes decreases, eventually resulting in correct identifying the max-min allocation. 

To theoretically substantiate this, we next show that our algorithm achieves a sub-linear regret bound. Define $\phi^*=\mathcal{O}_1(\bm{\mu},\overset{K}{\underset{i=1}{\cup}}\mathcal{A}_j,K)$ and $j^*=\mathcal{O}_2(\bm{\mu},\phi^*,K)$ as the true max-min allocation and the agent receiving the max-min objective, respectively.
Further note that, in this setting, the regret as defined in \eqref{regret:def:Combinatorial} reduces to
\begin{equation*}
\begin{aligned}
R_T=\mathbb{E}\left[\sum_{t=1}^{T}\left(r^{j^*}(\bm{\mu}; \phi^*(j^*))-\min_{1\le j \le K}r^{j}(\bm{\mu}; \phi_t(j))\right)\right]
\end{aligned} 
\end{equation*}
where $\phi_t(\cdot)$ is the allocation chosen at time $t$. Since both $j^*$ and $\phi(j^*)$ are optimal, this regret is always positive. It's important to note that this regret mirrors the definition of regret in the unit demand case. It's defined in relation to the true minimum of the selected allocation, rather than the revealed arm. 

We state our assumptions for the main results of this section. 
\begin{assumption}\label{assm:reward:Lipschitz}
	For any agent $j \in \mathcal{K}$ and any bundle $S \in  \mathcal{A}_j$, there is some positive constant $c$ such that
	$\left|r^{j}(\bm{\mu};S)-r^{j}(\bm{\nu};S)\right|\le c \sum_{i \in S}\left|\mu_i-\nu_i\right|$
	for any two vectors $\bm{\mu}$ and $\bm{\nu}$ in $\mathbb{R}^N$. Furthermore, if $\mu_i\le \nu_i$ for all $i \in S$, one has $r^{j}(\bm{\mu};S) \le r^{j}(\bm{\nu};S)$, where $\mu_i$ and $\nu_i$ represent the i-th element of $\bm{\mu}$ and $\bm{\nu}$, respectively.
\end{assumption}

Assumption~\ref{assm:reward:Lipschitz} implies that the total reward increases when the reward for an item increases while other rewards do not decrease. Also a change in the reward for each agent is bounded with change in the rewards of base items. Such assumptions are common in the literature on combinatorial multi-armed bandits~\citep{chen2013combinatorial}.

Next, we state one of our assumptions for the analysis in this setting which is akin to the gap assumption in the unit demand case.

\begin{assumption}\label{assm:gap}
	For any $\phi_1 \in \Phi^*$ and $\phi_2 \in \mathcal{M}\backslash \Phi^*$, there exists $\tilde{\Delta}_{\min}>0$ such that 
	\[ \left( \min_{j \in \mathcal{K}} r^{j}(\bm{\mu};\phi_1(j))-\min_{j \in \mathcal{K}} r^{j}(\bm{\mu};\phi_2(j))\right)> \tilde{\Delta}_{\min}.\]
	Further, for any $\phi_1, \, \phi_2  \in \mathcal{M}$ and pair $i,j \in \mathcal{K}, \, i\ne j$, 
	\[\left|r^{j}(\bm{\mu};\phi_1(j))-r^{i}(\bm{\mu};\phi_2(i)) \right|\le \tilde{\Delta}_{\max}.\]
\end{assumption}
One may note that for any $\phi_1 \in \Phi^*$, $\min_{j \in \mathcal{K}} r^{j}(\bm{\mu};\phi_1(j))=\text{opt}^{\maxmin}$. Therefore Assumption~\ref{assm:gap} states that the optimal and the sub-optimal allocations are separated by some quantity $\tilde{\Delta}_{\min}$ in terms of the max-min objective. In essence, the optimal set of allocations are identifiable or discernible from the sub-optimal set of allocations by some identifiability constant in terms of the max-min objective. This identifiability assumption parallels Assumption~\ref{assm:gap:MAB} as stated in Section~\ref{sec:MAB}, in the bundle setting.

\begin{theorem}\label{coro:regret}
	Under Assumptions~\ref{assm:reward:Lipschitz}-\ref{assm:gap}, for Algorithm~\ref{algo1} , one has 
	\[R_T \le 3\,\tilde{\Delta}_{\max} \, N\left[ \frac{\left(\sqrt{2\alpha}+2\right)^2 c^2 \, N^2\, \sigma^2}{\tilde{\Delta}^2_{\min}}\,  \log T+\frac{\alpha-1}{\alpha-2}+2\right].  \]
\end{theorem}
Theorem~\ref{coro:regret} establishes a sub-linear rate for the regret of Algorithm~\ref{algo1}. Specifically, it makes at most $O(N^3\, \log T)$ errors within total time horizon $T$. Similar to the unit demand case, an increase in $\tilde{\Delta}_{\max}$ or a decrease in $\tilde{\Delta}_{\min}$ enlarges the regret bound. This is because, in the former scenario, errors are penalized more severely, while in the latter scenario, detection becomes more challenging due to a narrower gap. 

The primary challenge in establishing this result lies in the reduction of this problem from super-arms to base arms. The combinatorial bandit can be conceptualized as a MAB where the super-arms are treated as base arms. However, this approach faces two primary challenges. Firstly, the number of super-arms can be exponential in the number of base arms, which would make the regret bound loose if we substituted $N$ for the number of super-arms. Second, in the combinatorial setting, unlike in the unit demand setting in Section \ref{sec:MAB}, the arms interact with one another via the reward function and feasibility considerations. To overcome these challenges in our proofs, we employ the novel idea that incremental knowledge of a few entries in low-dimensional quality vector, $\bm{\mu}$, improves our knowledge about rewards for all super-arms that contain those entries, and hence making the analysis tractable.

\section{Minimal Envy Fairness}\label{sec:Envy}

In this section, we illustrate the versatility of the proposed dueling UCB and LCB approach beyond the conventional max-min fair allocation problem. Here, we extend our methodology to address minimal envy \citep{aleksandrov2015online,aleksandrov2017expected,aleksandrov2017pure} as an alternative fairness metric and tailor our method to identify online allocations that minimize envy.

The principle idea in envy is that each agent receives a satisfactory reward and does not desire another item. In this section, we define envy in the setting of our problem and propose an algorithm to minimize envy. As before, our set-up consists of $K$ agents and $N$ items. We start by defining the notion of envy among two agents. Given an allocation $\phi \in \mathcal{M}$ and any two agents $i,j \in \mathcal{K}$, the envy between the two agents $i,j$ under allocation $\phi$ is given as 
\[   ev_{i\rightarrow j}(\bm{\mu},\phi)=\max  \{r^{i}(\bm{\mu};\phi(j))-r^i(\bm{\mu};\phi(i)), 0 \} \]
where $ev_{i\rightarrow j}(\bm{\mu},\phi)$ quantifies the extra reward agent $i$ would attain upon acquiring the allocation originally designated for agent $j$. Further, for each allocation $\phi \in \mathcal{M}$, we define the envy of the allocation as 
\begin{equation}\begin{aligned}\label{fixed:all:envy}
ev(\bm{\mu},\phi)=\max_{i,j \in \mathcal{K}} ev_{i\rightarrow j}(\bm{\mu},\phi)
\end{aligned}\end{equation}
which is the maximum envy between any two agents in the allocation $\phi$.
Our objective is to find the allocation that minimizes envy \eqref{fixed:all:envy}. In other words, we seek to find:
\begin{equation}\begin{aligned}\label{envy:opt:set}
\phi^* = \argmin_{\phi \in \mathcal{M}} ev(\bm{\mu},\phi).
\end{aligned}\end{equation} 
Let $\mathcal{E}^*$ represent the set of all allocations that achieve true minimal envy, or in other words, solutions of equation \eqref{envy:opt:set}. To evaluate the policy that allocates $\phi_t$ at time $t$, we compare its envy objective to that of the optimal benchmark when the reward function and the true base rewards $\bm{\mu}$ are known. That is, for any $\phi^* \in \mathcal{E}^*$, we define the regret over time $T$ as  
\begin{equation}\begin{aligned}\label{envy:regret}
R_T=\mathbb{E}\left[\sum_{t=1}^{T}\left(ev(\bm{\mu};\phi_t)-ev(\bm{\mu};\phi^*)\right)\right]. 
\end{aligned}\end{equation}
By definition, this regret is always non-negative and is zero when $\phi_t \in \mathcal{E}^*$. 

In practice, the true reward vector $\bm{\mu}$ is typically unknown, requiring us to learn it from agents' sequential feedback. Unlike the active feedback framework discussed earlier, where feedback was restricted to a single agent, in this minimal envy setting, we can collect feedback from two agents at each time point. This adjustment is essential for computing envy fairness, as it enables comparisons between two agents to ensure a fair assessment.

In line with our previous discussions, we could approach the problem within a bandit framework by employing UCB and LCB estimates of the base arms. However, the dueling ULCB approach used in previous sections is inadequate here. Blindly applying UCB and LCB values fails to accurately estimate the true envy and leads to erroneous identification. To address this issue, our core idea is to maintain an optimistic estimate of the true reward that steadily increases until exploration. Conversely, the LCB embodies the principle of pessimism, serving as an underestimation of the true reward that diminishes until exploration. Drawing on this principle, we recognize the need for both an optimistic and a pessimistic estimate of envy, each dynamically adjusting until exploration. Recall that we denote the UCB vector of the base arms as $\bar{\bm{\nu}}(t)=(\bar{\nu}_1(t),\bar{\nu}_2(t),\cdots,\bar{\nu}_N(t))$ and the LCB vector of the base arms as $\underline{\bm{\nu}}(t)=(\underline{\nu}_1(t),\underline{\nu}_2(t),\cdots,\underline{\nu}_N(t))$.
For any allocation $\phi$ and agents $i,j \in \mathcal{K}$, we propose the following as the optimistic and pessimistic estimates of envy between agents $i$ and $j$, respectively 
\begin{equation}
\begin{aligned}
ev_{i \rightarrow j}(\bar{\bm{\nu}}(t), \underline{\bm{\nu}}(t), \phi_t)&=  \left(r^{i}(\bar{\bm{\nu}}(t);\phi(j))-r^i(\underline{\bm{\nu}}(t);\phi(i))\right)^{+},\\
ev_{i \rightarrow j}(\underline{\bm{\nu}}(t),\bar{\bm{\nu}}(t),\phi_t)&=\left(r^i(\underline{\bm{\nu}}(t);\phi(j))-r^i(\bar{\bm{\nu}}(t);\phi(i))\right)^{+},
\end{aligned}
\end{equation}
where $x^+=\max\{0,x\}$.
These values can be viewed as surrogates for the UCB and LCB estimates of envy. We designate the optimistic estimate as the upper estimate and the pessimistic estimate as the lower estimate of envy. Our key idea is to first use the lower estimate of the envy to solve 
\[\argmin_{\phi \in \mathcal{M}}\max_{i,j \in \mathcal{K}}   \, ev_{i \rightarrow j}(\underline{\bm{\nu}}(t),\bar{\bm{\nu}}(t),\phi_t),\]
which is a proxy of \eqref{envy:opt:set}, and then use the upper estimate to find the pair of agents in an allocation with maximum envy, namely,
\[\argmax_{i,j \in \mathcal{K}}  \, ev_{i \rightarrow j}(\bar{\bm{\nu}}(t), \underline{\bm{\nu}}(t), \phi_t).\]
This approach reverses our previous application of upper and lower estimates (UCB and LCB) for addressing the max-min objective. The core principle is that objectives needing minimization should be approached with underestimation of true values, whereas maximization objectives warrant overestimation. We propose that this dueling dynamic of underestimation and overestimation will guide us in iteratively uncovering the true objective.
Before presenting our algorithm, we assume the existence of two computational oracles $\mathcal{O}^E_1$ and $\mathcal{O}^E_2$ which solve \eqref{envy:opt:set} and \eqref{fixed:all:envy} respectively given any $\textbf{x}_1, \textbf{x}_2 \in \mathbb{R}^N$, i.e.,
\begin{align}
\begin{split}
\tilde{\phi}&=\argmin_{\phi \in \mathcal{M}}\max_{i,j \in \mathcal{K}}   \, ev_{i \rightarrow j}(x_1,x_2,\phi,)=\mathcal{O}^E_1(x_1,x_2,\overset{K}{\underset{j=1}{\cup}}\mathcal{A}_j,K) \label{oracle:solves:this_1}
\end{split}\\
\begin{split}
(\tilde{i},\tilde{j})&=\argmax_{i,j \in \mathcal{K}}  ev_{i \rightarrow j}(x_2,x_1,\phi_t)=\mathcal{O}^E_2(x_2,x_1,\tilde{\phi},K) \label{oracle:solves:this_2}.
\end{split}
\end{align}

\begin{algorithm}
	\textbf{Input}: $K, \sigma, \alpha$.\\
	\textbf{for}: $t=1,2,\cdots, N$;\\
	Pull each arm.\\
	\textbf{Update}: $T_i(t), \hat{\mu}_i(t),\bar{\nu}_i(t), \underline{\nu}_i(t)$.\\
	\textbf{for} $t=N+1,\cdots,T$ \textbf{calculate}:\\
	UCB: $\bar{\nu}_i(t)=\hat{\mu}_i(t)+\sqrt{\frac{1}{T_{i}(t-1)}\,2\sigma^2 \log t^{\alpha}}$.\\
	and \\
	LCB: $
	\underline{\nu}_i(t)=\hat{\mu}_i(t)-\sqrt{\frac{1}{T_{i}(t-1)}\,2\sigma^2 \log t^{\alpha}}$.\\
	$\phi_t=\mathcal{O}^{E}_1(\underline{\bm{\nu}}(t),\bar{\bm{\nu}}(t),\overset{K}{\underset{i=1}{\cup}} \mathcal{A}_j,K)$.\\
	$(i_t,j_t)=\mathcal{O}^{E}_2(\bar{\bm{\nu}}(t),\underline{\bm{\nu}}(t),\phi_t,K)$. \\
	\textbf{Pull} $(\phi_t(i_t),\phi_t(j_t))$ for $t=N+1,2,\cdots,T$.\\
	\textbf{Output} $(\phi_t,\phi_t(i_t),\phi_t(j_t))$ for $t=N+1,2,\cdots,T$.
	\caption{Envy-combinatorial ULCB Algorithm}
	\label{algo:envy:cucblcb}
\end{algorithm} 

Our Envy-combinatorial ULCB algorithm is presented in Algorithm \ref{algo:envy:cucblcb}. After pulling each arm once, at each time instance we update the UCB and LCB values. Based on these, we solve \eqref{oracle:solves:this_1} and \eqref{oracle:solves:this_2} using $\mathcal{O}^E_1$ and $\mathcal{O}^E_2$ respectively with $\textbf{x}_1=\bar{\bm{\nu}}(t)$ and $\textbf{x}_2=\underline{\bm{\nu}}(t)$, then explore the outputs by pulling the two super-arms returned by the algorithm.

We next establish the upper bound of the cumulative regret for our algorithm. It's important to note that regret only arises when $\phi_t \not\in \mathcal{E}^*$. To begin, we introduce an identifiability assumption.
\begin{assumption}\label{envy:assm:gap}
	For any $\phi_1 \not \in \mathcal{E}^*$ and $\phi_2 \in \mathcal{E}^*$, there exists $\Delta_{e,\min}>0$, such that 
	\begin{equation*}
	ev(\bm{\mu};\phi_1)-ev(\bm{\mu};\phi_2)\ge \Delta_{e,\min}.
	\end{equation*}
	Further, for any $\phi_1,\phi_2$ and any pairs $(i,j) \ne (i',j') \in \mathcal{K}\times \mathcal{K}$, there exists $\Delta_{e,\max}>0$ such that
	\begin{equation*}
	\left|ev_{i\to j}(\bm{\mu},\phi_1)- ev_{i'\rightarrow j'}(\bm{\mu};\phi_2) \right| \le \Delta_{e,\max}.
	\end{equation*}   
\end{assumption}

Assumption~\ref{envy:assm:gap} is an identifiability assumption for the set of optimal envy allocations; akin to the previous assumptions in Sections~\ref{sec:MAB} and~\ref{sec:CMAB}, allowing us to discern an allocation as being optimal envy or not.

\begin{theorem}\label{envy:main:thm}
	Under Assumptions~\ref{assm:reward:Lipschitz} and~\ref{envy:assm:gap}, we have
	\begin{equation*}
	\begin{aligned}
	R_T =O\left(\frac{N^3\, \log T}{\Delta^2_{e,\min}}\right).
	\end{aligned}
	\end{equation*}
\end{theorem}

Theorem~\ref{envy:main:thm} implies that Algorithm~\ref{algo:envy:cucblcb} makes at most $O(N^3\, \log T)$ errors in selecting the allocation with minimal envy. The challenges in the proof lie in the formulation of the problem, obtaining tail bounds and finally establishing the envy regret. Due to the need to account for comparisons between pairs of agents, envy is not necessarily a monotone function of the reward. This necessitates a more intricate analysis compared to previous formulations where the objective was more  amenable to simpler methods. To address these challenges, we prove the validity of the constructed upper and lower estimates of the reward by deriving tail bounds for them. Moreover, we prove that the relative order between these estimates remains consistent after a substantial number of pulls. 


\section{Stable Assignment}\label{sec:stable}
In this section we apply our dueling ULCB framework to address more general problems of stability. 
The primary distinction in the stable matching problem, compared to the fair allocation discussed in the previous section, lies in the absence of goods. Instead, agents are paired together, and it becomes essential to account for the potential deviation of a subset of agents from the centralized solution.

We first present a framework that closely aligns with the one discussed in the preceding section, yet it is sufficiently versatile to encompass various stable matching models, such as college admissions \citep{gale1962college, roth2008deferred}, ride-sharing \citep{lokhandwala2018dynamic,shi2023multiagent}, and matching with couples \citep{roth1984evolution,nguyen2018near}.

\subsection{Model}
We continue to use the notation outlined in previous sections, where $\mathcal{N}$ denotes the set of goods and $\mathcal{K}$ denotes the set of agents. Each agent $j$ has  a set of accessible bundles from $\mathcal{K}$, denoted by $\mathcal{A}_j$, along with a reward function given as $r^{j}: \mathbb{R}^N \times \mathcal{A}_j\to \mathbb{R}$ for $j=1,2,\cdots,K$.
We further define the desired set of \emph{feasible} assignments as 
\[\mathcal M^* \subseteq \{\phi: \mathcal{K}\rightarrow \cup \mathcal{A}_j \mid \phi(j)\in \mathcal{A}_j \}.\]
$\mathcal{M}^*$ represents a \emph{subset} of assignments with unspecified feasibility constraints, tailored to the specific problem. Allowing $\mathcal{M}^*$ to be arbitrary allows us to flexibly model a wide range of settings.


For example, in a marriage  model (dating market), the set of agents consists of a union of two sets: ${\mathcal{K}}=\mathcal{G}_1 \cup \mathcal{G}_2$
where $\mathcal{G}_1$ (resp. $\mathcal{G}_2$) is the set of men (resp. women) who are to be matched. The set of ``goods'' corresponds to  all possible pairs of men and women to be ${\mathcal{N}}=\{(m,w), (w,m):\,  m \in \mathcal{G}_1\text{ and }w \in \mathcal{G}_2\}$.\footnote{We can allow for  agents to stay single by adding $(m,\emptyset)$ and $(w,\emptyset)$ to $\mathcal{N}$.} Each agent is interested to consume a single good. For an agent $j\in \mathcal{K}$, the set of accessible goods is $\mathcal{A}_j:=\{(j,j')\mid j' \text{ is in the different gender group}\}$.  A matching of men and women is a map $\phi : {\mathcal{K}}\rightarrow {\mathcal{N}}$ such that if $\phi(m)=(m,w)$ for $m \in \mathcal{G}_1$ then $\phi(w)=(w,m)$ for that particular $w \in \mathcal{G}_2$.  Therefore, with this additional constraint, we define the  set of all matching as: 
\[\mathcal{M}^*_{\text{matching}}=\left\{ \phi : {\mathcal{K}}\rightarrow {\mathcal{N}}\mid \text{if }\phi(a)=(a,b) \, \text{then } \phi(b)=(b,a)\right\}.\]
Applying similar logic, the model can accommodate other matching problem variants. More specifically, the fundamental matching elements need not be confined to simple man-women matches; they can extend to scenarios such as matching couples (where two students are paired with two hospitals) or many-to-one matching (where a group of students is matched to a single school).

Next we consider  stability constraints. For example, in the marriage  model, a matching is stable if  there is no $(m,w)$ who are not matched, but prefer to be together. To formulate this constraint in our general framework, we consider for each  matching $\phi\in \mathcal M^*$, and a group of agents $L\subseteq \mathcal{K}$ there is a set of ``deviating'' assignment depending on $L,\phi$, denoted as $\mathcal M^*|_{\phi,L}$, where
$$
\mathcal M^*|_{\phi,L} \subseteq  \{\phi': L \rightarrow \cup_{j\in L} \mathcal{A}_j \mid \phi'(j)\in \mathcal{A}_j \}.
$$
For instance, in a marriage model, given a matching function $\phi$ and two agents of different genders $L=\{m,w\}$ such that $\phi(m)\neq w$, the set of ``deviating'' assignments consists of a single element: $\phi'(m)=(m,w); \phi'(w)=(w,m)$. In alternative matching models, like matching with couples, a more comprehensive set of deviating agents needs to be considered. This set could include, for example, 2 students and 2 hospitals.

We now define the quality of the possible deviation which measures how much more reward an agent gets under the current matching compared with the deviation. Thus, the true benefit for an agent to stay in the current assignment given the deviation is computed as 
\[g^j(\bm{\mu};\phi\rightarrow \phi'):=r^j(\bm \mu,\phi(j))-r^j(\bm \mu,\phi'(j))\]
when the base rewards are $\bm\mu$ with the original matching being $\phi$ and the deviation being $\phi'$.

If agent $j$  experiences a negative benefit, it implies he is motivated to deviate from the current assignment $\phi$. To measure the collective incentive of the entire agent group $L$ to stick with the current matching rather than deviating to $\phi'$, we calculate:

$$g^L(\bm\mu;\phi\rightarrow \phi') := \max_{j \in L} \bigl\{g^j(\bm{\mu};\phi\rightarrow \phi')\bigr\}.
$$
A negative value for $g^L(\bm\mu;\phi\rightarrow \phi')$  indicates that the group $L$ has a collective incentive to deviate and a positive value indicates that there exits at least one agent who has no incentive to deviate.

In the context of models such as the marriage model or matching with couples, we impose a restriction allowing only a limited number of agents to form a deviate coalition. Consequently, we assume that the size of the set $L$, denoted by $|L|$, is constrained such that $|L| \leq \kappa$, where $\kappa$ is a small constant. For example in  pairwise stable setups, $\kappa$ is typically set to 2, while in the case of matching for couples, $\kappa$ is set to 4.

Consider the following set that plays a key role in our problem:
\[
\mathbb{F}(\bm\mu,\theta) = \left\{\phi \in \mathcal{M}^*\;\middle|\; 
g^L(\bm \mu,\phi\rightarrow\phi')\ge \theta\text{ for all } L\subseteq \mathcal{K}, |L|\le \kappa, \phi'\in \mathcal{M^*}|_{L,\phi}
\right\}.
\]
The set $\mathbb{F}(\bm\mu,\theta)$ represents all matchings wherein any group of agents
with at most $\kappa$ members has an incentive of at least $\theta$ to resist changing their current assignment. We refer to this matching set as \emph{$\theta$-stable} and any element $\phi$ within it is termed a $\theta$-stable matching. Additionally, when $\theta=0$, we colloquially refer to the set as stable and its elements as stable matchings.


\subsection{Learning Task and Regret Measures}
Our objective is to determine whether the set of of stable matching $\mathbb{F}(\bm{\mu},0)$ is empty or not, and to select a stable matching if the set is non-empty without knowing the true $\bm{\mu}$. 
For this purpose, we pose this question as a statistical hypothesis testing problem. Given \emph{any} $\eta>0$, we formulate the null and the alternate hypothesis as follows
$$    
H_0: \mathbb{F}(\bm{\mu},0)=\emptyset;~~
H_a: \mathbb{F}(\bm{\mu},\eta) \ne \emptyset.
$$
The null hypothesis posits the absence of any stable matching, whereas the alternative hypothesis suggests the presence of a strong stable matching, wherein every deviation involves at least one agent with a strict preference to remain.

To study this hypothesis we consider this problem in an online setting where we aim to learn the stability set by gathering feedback from agents. Like in previous section, we are not allowed to collect feedback regarding the entire matching, but are restricted to at most $\kappa$ agents in each period.   In each period, there is an identical group of agents with unknown characteristics $\bm{\mu}$. Our algorithm tests the hypothesis using noisy estimates of $\bm{\mu}$ through an online decision rule $\delta_t$, which accepts or rejects the null hypothesis at time instance $t$. Additionally, we aim to output a matching believed to be stable when the alternate hypothesis is true.


Since we are analyzing a hypothesis testing problem, there are two errors- the Type I and the Type II errors that need to be taken into consideration. Thus, we naturally have three different regrets that measure the goodness of our decision rule and the quality of our solution. If the decision rule is such that we reject the null hypothesis if $\delta_t=1$ and do not reject it when $\delta_t=0$ then the three regret measures are as follows:
\begin{equation}\begin{aligned}\label{null:regret}
\textrm{Type I error:}~~R^{H_0}_T =\mathbb{E}_{H_0} \left[\sum_{t=1}^{T}\mathbf{1}\left\{ \delta_t=1\right\}\right], ~~~
\textrm{Type II error:}~~R^{H_a}_T =\mathbb{E}_{H_a} \left[\sum_{t=1}^{T}\mathbf{1}\left\{ \delta_t=0\right\}\right],
\end{aligned}\end{equation}
and the regret associated with not identifying a feasible solution when it exists:
\begin{equation}\begin{aligned}\label{regret:solution}
R_T=\mathbb{E}_{H_a}\left[\sum_{t=1}^{T}\mathbf{1}\left\{\phi_t \not \in \mathbb{F}(\bm{\mu},\eta)\right\}\right].
\end{aligned}\end{equation}
The first two regrets measure the rate of the Type I and Type II errors and the third one measures the rate at which we identify a solution in the stable set. We shall develop an algorithm that identifies the correct decision and solution with sub-linear regret.

\subsection{Algorithm}
Note that to analyze this setting, we use the idea of dueling UCB and LCB in previous sections. For any $\phi$, $L \subseteq \mathcal{K}$, with $\phi' \in \mathcal{M}^*|_{\phi,L}$, we define the upper estimate and the lower estimate of $g^L(\bm \mu, \phi \rightarrow \phi')$ over $[\bar{\bm{\nu}},\ub{\bm{\nu}}]$ as 
\[\textrm{upper}: g^L(\bar{\bm{\nu}}(t);\underline{\bm{\nu}}(t);\phi\rightarrow \phi')= \max_{j \in L} r^j(\bar{\bm{\nu}}(t),\phi(j))-r^j(\underline{\bm{\nu}}(t),\phi'(j))\]
and 
\[\textrm{lower}: g^L(\underline{\bm{\nu}}(t); \bar{\bm{\nu}}(t); \phi\rightarrow \phi')= \max_{j \in L} r^j(\underline{\bm{\nu}}(t),\phi(j))-r^j(\bar{\bm{\nu}}(t),\phi'(j)), \]
where $\bar{\bm{\nu}}(t)$ and $\underline{\bm{\nu}}(t)$ are the UCB and LCB estimates of the true base rewards respectively at time $t$.

The decision function for the hypothesis is given as, 
\begin{equation*}\begin{aligned}
\delta^{\epsilon}_t&=1 \ \text{if there exists } \phi \text{ such that } \  g^L(\underline{\bm{\nu}}(t); \bar{\bm{\nu}}(t); \phi\rightarrow \phi') \ge  \epsilon, \ \,\forall \, L \subseteq \mathcal{K} \text{ and } \phi' \in \mathcal{M}^*|_{\phi,L}\\
&=0 \ \text{otherwise}.
\end{aligned}\end{equation*}
Note that this decision rule is the same as the decision rule $\delta_t$ while defining the regret. We write it as $\delta^{\epsilon}_t$ to emphasize its dependence on $\epsilon$, which is a relaxation parameter for testing the hypothesis.

For a given $\phi$,  if $g^L(\underline{\bm{\nu}}(t); \bar{\bm{\nu}}(t); \phi\rightarrow \phi')$ is at least $\epsilon$ for all  $L \subseteq \mathcal{K}$ and  $\phi' \in \mathcal{M}^*|_{\phi,L}$, we say that $\phi$ is estimated to be $\epsilon$-stable over \emph{all} reasonable values of $\bm\mu$. On the other hand, if the upper-estimates $g^L(\bar{\bm{\nu}}(t);\underline{\bm{\nu}}(t); \phi\rightarrow \phi')$ are at least $\epsilon$ for $L \subseteq \mathcal{K}$ and  $\phi' \in \mathcal{M}^*|_{\phi,L}$, we say that $\phi$ could be $\epsilon$-stable over \emph{some} reasonable value of $\bm\mu$. Now, we consider an oracle $\mathcal{O}_F$ which, given  $\bar{\bm{\nu}},\underline{\bm{\nu}}$ each in  $\mathbb{R}^{N}$ and an $\eta\in \mathbb{R}$, solves the static stability problem, and if $\delta^\epsilon_t = 1$ returns a corresponding $\phi \in \mathcal{M}^*$. Recall that, by definition, this $\phi$ is estimated to be $\epsilon$-stable over \emph{all} reasonable values of $\bm\mu$. On the other hand, if  $\delta^{\epsilon}_t=0$ no such $\phi$ exists. Instead, the oracle checks if there is a $\phi \in \mathcal{M}^*$ that could be $\eta$-stable with $g^{L}(\bar{\bm\nu}(t);\underline{\bm\nu}(t);\phi\rightarrow \phi')\ge \eta$ for all $L\subseteq \mathcal{K},\,  \phi' \in \mathcal{M}^*|_{L,\phi}$ which we define as a solution being $\eta$-stable for some value of $\bm\mu$. If so, the oracle returns this $\phi$. Otherwise, it returns an arbitrary $\phi\in M^*$. Regardless of whether our estimates find that $\phi$  could be $\eta$-stable or not, since $\delta^\epsilon_t = 0$, $\phi$ is not estimated to $\epsilon$-stable over all reasonable values of $\bm\mu$ and, therefore, there exists an $L \subseteq \mathcal{K}$ and a $\phi'\in M^*|_{L,\phi}$ such that the lower estimate $g^L(\underline{\bm{\nu}}(t); \bar{\bm{\nu}}(t); \phi\rightarrow \phi')<\epsilon$. The oracle returns this set which provides evidence of why $\phi$ was not estimated to be $\epsilon$-stable.  
We present our algorithm in Algorithm \ref{algo:gen:problem}. Note that the selection of $\epsilon$ is fundamental to the algorithm and hence practitioners are advised to use multiple values to see which works best in practise.

\begin{algorithm}
	\textbf{Input}: $\mathcal{N}, \mathcal{K}, \sigma, \alpha, \eta$.\\
	\textbf{for}: $t=1,2,\cdots, N$;\\
	Pull each arm.\\
	\textbf{Update}: $T_i(t), \hat{\mu}_i(t),\bar{\nu}_i(t), \underline{\nu}_i(t)$.\\
	\textbf{for} $t=N+1,\cdots,T$ \textbf{do}:\\
	UCB: $\bar{\nu}_i(t)=\hat{\mu}_i(t)+\sqrt{\frac{1}{T_{i}(t-1)}\,2\sigma^2 \log t^{\alpha}}$\\
	and \\
	LCB: $
	\underline{\nu}_i(t)=\hat{\mu}_i(t)-\sqrt{\frac{1}{T_{i}(t-1)}\,2\sigma^2 \log t^{\alpha}}$\\
	$(\delta^{\epsilon}_t, \phi_t, L_t,\phi'_t)=\mathcal{O}_F(\eta,\underline{\nu}_1(t),\underline{\nu}_2(t),\cdots, \underline{\nu}_N(t),\bar{\nu}_1(t),\bar{\nu}_2(t),\cdots, \bar{\nu}_N(t))$ \\ 
	\textbf{Output}\,  $\phi_t$ \textbf{and Pull} $\phi_t(j), \phi'_t(j)$ for $t=N+1,2,\cdots,T$, $j \in L_t$.	
	\caption{Feasibility ULCB}
	\label{algo:gen:problem}
\end{algorithm} 



Next we study the theoretical properties of our proposed feasibility ULCB algorithm by deriving the upper bounds of the type-I and type-II errors in $(\ref{null:regret})$ and the regret in $(\ref{regret:solution})$. We first list the main assumption for this section. Define, for each $\phi$, the set $\mathcal{B}_{\phi}=\bigl\{(L,\phi'): g^{L}(\bm\mu, \phi \rightarrow \phi') < \eta\bigr\}$. Note that $\mathcal{B}_{\phi}$ may be empty for a particular $\phi$. 
\begin{assumption}\label{assm:feas:sep}
	There exists an $\eta>0$ and $\Delta_{\mathcal{M}^*,q}>0$ such that 
	\[\inf_{\phi\in \mathcal{M}^*\backslash \mathbb{F}(\bm{\mu},\eta), \, (L,\phi') \in \mathcal{B}_{\phi}}\left(\eta- g^{L}(\bm\mu, \phi \rightarrow \phi')) \right)\ge \Delta_{\mathcal{M}^*,q}>0.\] Further, we assume $\eta-\Delta_{\mathcal{M}^*,q} <\epsilon <\eta$.
\end{assumption}
Assumption~\ref{assm:feas:sep} implies that there exists a gap between the stable set and its complement. The reason for this is that in the stable set $g^{L}(\bm\mu, \phi \rightarrow \phi') \ge \eta$ for all $L,\phi'$ and on the complement of a stable set, we have $g^{L}(\bm\mu, \phi \rightarrow \phi') < \eta-\Delta_{\mathcal{M},q}$. This is akin to canonical hypothesis testing scenarios where there needs to be difference between the signals for a true discovery to be identified. As in previous sections, this may again be viewed as an identifiability condition for the stable set. Note that if $H_0$ holds then 
$\sup_{ \phi \in \mathcal{M}^*} \inf_{(L,\phi') \in \mathcal{B}_{\phi}} g^{L}(\bm\mu, \phi \rightarrow \phi'))<0$. 

We are now ready to state our main results, of which the proof is provided in the Appendix. 
\begin{proposition}\label{prop:null:hyp}
	Assume that Assumption~\ref{assm:reward:Lipschitz} holds. The expected number of type-I errors made by Algorithm \ref{algo:gen:problem} satisfies 
	$$R^{H_0}_T \le \frac{2\, N\, \left(\alpha-1\right)}{\alpha -2}.$$ 
\end{proposition}
Proposition~\ref{prop:null:hyp} implies that our algorithm is expected to make no more than a constant number of errors in identifying the null hypothesis regardless of the time horizon as long as $T>N$. 
\begin{theorem}\label{thm:main:feas}
	Let Assumptions~\ref{assm:reward:Lipschitz} and~\ref{assm:feas:sep} hold. Then, the type-II error of our algorithm satisfies
	\begin{equation*}\begin{aligned}
	R^{H_a}_T \le C_1^*(\kappa, \sigma^2, r,\alpha)\left(\Delta^{-2}_{\mathcal{M},q}+2\, (\eta-\epsilon)^{-2}\right) N^3\, \log T  +N\,C^*_2(\alpha) 
	\end{aligned}\end{equation*}
	where $C_1^*(\kappa, \sigma^2, r, \alpha)$, $C^*_2(\alpha)$ are quantities independent of $N,T$.
\end{theorem}

Theorem~\ref{thm:main:feas} implies that our algorithm expects to make $O(N^3 \log T)$ type-II errors within a total time horizon of $T$, thereby accurately identifying feasibility in most iterations.
\begin{theorem}\label{Prop:feasibility}
	Under Assumptions~\ref{assm:reward:Lipschitz} and~\ref{assm:feas:sep}, we have
	\begin{equation*}\begin{aligned}
	R_T \le \tilde{C}_1^*(\alpha,r) \left(\Delta^{-2}_{\mathcal{M},q}+(\eta-\epsilon)^{-2}\right) N^3\,\log T+N\, \tilde{C}_2^*(\alpha)   
	\end{aligned}\end{equation*}
	where $\tilde{C}_1^*(\alpha,r)$ and $\tilde{C}_2^*(\alpha)$ are independent of $N$ and $T$.
\end{theorem}

Theorem~\ref{Prop:feasibility} implies that our algorithm identifies a stable solution in most iterations, making at most $O(N^3\, \log T)$ errors within a total time horizon of $T$. 

Taken together, these results show that, even though the feedback we collect is restricted to sets of agents of size $\kappa$, each of which participates by assessing their local stability, our algorithm correctly detects the existence/non-existence of a globally-stable solution and returns stable assignments in most time-periods, provided $\eta$-stable solutions exist. Note that the smaller the value of $\Delta_{\mathcal{M},q}$ is the harder the problem is as not only does the signal become hard to detect but also the range of values for $\epsilon$ decreases. 

\section{Simulations}\label{sec:sim}

\begin{figure}[htb]
	\centering
	\includegraphics[height=5cm,width=7cm]{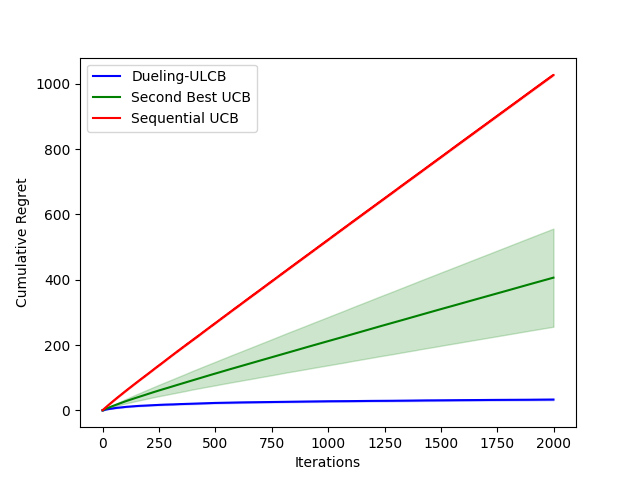}
	\caption{Comparison between the proposed Dueling ULCB and the two benchmark methods. 
	}
	\label{fig:toy:example}
\end{figure}

In this section we present simulation experiments which bolster our theoretical understanding of the problem. We first present the example considered in Section~\ref{sec:top-K} with $N=3$ goods and $K=2$ agents. The true arm rewards are $\mu_1=1,\mu_2=2,\mu_3=3$ respectively for the three arms and the observed rewards at time $t$ are $X_i(t)\sim N(\mu_i,1)$ where $i=1,2,3$. Our objective is to select the second-largest arm, i.e., arm 2. Figure~\ref{fig:toy:example} illustrates a comparison of cumulative regrets between our proposed Dueling ULCB and two benchmark methods. The first benchmark selects the arm with the second-best UCB and is referred to as ``Second Best UCB". The second benchmark method sequentially selects the arm with the best UCB and subsequently the second best UCB. This refers to the sequential version of existing top-k-arm selection algorithm~\citep{heckel2018approximate}. We refer to this as ``Sequential UCB". Evidently, the benchmark methods fail to converge to the correct solution, while our approach demonstrates a substantial improvement in regret. This indicates that in our considered limited feedback setting, only using UCB is insufficient and it is critical to incorporate both UCB and LCB for the active learning.
\vspace{-1.5em}
\FloatBarrier

In fact, for any algorithm with the property that an incorrect arm is chosen at each instance with some fixed probability, i.e., $\mathbb{P}(\text{choosing some incorrect arm at time }$t$)\ge c$ where $c>0$ is some constant, it would incur a linear regret. This is demonstrated in the Sequential UCB algorithm with $3$ arms with $K=2$ in Figure~\ref{fig:toy:example}. Therefore, for an algorithm to exhibit sub-linear regret, the probability of selection of incorrect arms should decrease to $0$ when time horizon increases.

\begin{figure}[h!]
	\centering  
	\begin{subfigure}[t]{0.48\textwidth} 
		\centering \includegraphics[width=6cm]{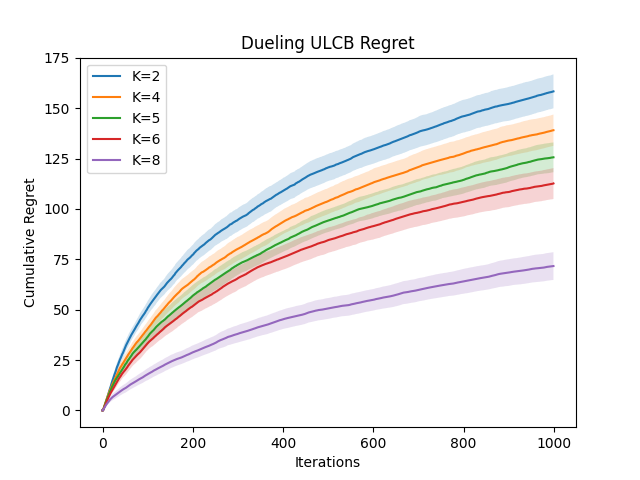}
		\caption{Varying $K$.}
		\label{fig:ulcb_regret:varyingK}
	\end{subfigure}
	\hfill
	\begin{subfigure}[t]{0.48\textwidth}
		\centering
		\includegraphics[width=6 cm]{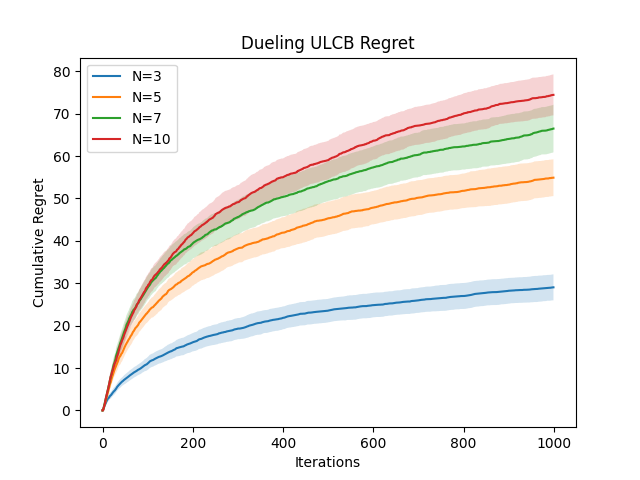}
		\caption{Varying $N$.}	\label{fig:ulcb_regret:varyingN}	
	\end{subfigure}
	\caption{Cumulative regrets of our Dueling ULCB algorithm with varying $K$ and varying $N$.}
	\label{fig:ulcb_regret}
\end{figure}

Next we study the effects of $K$ and $N$ on the cumulative regret of the Dueling ULCB algorithm. We consider a MAB setting with the number of goods $N=10$ and the number of agents $K=2,3,5,7,8$. Further, we generate the reward for the $i$-th agent as $X_i(t)\sim N(\mu_i,1)$, where $\mu_i=i$, $i=1,2,\cdots,N$. The arm differences are equal to $1$ in this case. In Figure~\ref{fig:ulcb_regret} we see that the cumulative regret is sub-linear with respect to the number of iterations. Further, in Figure~\ref{fig:ulcb_regret:varyingK} the cumulative regret of the algorithm shows that as $K$ increases the regret decreases.  Moreover, Figure~\ref{fig:ulcb_regret:varyingN} shows that the regret is increasing in $N$, which is expected as more exploration is needed for larger $N$. This result is also consistent with our theoretical findings.

\begin{figure}[!ht]
	\centering  
	\begin{subfigure}[t]{0.48\textwidth} 
		\centering \includegraphics[width=6cm]{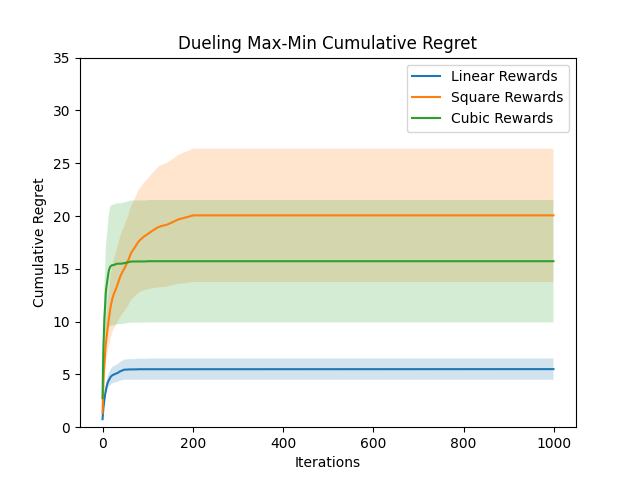}
		\caption{Same reward function for both agents.}
		\label{fig:Cumulative_Regret_vs_Rewards}
	\end{subfigure}
	\hfill
	\begin{subfigure}[t]{0.48\textwidth}
		\centering
		\includegraphics[width=6cm]{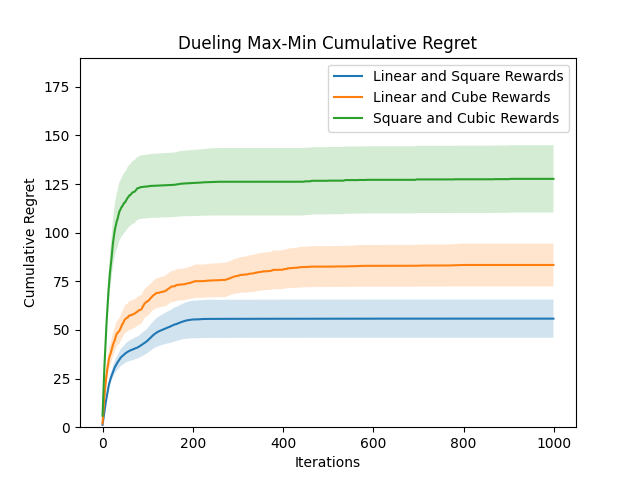}
		\caption{Different reward functions for both agents.}	\label{fig:Cumulative_Regret_vs_Mixture_Rewards}	
	\end{subfigure}
	\caption{Cumulative regrets of the Dueling Max-Min ULCB algorithm with varying reward functions for both agents.}
	\label{fig:regret_vs_rewards}
\end{figure}

Following this, we run experiments for the case where we have base arms $1,2,3,4$ and the number of agents is $2$. The base arm rewards are again $N(\mu_i,1)$, with $\mu_i =i$, like in the last experiment. We use three different reward functions and their combinations for the purpose of our simulations which are $r(\bm{\mu};S)=\sum_{i\in S} \mu_i, \ r(\bm{\mu};S)=\sum_{i\in S} \mu^3_i$ and $r(\bm{\mu};S)=\sum_{i\in S} (\mu_i\vee 0)^2$ respectively. In Figure~\ref{fig:Cumulative_Regret_vs_Rewards} we present the cumulative regret of the Dueling Max-Min ULCB algorithm with same rewards for all agents. We vary the reward function between the three reward functions as stated above. In Figure~\ref{fig:Cumulative_Regret_vs_Mixture_Rewards} we present the cumulative regret of the Dueling Max-Min ULCB algorithm where the agents share different rewards. We vary the reward function between the three reward functions as stated above. As shown in Figure \ref{fig:regret_vs_rewards}, among all the settings of the reward function, our Dueling Max-Min ULCB Algorithm achieves a clear sub-linear pattern.

\begin{figure}[!ht]
	\centering  
	\begin{subfigure}[t]{0.48\textwidth} 
		\centering \includegraphics[width=6cm]{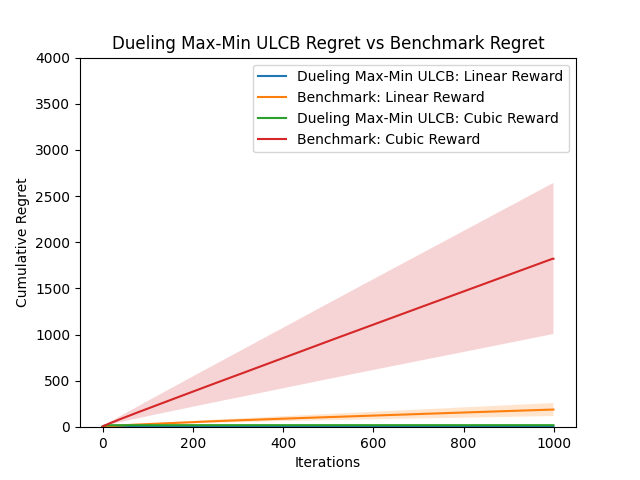}
		\caption{Same reward function for both agents.}
		\label{fig:Cumulative_Regret_vs_Benchmark_Multiple_Rewards}
	\end{subfigure}
	\hfill
	\begin{subfigure}[t]{0.48\textwidth}
		\centering
		\includegraphics[width=6 cm]{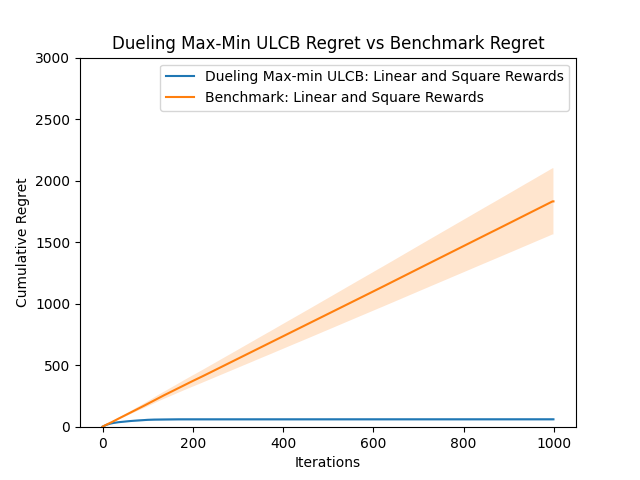}
		\caption{Different reward functions for both agents.}	\label{fig:Cumulative_Regret_vs_Benchmark_Mixed_Rewards}	
	\end{subfigure}
	\caption{Cumulative regrets of Dueling Max-Min ULCB vs Benchmark with varying reward functions for both agents.}
	\label{fig:reward:mix}
\end{figure}

Finally we compare the Dueling Max-Min ULCB algorithm to a benchmark algorithm which excludes the LCB step from the algorithm and only solves the Max-Min problem using the UCB values. In Figure~\ref{fig:Cumulative_Regret_vs_Benchmark_Multiple_Rewards} we do this comparison with the agents sharing the same reward and in Figure~\ref{fig:Cumulative_Regret_vs_Benchmark_Mixed_Rewards}, we perform the same experiments with the agents having different rewards. Across all scenarios, the benchmark method exhibits a distinct linear regret, suggesting that relying solely on UCB values is inadequate for resolving our max-min allocation problem. Conversely, our Dueling Max-Min ULCB algorithm demonstrates a significant enhancement in regret.

\FloatBarrier

\section{Conclusion}
Our paper proposes a framework for learning with limited feedback. The constraint on feedback arises from practical considerations, as obtaining feedback, while providing crucial information for online decision-making, can incur significant costs. We apply this framework to various allocation problems, aiming to achieve fairness or stability. Our results uncover important structures indicating that not all information about the allocation is necessary to achieve the desired outcome. Our algorithm has practical applications in real-world scenarios, including online dating markets, job matching, and food allocation.


Future work involves addressing problems with more complex objectives and constraints, where it is possible that limiting feedback to one or a small constant number of agents  may not be sufficient to achieve sub-linear regret. An immediate example of this is a constrained optimization problem or an LP with noisy coefficients. Investigating the trade off between the regret and the size of feedback is interesting and can provide insights into the fundamental structure of the problem. Moreover, our paper currently focuses on the sample complexity of the considered problem and does not discuss the computational issues of the offline oracles. It is an interesting to study the trade-off of computational and sample complexity in practical polynomial-time approximate algorithms. 

\bibliographystyle{plain}
\bibliography{ec_bib}

\begin{thebibliography}{10}

\bibitem{aleksandrov2015online}
Martin Aleksandrov, Haris Aziz, Serge Gaspers, and Toby Walsh.
\newblock Online fair division: Analysing a food bank problem.
\newblock {\em arXiv preprint arXiv:1502.07571}, 2015.

\bibitem{aleksandrov2017expected}
Martin Aleksandrov and Toby Walsh.
\newblock Expected outcomes and manipulations in online fair division.
\newblock In {\em KI 2017: Advances in Artificial Intelligence: 40th Annual German Conference on AI, Dortmund, Germany, September 25--29, 2017, Proceedings 40}, pages 29--43. Springer, 2017.

\bibitem{aleksandrov2017most}
Martin Aleksandrov and Toby Walsh.
\newblock Most competitive mechanisms in online fair division.
\newblock In {\em KI 2017: Advances in Artificial Intelligence: 40th Annual German Conference on AI, Dortmund, Germany, September 25--29, 2017, Proceedings 40}, pages 44--57. Springer, 2017.

\bibitem{aleksandrov2017pure}
Martin Aleksandrov and Toby Walsh.
\newblock Pure nash equilibria in online fair division.
\newblock In {\em IJCAI}, pages 42--48, 2017.

\bibitem{aleksandrov2019strategy}
Martin Aleksandrov and Toby Walsh.
\newblock Strategy-proofness, envy-freeness and pareto efficiency in online fair division with additive utilities.
\newblock In {\em PRICAI 2019: Trends in Artificial Intelligence: 16th Pacific Rim International Conference on Artificial Intelligence, Cuvu, Yanuca Island, Fiji, August 26--30, 2019, Proceedings, Part I 16}, pages 527--541. Springer, 2019.

\bibitem{aleksandrov2020online}
Martin Aleksandrov and Toby Walsh.
\newblock Online fair division: A survey.
\newblock In {\em Proceedings of the AAAI Conference on Artificial Intelligence}, volume~34, pages 13557--13562, 2020.

\bibitem{banerjee2022proportionally}
Siddhartha Banerjee, Vasilis Gkatzelis, Safwan Hossain, Billy Jin, Evi Micha, and Nisarg Shah.
\newblock Proportionally fair online allocation of public goods with predictions.
\newblock {\em arXiv preprint arXiv:2209.15305}, 2022.

\bibitem{banerjee2023online}
Siddhartha Banerjee, Chamsi Hssaine, and Sean~R Sinclair.
\newblock Online fair allocation of perishable resources.
\newblock {\em ACM SIGMETRICS Performance Evaluation Review}, 51(1):55--56, 2023.

\bibitem{benade2022dynamic}
Gerdus Benad{\`e}, Daniel Halpern, and Alexandros Psomas.
\newblock Dynamic fair division with partial information.
\newblock {\em Advances in neural information processing systems}, 35:3703--3715, 2022.

\bibitem{benade2023fair}
Gerdus Benad{\`e}, Aleksandr~M Kazachkov, Ariel~D Procaccia, Alexandros Psomas, and David Zeng.
\newblock Fair and efficient online allocations.
\newblock {\em Operations Research}, 2023.

\bibitem{benade2018make}
Gerdus Benade, Aleksandr~M Kazachkov, Ariel~D Procaccia, and Christos-Alexandros Psomas.
\newblock How to make envy vanish over time.
\newblock In {\em Proceedings of the 2018 ACM Conference on Economics and Computation}, pages 593--610, 2018.

\bibitem{bistritz2020my}
Ilai Bistritz, Tavor Baharav, Amir Leshem, and Nicholas Bambos.
\newblock My fair bandit: Distributed learning of max-min fairness with multi-player bandits.
\newblock In {\em International Conference on Machine Learning}, pages 930--940. PMLR, 2020.

\bibitem{NIPS2015_ab233b68}
Wei Cao, Jian Li, Yufei Tao, and Zhize Li.
\newblock On top-k selection in multi-armed bandits and hidden bipartite graphs.
\newblock In C.~Cortes, N.~Lawrence, D.~Lee, M.~Sugiyama, and R.~Garnett, editors, {\em Advances in Neural Information Processing Systems}, volume~28. Curran Associates, Inc., 2015.

\bibitem{cen2022regret}
Sarah~H Cen and Devavrat Shah.
\newblock Regret, stability \& fairness in matching markets with bandit learners.
\newblock In {\em International Conference on Artificial Intelligence and Statistics}, pages 8938--8968. PMLR, 2022.

\bibitem{cesa2012combinatorial}
Nicolo Cesa-Bianchi and G{\'a}bor Lugosi.
\newblock Combinatorial bandits.
\newblock {\em Journal of Computer and System Sciences}, 78(5):1404--1422, 2012.

\bibitem{chen2013combinatorial}
Wei Chen, Yajun Wang, and Yang Yuan.
\newblock Combinatorial multi-armed bandit: General framework and applications.
\newblock In {\em International conference on machine learning}, pages 151--159. PMLR, 2013.

\bibitem{fikioris2023online}
Giannis Fikioris, Siddhartha Banerjee, and {\'E}va Tardos.
\newblock Online resource sharing via dynamic max-min fairness: Efficiency, robustness and non-stationarity.
\newblock {\em arXiv preprint arXiv:2310.08881}, 2023.

\bibitem{gale1962college}
David Gale and Lloyd~S Shapley.
\newblock College admissions and the stability of marriage.
\newblock {\em The American Mathematical Monthly}, 69(1):9--15, 1962.

\bibitem{pmlr-v151-garcelon22b}
Evrard Garcelon, Vashist Avadhanula, Alessandro Lazaric, and Matteo Pirotta.
\newblock Top k ranking for multi-armed bandit with noisy evaluations.
\newblock In Gustau Camps-Valls, Francisco J.~R. Ruiz, and Isabel Valera, editors, {\em Proceedings of The 25th International Conference on Artificial Intelligence and Statistics}, volume 151 of {\em Proceedings of Machine Learning Research}, pages 6242--6269. PMLR, 28--30 Mar 2022.

\bibitem{golovin2005max}
Daniel Golovin.
\newblock {\em Max-min fair allocation of indivisible goods}.
\newblock School of Computer Science, Carnegie Mellon University, 2005.

\bibitem{heckel2018approximate}
Reinhard Heckel, Max Simchowitz, Kannan Ramchandran, and Martin Wainwright.
\newblock Approximate ranking from pairwise comparisons.
\newblock In {\em International Conference on Artificial Intelligence and Statistics}, pages 1057--1066. PMLR, 2018.

\bibitem{jagadeesan2023learning}
Meena Jagadeesan, Alexander Wei, Yixin Wang, Michael~I Jordan, and Jacob Steinhardt.
\newblock Learning equilibria in matching markets with bandit feedback.
\newblock {\em Journal of the ACM}, 70(3):1--46, 2023.

\bibitem{kalyanakrishnan2012pac}
Shivaram Kalyanakrishnan, Ambuj Tewari, Peter Auer, and Peter Stone.
\newblock Pac subset selection in stochastic multi-armed bandits.
\newblock In {\em ICML}, volume~12, pages 655--662, 2012.

\bibitem{kawase2022online}
Yasushi Kawase and Hanna Sumita.
\newblock Online max-min fair allocation.
\newblock In {\em International Symposium on Algorithmic Game Theory}, pages 526--543. Springer, 2022.

\bibitem{lattimore2020bandit}
Tor Lattimore and Csaba Szepesv{\'a}ri.
\newblock {\em Bandit algorithms}.
\newblock Cambridge University Press, 2020.

\bibitem{lattimore2020learning}
Tor Lattimore, Csaba Szepesvari, and Gellert Weisz.
\newblock Learning with good feature representations in bandits and in rl with a generative model.
\newblock In {\em International Conference on Machine Learning}, pages 5662--5670. PMLR, 2020.

\bibitem{leshem2024fair}
Amir Leshem.
\newblock Fair multi-agent bandits.
\newblock {\em arXiv preprint arXiv:2306.04498}, 2024.

\bibitem{li2023double}
Yuantong Li, Guang Cheng, and Xiaowu Dai.
\newblock Double matching under complementary preferences.
\newblock {\em arXiv preprint arXiv:2301.10230}, 2023.

\bibitem{li2023rate}
Yuantong Li, Chi-hua Wang, Guang Cheng, and Will~Wei Sun.
\newblock Rate-optimal contextual online matching bandit.
\newblock {\em arXiv preprint arXiv:2205.03699}, 2023.

\bibitem{liu2020competing}
Lydia~T Liu, Horia Mania, and Michael Jordan.
\newblock Competing bandits in matching markets.
\newblock In {\em International Conference on Artificial Intelligence and Statistics}, pages 1618--1628. PMLR, 2020.

\bibitem{locatelli2016optimal}
Andrea Locatelli, Maurilio Gutzeit, and Alexandra Carpentier.
\newblock An optimal algorithm for the thresholding bandit problem.
\newblock In {\em International Conference on Machine Learning}, pages 1690--1698. PMLR, 2016.

\bibitem{lokhandwala2018dynamic}
Mustafa Lokhandwala and Hua Cai.
\newblock Dynamic ride sharing using traditional taxis and shared autonomous taxis: A case study of nyc.
\newblock {\em Transportation Research Part C: Emerging Technologies}, 97:45--60, 2018.

\bibitem{markakis2011worst}
Evangelos Markakis and Christos-Alexandros Psomas.
\newblock On worst-case allocations in the presence of indivisible goods.
\newblock In {\em International Workshop on Internet and Network Economics}, pages 278--289. Springer, 2011.

\bibitem{min2022learn}
Yifei Min, Tianhao Wang, Ruitu Xu, Zhaoran Wang, Michael Jordan, and Zhuoran Yang.
\newblock Learn to match with no regret: Reinforcement learning in markov matching markets.
\newblock {\em Advances in Neural Information Processing Systems}, 35:19956--19970, 2022.

\bibitem{muthirayan2023competing}
Deepan Muthirayan, Chinmay Maheshwari, Pramod Khargonekar, and Shankar Sastry.
\newblock Competing bandits in time varying matching markets.
\newblock In {\em Learning for Dynamics and Control Conference}, pages 1020--1031. PMLR, 2023.

\bibitem{nguyen2018near}
Thanh Nguyen and Rakesh Vohra.
\newblock Near-feasible stable matchings with couples.
\newblock {\em American Economic Review}, 108(11):3154--3169, 2018.

\bibitem{roth1984evolution}
Alvin~E Roth.
\newblock The evolution of the labor market for medical interns and residents: a case study in game theory.
\newblock {\em Journal of political Economy}, 92(6):991--1016, 1984.

\bibitem{roth2008deferred}
Alvin~E Roth.
\newblock Deferred acceptance algorithms: History, theory, practice, and open questions.
\newblock {\em international Journal of game Theory}, 36(3):537--569, 2008.

\bibitem{JMLR:v18:16-206}
Nihar~B. Shah and Martin~J. Wainwright.
\newblock Simple, robust and optimal ranking from pairwise comparisons.
\newblock {\em Journal of Machine Learning Research}, 18(199):1--38, 2018.

\bibitem{shi2023multiagent}
Chengchun Shi, Runzhe Wan, Ge~Song, Shikai Luo, Hongtu Zhu, and Rui Song.
\newblock A multiagent reinforcement learning framework for off-policy evaluation in two-sided markets.
\newblock {\em The Annals of Applied Statistics}, 17(4):2701--2722, 2023.

\bibitem{yamada2023learning}
Hakuei Yamada, Junpei Komiyama, Kenshi Abe, and Atsushi Iwasaki.
\newblock Learning fair division from bandit feedback.
\newblock {\em arXiv preprint arXiv:2311.09068}, 2023.

\bibitem{zhang2021quantile}
Mengyan Zhang and Cheng~Soon Ong.
\newblock Quantile bandits for best arms identification.
\newblock In {\em International Conference on Machine Learning}, pages 12513--12523. PMLR, 2021.

\bibitem{zhou2022approximate}
Ruida Zhou and Chao Tian.
\newblock Approximate top-$ m $ arm identification with heterogeneous reward variances.
\newblock In {\em International Conference on Artificial Intelligence and Statistics}, pages 7483--7504. PMLR, 2022.

\end{thebibliography}

\appendix

	\begin{center}
		{\Large\bf Appendices} \\
		\medskip
		{\Large\bf ``Active Learning for Fair and Stable Online Allocations"}  \\
		\bigskip
	\end{center}
	In this file, we provide all detailed proofs in Sections~\ref{sec:proof-MAB}-\ref{sec:proof-general}. In Section~\ref{sec:proof-MAB}, we provide proofs for the unit demand case which is presented in Section~\ref{sec:MAB} in the main body. We also present some key results in this Section which shall be used through the rest of the Appendix. 
	
	


\section{Proofs for Section~\ref{sec:MAB}}\label{sec:proof-MAB}
We further present another Lemma which is fundamental for our main results.


\begin{lemma}\label{subgaus:lemma}
	Consider a MAB with $N$ arms, where the $i$-th arm has noisy rewards $X_i(t)=\mu_i+\epsilon_i(t)$ with $\epsilon_i(t)\sim \sigma^2$-subgaussian, $i=1,\ldots,N$. Further, denote by $\bar{\nu}_i(t),\underline{\nu}_i(t)$ the UCB and LCB of the reward estimate for the $i$-th arm at time $t\in \{1,2,\cdots,T\}$ defined in $(\ref{UCBdef})$ and $(\ref{LCBdef})$. Then for any $\hat{\Delta} \ge 2, \, \delta>0$, we have
	
	\begin{equation*}\begin{aligned}
	\mathbb{E}\left[\sum_{t=1}^{T}\mathbf{1}\left\{\exists \, i \in \mathcal{N}: 	\left(\left|\bar{\nu}_i(t)-\mu_{i}\right|\ge \delta \right)\cap\left(I_t=i\right)\cap \left(T_i(t-1)> C\, \log T\right) \right\}\right] \le  \frac{2\,N}{T^{\hat{\Delta}^2/2-2}}
	\end{aligned}\end{equation*}
	
	and 
	
	\begin{equation*}\begin{aligned}
	\mathbb{E}\left[\sum_{t=1}^{T}\mathbf{1}\left\{\exists \, i \in \mathcal{N}: 	\left(\left|\underline{\nu}_i(t)-\mu_{i}\right|\ge \delta \right)\cap\left(I_t=i\right)\cap \left(T_i(t-1)> C\, \log T\right) \right\}\right] \le  \frac{2\,N}{T^{\hat{\Delta}^2/2-2}}
	\end{aligned}\end{equation*}
	where $C=\left(\sqrt{2\alpha }+\hat{\Delta}\right)^2
	\sigma^2\, \delta^{-2}.$
\end{lemma}
\noindent\textbf{Proof of Lemma~\ref{subgaus:lemma}.} 
We begin by noting that 

\begin{equation*}\begin{aligned}
&\mathbb{E}\left[\sum_{t=1}^{T}\mathbf{1}\left\{\exists \, i \in \mathcal{N}: 	\left(\left|\bar{\nu}_i(t)-\mu_{i}\right|\ge \delta \right)\cap\left(I_t=i\right)\cap \left(T_i(t-1)> C\, \log T\right) \right\}\right]\\
&\le \mathbb{E}\left[\sum_{t=1}^{T}\mathbf{1}\left\{\exists \, i \in \mathcal{N}: 	\left(\left|\bar{\nu}_i(t)-\mu_{i}\right|\ge \delta\right)\cap \left(T_i(t-1)> C\, \log T\right)\right\}\right].
\end{aligned}\end{equation*}
Therefore, we have 

\begin{equation*}\begin{aligned}
&\mathbb{E}\left[\sum_{t=1}^{T}\mathbf{1}\left\{\exists \, i \in \mathcal{N}: 	\left(\left|\bar{\nu}_i(t)-\mu_{i}\right|\ge \delta \right)\cap\left(I_t=i\right)\cap \left(T_i(t-1)> C\, \log T\right) \right\}\right]\\
&\le \sum_{i\in \mathcal{N}}\sum_{t=1}^{T}\mathbb{P}\left\{	\left(\left|\bar{\nu}_i(t)-\mu_{i}\right|\ge \delta \right)\cap \left(T_i(t-1)> C\, \log T\right)\right\}\\
&\le \sum_{i\in \mathcal{N}}\sum_{t=1}^{T}\sum_{j=[C\log T]+1}^{T}\mathbb{P}\left[	\left(\left|\bar{\nu}_i(t)-\mu_{i}\right|\ge \delta \right)\cap \left(T_i(t-1)=j\right)\right]
\end{aligned}\end{equation*}

where the second step follows from the union bound and the third step follows from the values that $T_i(t-1)$ can take. Further, $[x]$ is the floor function denoting the greatest integer less than $x$. 
Now, for any $j=[C\log T]+1,\cdots, T$, we have

\begin{equation*}
\begin{aligned}
&\mathbb{P}\left\{\left(\bar{\nu}_i(t)-\mu_{i}\ge \delta \right)\cap \left(T_i(t-1)=j\right)\right\}\\
&\overset{(1)}{\le} \mathbb{P}\left\{\left(\frac{\sqrt{j}}{\sigma}\left(\hat{\mu}_i(t)-\mu_{i}\right)\ge \frac{\delta \sqrt{C\, \log T}}{\sigma}-\sqrt{2\alpha \log t} \right)\cap \left(T_i(t-1)=j\right)\right\}\\
&\overset{(2)}{\le} \exp\left\{-\frac{1}{2}\left(\frac{\delta\sqrt{C\, \log T}}{\sigma}-\sqrt{2\alpha \log t}\right)^2\right\}\\
&\overset{(3)}{\le} \exp\left\{-\frac{1}{2}\left(\frac{\delta \sqrt{C\, \log T}}{ \sigma}-\sqrt{2\alpha \log T}\right)^2\right\}\\
&\overset{(4)}{\le} \exp\left\{-\frac{\hat{\Delta}^2}{2}\log T\right\}\\
&\le \frac{1}{T^{\hat{\Delta}^2/2}}.
\end{aligned}
\end{equation*}

Note that $(1)$ follows using simple algebra and the fact that $j>C \log T$, $(2)$ follows as the arm rewards are subgaussian, $(3)$ follows as the expression is maximized for $t=T$ as the expression inside the square is positive and monotonically decreasing in $t$ with $T$ fixed (since the square term is minimized at $t=T$), 
and $(4)$ follows by the definition of $\hat{\Delta}$. 

Similarly, for the left tail, we have

\begin{equation*}
\begin{aligned}
&\mathbb{P}\left\{\left(\bar{\nu}_i(t)-\mu_{i}\le -\delta \right)\cap \left(T_i(t-1)=j\right)\right\}\\
&\overset{(5)}{\le} \mathbb{P}\left\{\left(\frac{\sqrt{j}}{\sigma}\left(\hat{\mu}_i(t)-\mu_{i}\right)\le -\frac{\delta \sqrt{C\, \log T}}{\sigma}-\sqrt{2\alpha \log t} \right)\cap \left(T_i(t-1)=j\right)\right\}\\
&\overset{(6)}{\le} \exp\left\{-\frac{1}{2}\left(\frac{\delta\sqrt{C\, \log T}}{\sigma}+\sqrt{2\alpha \log t}\right)^2\right\}\\
&\overset{(7)}{\le} \exp\left\{-\frac{1}{2}\left(\frac{\delta \sqrt{C\, \log T}}{ \sigma}-\sqrt{2\alpha \log t}\right)^2\right\}\\
&\le \frac{1}{T^{\hat{\Delta}^2/2}},
\end{aligned}
\end{equation*}
where $(5)$ follows from algebra and the fact $j>C \log T$, $(6)$ follows from the fact that the rewards are subgaussian and $(7)$ follows as 
$0<\frac{\delta \sqrt{C\, \log T}}{ \sigma}-\sqrt{2\alpha \log t}<\frac{\delta \sqrt{C\, \log T}}{ \sigma}+\sqrt{2\alpha \log t}$
and the monotone property of the exponential function. Therefore, we have

\begin{equation*}\begin{aligned}
&\mathbb{E}\left[\sum_{t=1}^{T}\mathbf{1}\left\{\exists \, i \in \mathcal{N}: 	\left(\left|\bar{\nu}_{i}(t)-\mu_{i}\right|\ge \delta\right)\cap\left(I_t=i\right)\cap \left(T_i(t-1)> C\, \log T\right) \right\}\right]\\
&\le \frac{2\,N}{T^{\hat{\Delta}^2/2-2}}.
\end{aligned}\end{equation*}
For the LCB values, we have

\begin{equation*}\begin{aligned}
&\mathbb{E}\left[\sum_{t=1}^{T}\mathbf{1}\left\{\exists \, i \in \mathcal{N}: 	\left(\left|\underline{\nu}_i(t)-\mu_{i}\right|\ge \delta \right)\cap\left(I_t=i\right)\cap \left(T_i(t-1)> C\, \log T\right) \right\}\right]\\
&\le \sum_{i\in \mathcal{N}}\sum_{t=1}^{T}\mathbb{P}\left\{	\left(\left|\underline{\nu}_i(t)-\mu_{i}\right|\ge \delta \right)\cap \left(T_i(t-1)> C\, \log T\right)\right\}\\
&\le \sum_{i\in \mathcal{N}}\sum_{t=1}^{T}\sum_{j=[C\log T]+1}^{T}\mathbb{P}\left[	\left(\left|\underline{\nu}_i(t)-\mu_{i}\right|\ge \delta \right)\cap \left(T_i(t-1)=j\right)\right].
\end{aligned}\end{equation*}
And we further have,

\begin{equation*}
\begin{aligned}
&\mathbb{P}\left\{\left(\underline{\nu}_i(t)-\mu_{i}\le -\delta \right)\cap \left(T_i(t-1)=j\right)\right\}\\
&\overset{(8)}{\le} \mathbb{P}\left\{\left(\frac{\sqrt{j}}{\sigma}\left(\mu_{i}-\hat{\mu}_i(t)\right)\ge \frac{\delta \sqrt{C\, \log T}}{\sigma}-\sqrt{2\alpha \log t} \right)\cap \left(T_i(t-1)=j\right)\right\}\\
&\overset{(9)}{\le} \exp\left\{-\frac{1}{2}\left(\frac{\delta\sqrt{C\, \log T}}{\sigma}-\sqrt{2\alpha \log t}\right)^2\right\}\\
&\le \frac{1}{T^{\hat{\Delta}^2/2}},
\end{aligned}
\end{equation*}
where $(8)$ follows from algebraic manipulation and $j>C\log T$ and $(9)$ follows from the subgaussian arm errors. For the right tail, we have

\begin{equation*}
\begin{aligned}
&\mathbb{P}\left\{\left(\underline{\nu}_i(t)-\mu_{i}\ge \delta \right)\cap \left(T_i(t-1)=j\right)\right\}\\
&\overset{(10)}{\le} \mathbb{P}\left\{\left(\frac{\sqrt{j}}{\sigma}\left(\hat{\mu}_i(t)-\mu_i\right)\ge \frac{\delta \sqrt{C\, \log T}}{\sigma}+\sqrt{2\alpha \log t} \right)\cap \left(T_i(t-1)=j\right)\right\}\\
&\le \frac{1}{T^{\hat{\Delta}^2/2}},
\end{aligned}
\end{equation*}
where $(10)$ follows from subgaussian arm rewards and the rest of the argument is identical to the argument for left tail of the UCB.

Therefore
\begin{equation*}\begin{aligned}
&\mathbb{E}\left[\sum_{t=1}^{T}\mathbf{1}\left\{\exists \, i \in \mathcal{N}: 	\left(\left|\underline{\nu}_{i}(t)-\mu_{i}\right|\ge \delta\right)\cap\left(I_t=i\right)\cap \left(T_i(t-1)> C\, \log T\right) \right\}\right]\\
&\le \frac{2\,N}{T^{\hat{\Delta}^2/2-2}}.
\end{aligned}\end{equation*}

Hence we are done. 
\hfill \Halmos 

\begin{corollary}\label{main:coro:suffpull}
	Under the conditions of Lemma \ref{subgaus:lemma}, we have
	\begin{equation*}\begin{aligned}
	\mathbb{E}\left[\sum_{t=1}^{T}\mathbf{1}\left\{\exists \, i \in \mathcal{N}: 	\left(\left|\bar{\nu}_i(t)-\mu_{i}\right|\ge \delta \right)\cap \left(I_t=i\right) \right\}\right] \le N\,C\log T +\frac{2\,N}{T^{\hat{\Delta}^2/2-2}}
	\end{aligned}\end{equation*}
	
	where $C=\left(\sqrt{2\alpha }+\hat{\Delta}\right)^2
	\sigma^2\, \delta^{-2}.$
\end{corollary}

\noindent\textbf{Proof of Corollary~\ref{main:coro:suffpull}.}
Note that for the event in question we have 
\begin{equation*}\begin{aligned}
&\mathbb{E}\left[\sum_{t=1}^{T}\mathbf{1}\left\{\exists \, i \in \mathcal{N}: 	\left(\left|\bar{\nu}_i(t)-\mu_{i}\right|\ge \delta \right)\cap\left(I_t=i\right) \right\}\right] \\
& \le \mathbb{E}\left[\sum_{t=1}^{T}\mathbf{1}\left\{\exists \, i \in \mathcal{N}: 	\left(\left|\bar{\nu}_i(t)-\mu_{i}\right|\ge \delta \right)\cap\left(I_t=i\right)\cap \left(T_i(t-1)> C\log T\right) \right\}\right]\\ &+\mathbb{E}\left[\sum_{t=1}^{T}\mathbf{1}\left\{\exists \, i \in \mathcal{N}: 	\left(\left|\bar{\nu}_i(t)-\mu_{i}\right|\ge \delta \right)\cap\left(I_t=i\right)\cap \left(T_i(t-1)\le  C\log T\right) \right\}\right].
\end{aligned}\end{equation*} 
Using Lemma~\ref{subgaus:lemma} we see that

\begin{equation*}
\begin{aligned}
&\mathbb{E}\left[\sum_{t=1}^{T}\mathbf{1}\left\{\exists \, i \in \mathcal{N}: 	\left(\left|\bar{\nu}_i(t)-\mu_{i}\right|\ge \delta \right)\cap\left(I_t=i\right)\cap \left(T_i(t-1)> C\log T\right) \right\}\right]\\ 
& \le \frac{2N}{T^{\hat{\Delta}^2/2-2}}.
\end{aligned}
\end{equation*}

Further, we see that

\begin{equation*}\begin{aligned}
&\mathbb{E}\left[\sum_{t=1}^{T}\mathbf{1}\left\{\exists \, i \in \mathcal{N}: 	\left(\left|\bar{\nu}_i(t)-\mu_{i}\right|\ge \delta \right)\cap\left(I_t=i\right)\cap \left(T_i(t-1)\le  C\log T\right) \right\}\right]\\
&\le \mathbb{E}\left[\sum_{t=1}^{T}\mathbf{1}\left\{\exists \, i \in \mathcal{N}: 	\left(I_t=i\right)\cap \left(T_i(t-1)\le  C\log T\right) \right\}\right]\\
& \le N \, C \, \log T.
\end{aligned}\end{equation*}
Hence we are done.
\hfill \Halmos 

The following result is present in  \citep{cen2022regret}; we furnish this in our work for completeness. 
\begin{lemma}\label{Lemma:cen:shah}
	Consider a MAB with $N$ arms having arm rewards $X_i(t)=\mu_i+\epsilon_i(t)$ with $\epsilon_i(t)$, $\sigma^2$-subgaussian and $\mu_i$ as the true arm reward. Further, denote by $\bar{\nu}_i(t),\underline{\nu}_i(t)$ the UCB and LCB of the reward estimate for the $i$-th arm at time $t\in \{1,2,\cdots,T\}$ defined in $(\ref{UCBdef})$ and $(\ref{LCBdef})$. Then 
	$$\mathbb{P}\left(\bar{\nu}_i(t)\le \mu_i\right)\le \frac{1}{t^{\alpha-1}}
	\textrm{~~and~~}
	\mathbb{P}\left(\underline{\nu}_i(t)\ge \mu_i\right)\le \frac{1}{t^{\alpha-1}}.$$
\end{lemma}
\textbf{Proof of Lemma~\ref{Lemma:cen:shah}.} The proof follows by decomposing the event in terms of the number of times arm $i$ has been pulled. Note that we may pull arm $i$ at most $t$ times if we are at time instance $t$.

\begin{equation*}
\begin{aligned}
&\mathbb{P}\left( \bar{\nu}_{i}(t) \le  \mu_{i} \right)\\
&=\sum_{\tau=1}^{t} \mathbb{P}\left( \bar{\nu}_{i}(t) \le  \mu_{i} \mid T_i(t-1)=\tau\right) \, \mathbb{P}\left(T_i(t-1)=\tau\right)\\
&\le \sum_{\tau=1}^{t} \mathbb{P}\left( \bar{\nu}_{i}(t) \le  \mu_{i} \mid T_i(t-1)=\tau\right)\\
&\le \sum_{\tau=1}^{t} \mathbb{P}\left(  \frac{\sqrt{\tau}\left(\mu_{i}-\hat{\mu}_i(t)\right)}{\sigma} \ge \sqrt{2\log t^{\alpha}}\right)\\
&\le \sum_{\tau=1}^{t} \frac{1}{t^{\alpha}}\\
&=\frac{1}{t^{\alpha-1}}.
\end{aligned}   
\end{equation*}

The proof for $\mathbb{P}\left(\underline{\nu}_i(t)\ge \mu_i\right)$ is similar.
Hence we are done. \hfill \Halmos

\noindent\textbf{Proof of Theorem~\ref{thm:unit:demand}.}
To establish this result we will first show a concentration bound for the UCB and the LCB estimates of the arms. Subsequently, we will establish that if the arms are sufficiently pulled then our algorithm chooses the correct max-min allocation. Namely, we show two steps-given an allocation not exploring the true minimum in the allocation occurs at most $O(\log T)$ times and given any two allocations, not choosing the allocation with higher minimal reward occurs at most $O(\log T)$ times. Finally we combine the results which results in our regret bound.

First, we establish that for any $\phi \in \mathcal{M}$, 

\begin{equation*}\begin{aligned}
\mathbb{P}\left(\exists \,\phi \in \mathcal{M} :\, \min_{j \in \mathcal{K}} \bar{\nu}^j_{\phi(j)}(t) \le \min_{j \in \mathcal{K}} \mu^{j}_{\phi(j)}\right)\le \frac{K\, N}{t^{\alpha-1}}.
\end{aligned}\end{equation*}

To observe this, note that
\begin{equation*}\begin{aligned}
& \mathbb{P}\left(\exists \,\phi \in \mathcal{M} :\, \min_{j \in \mathcal{K}} \bar{\nu}^j_{\phi(j)}(t) \le \min_{j \in \mathcal{K}} \mu^{j}_{\phi(j)}\right)\\
& \overset{(\text{I})}{\le} \mathbb{P}\left( \exists \, j_1 \in \mathcal{K}, i_1\in \mathcal{N}: \, \bar{\nu}^{j_1}_{i_1}(t) \le  \mu^{j_1}_{i_1} \right)\\
& \overset{(\text{II})}{\le }\sum_{j_1=1}^{K}\sum_{i=1}^{N} \mathbb{P}\left( \bar{\nu}^{j_1}_{i_1}(t) \le  \mu^{j_1}_{i_1} \right)\\
&\le \frac{K\,N}{t^{\alpha-1}}.
\end{aligned}\end{equation*}

where $(\text{I})$ follows as there exits some $j_1 \in \mathcal{K}$ and $i_1=\phi(j_1)$ such that 
\[\bar{\nu}^{j_1}_{i_1}(t)=\min_{j \in \mathcal{K}} \bar{\nu}^j_{\phi(j)}(t) \le \min_{j \in \mathcal{K}} \mu^{j}_{\phi(j)}\le \mu^{j_1}_{i_1}.\] $(\text{II})$ follows from probability laws and where the last line 
follows from Lemma~\ref{Lemma:cen:shah}. Note that we may apply the same lemma here as this is indeed a MAB with each agent-item pair forming an arm. 
Next we establish that the number of discordant pairs between the UCB arm estimates and the true rewards till time $T$ is $O(\log T)$. 
Namely, we control the event $E^{\phi}_1(t)$ defined as
\begin{equation*}
\begin{aligned}
\left\{\exists \, \phi_1 \in \mathcal{M}\backslash \Phi^* , \,\phi_2 \in \Phi^*, \, j_1 \in \mathcal{K} \, : \left(I_t=(j_1, \phi_1(j_1))\right)\cap\left(\min_{j \in \mathcal{K}} \bar{\nu}^{j}_{\phi_1(j)}(t) \ge \min_{j \in \mathcal{K}} \bar{\nu}^{j}_{\phi_2(j)}(t)\right)\right\}
\end{aligned}
\end{equation*}
which represents a mismatch of ordering between the truth and the estimated. Further define the events 
\[A^{\phi}_1(t)=\left\{min_{j \in \mathcal{K}} \, \bar{\nu}^{j}_{\phi_2(j)}(t)
> \min_{j \in \mathcal{K}} \mu^{j}_{\phi_2(j)}\right\} \textrm{~and~}B^{\phi}_1(t)=\left\{I_t=\phi_1(j_1), \, j_1 \in \argmin_{j \in \mathcal{K}} \mu^{j}_{\phi_1(j)}\right\}\]
which respectively represent the events that the minimal UCB estimate is greater than the minimal true reward and the arm to explore belongs to the true minima of the chosen allocation. Therefore

\begin{equation*}\begin{aligned}
&\mathbb{E}\left[\sum_{t=1}^{T} \mathbf{1}\left\{E^{\phi}_1(t)\right\}\right]\\
& \le \mathbb{E}\left[\sum_{t=1}^{T} \mathbf{1}\left\{E^{\phi}_1(t)\cap A^{\phi}_1(t) \cap B^{\phi}_1(t)\right\}\right] +\mathbb{E}\left[\sum_{t=1}^{T} \mathbf{1}\left\{ E^{\phi}_1(t) \cap B^{\phi}_1(t)^c\right\}\right]+\mathbb{E}\left[\sum_{t=1}^{T} \mathbf{1}\left\{ A^{\phi}_1(t)^c\right\}\right].
\end{aligned}\end{equation*}
Note that on the event $E^{\phi}_1(t)\cap A^{\phi}_1(t) \cap B^{\phi}_1(t)$, one has 
\begin{equation*}\begin{aligned}
&\bar{\nu}^{j_1}_{\phi_1(j_1)}(t)- \mu^{j_1}_{\phi_1(j_1)}\\
&\overset{(1)}{\ge} \min_{j \in \mathcal{K}} \bar{\nu}^{j}_{\phi_1(j)}(t)-\min_{j \in \mathcal{K}} \mu^{j}_{\phi_1(j)}\\
& \overset{(2)}{\ge} \min_{j \in \mathcal{K}} \bar{\nu}^{j}_{\phi_2(j)}(t)-\min_{j \in \mathcal{K}} \mu^{j}_{\phi_1(j)}\\
& \overset{(3)}{\ge} \min_{j \in \mathcal{K}} \mu^{j}_{\phi_2(j)}- \min_{j \in \mathcal{K}} \mu^{j}_{\phi_1(j)}\\
& \overset{(4)}{\ge}  \Delta_{\min},
\end{aligned}\end{equation*}
where $(1), (2)$ follows from the event $E^{\phi}_1(t)$ as $\min_{j \in \mathcal{K}} \mu^{j}_{\phi_1(j)}=\mu^{j_1}_{\phi_1(j_1)}$, $(3)$ follows from the event $A^{\phi}_1(t)$ and $(4)$ follows from Assumption~\ref{assm:gap:MAB} due to the event $B^{\phi}_1(t)$. On the event $E^{\phi}_1(t)\cap B^{\phi}_1(t)^c$, consider any 
$j_2 \in \arg\min_{j \in \mathcal{K}} \mu^{j}_{\phi_1(j)}$. In this case,
we note that one of the two events hold-either $\mu_{\maxmin}-\mu^{j_1}_{\phi_1(j_1)}\ge  \Delta_{\min}/2$ or $\mu^{j_1}_{\phi_1(j_1)}-\mu^{j_2}_{\phi_1(j_2)}\ge  \Delta_{\min}/2$. This is because we know that $\mu_{\maxmin}-\mu^{j_2}_{\phi_1(j_2)}> \Delta_{\min}$ and then we divide the event based on which of the former two quantities $\mu^{j_1}_{\phi_1(j_1)}$ is closer to. Hence 

\begin{equation*}\begin{aligned}
\mathbb{E}\left[\sum_{t=1}^{T} \mathbf{1}\left\{ E^{\phi}_1(t)\cap B^{\phi}_1(t)^c\right\}\right] &\le \underset{(a)}{\underbrace{\mathbb{E}\left[\sum_{t=1}^{T} \mathbf{1}\left\{E^{\phi}_1(t) \cap \left( \mu_{\maxmin}-\mu^{j_1}_{\phi_1(j_1)}\ge  \Delta_{\min}/2\right)\right\}\right]}} \\
&\quad +\underset{(b)}{\underbrace{\mathbb{E}\left[\sum_{t=1}^{T} \mathbf{1}\left\{ E^{\phi}_1(t) \cap \left(\mu^{j_1}_{\phi_1(j_1)}-\mu^{j_2}_{\phi_1(j_2)}\ge  \Delta_{\min}/2\right)\right\}\right]}}.
\end{aligned} \end{equation*}

Further note that for the event in  $(a)$, $\exists \, j_3 \in \mathcal{K}$ such that 
\[\bar{\nu}^{j_3}_{\phi_2(j_3)}(t) = \min_{j \in \mathcal{K}} \bar{\nu}^{j}_{\phi_2(j)}(t)
\le \min_{j \in \mathcal{K}} \bar{\nu}^{j}_{\phi_1(j)}(t) = \bar{\nu}^{j_1}_{\phi_1(j_1)}(t);\]
however, since $\phi_2 \in \Phi^*$, 
\[\mu^{j_3}_{\phi_2(j_3)} \ge \mu_{\maxmin}>\mu^{j_1}_{\phi_1(j_1)}+\Delta_{\min}/2.\]
Consider the event $A^{\phi}_2(t)=\{\forall \, j \in \mathcal{K}: \, \bar{\nu}^{j}_{\phi_2(j)}(t) \ge \mu^{j}_{\phi(j)}\}$.  Note that on 
\[E^{\phi}_1(t) \cap \left( \mu_{\maxmin}-\mu^{j_1}_{\phi_1(j_1)}\ge  \Delta_{\min}/2\right)\cap A^{\phi}_2(t)\] one has

\begin{equation*}
\begin{aligned}
& \bar{\nu}^{j_1}_{\phi_1(j_1)}(t)- \mu^{j_1}_{\phi_1(j_1)} \\
& \overset{(4)}{\ge} \bar{\nu}^{j_3}_{\phi_2(j_3)}(t)- \mu^{j_1}_{\phi_1(j_1)}\\
& \overset{(5)}{\ge} \mu^{j_3}_{\phi_2(j_3)}(t)- \mu^{j_1}_{\phi_1(j_1)} \ge \Delta_{\min}/2    
\end{aligned}
\end{equation*}

where $(4),(5)$  hold due to the event in question. Therefore  
\begin{equation*}
\begin{aligned}
&\mathbb{E}\left[\sum_{t=1}^{T} \mathbf{1}\left\{E^{\phi}_1(t) \cap \left( \mu_{\maxmin}-\mu^{j_1}_{\phi_1(j_1)}\ge  \Delta_{\min}/2\right)\right\}\right]\\
& \le \mathbb{E}\left[\sum_{t=1}^{T} \mathbf{1}\left\{E^{\phi}_1(t) \cap \left( \mu_{\maxmin}-\mu^{j_1}_{\phi_1(j_1)}\ge  \Delta_{\min}/2\right)\cap A^{\phi}_2(t)\right\}\right] + \mathbb{E}\left[\sum_{t=1}^{T} \mathbf{1}\left\{ A^{\phi}_2(t)^c\right\}\right]\\
& \le \mathbb{E}\left[\sum_{t=1}^{T} \mathbf{1}\left\{I_t=\phi_1(j_1),\, \bar{\nu}^{j_1}_{\phi_1(j_1)}(t)- \mu^{j_1}_{\phi_1(j_1)} >\Delta_{\min}/2\right\}\right] + \mathbb{E}\left[\sum_{t=1}^{T} \mathbf{1}\left\{ A^{\phi}_2(t)^c\right\}\right].
\end{aligned}
\end{equation*}

On the event for $(b)$ note that both $\underline{\nu}^{j_1}_{\phi_1(j_1)}(t) \le \underline{\nu}^{j_2}_{\phi_1(j_2)}(t) $ and 
$\mu^{j_1}_{\phi_1(j_1)}>\mu^{j_2}_{\phi_1(j_2)}+\Delta_{\min}/2$ since the arm chosen to explore has minimal reward with respect to the LCB estimates but the true rewards have reverse ordering. Essentially the chosen arm is not the true minimal arm. Consider the event $$A^{\phi}_3(t)=\left\{\exists \, \phi_2, \, j \in \mathcal{K}\, : \underline{\nu}^{j_2}_{\phi_1(j_2)}(t)< \mu^{j_2}_{\phi(j_2)}\right\}$$
Therefore, on the event for \[E^{\phi}_1(t) \cap \left(\mu^{j_1}_{\phi_1(j_1)}-\mu^{j_2}_{\phi_1(j_2)}\ge  \Delta_{\min}/2\right)\cap A^{\phi}_3(t)\], we have 
\begin{equation*}
\begin{aligned}
& \mu^{j_1}_{\phi_1(j_1)}-\underline{\nu}^{j_1}_{\phi_1(j_1)}(t)\\
& \ge \mu^{j_1}_{\phi_1(j_1)}-\underline{\nu}^{j_2}_{\phi_1(j_2)}(t)\\
& \mu^{j_1}_{\phi_1(j_1)}-\mu^{j_2}_{\phi_1(j_2)} \ge \Delta_{\min}/2
\end{aligned}
\end{equation*}
\begin{equation*}
\begin{aligned}
& \mathbb{E}\left[\sum_{t=1}^{T} \mathbf{1}\left\{ E^{\phi}_1(t) \cap \left(\mu^{j_1}_{\phi_1(j_1)}-\mu^{j_2}_{\phi_1(j_2)}\ge  \Delta_{\min}/2\right)\right\}\right] \\
& \le \mathbb{E}\left[\sum_{t=1}^{T} \mathbf{1}\left\{I_t=\phi_1(j_1),\, \mu^{j_1}_{\phi_1(j_1)}-\underline{\nu}^{j_1}_{\phi_1(j_1)}(t)>\Delta_{\min}/2\right\}\right] + \mathbb{E}\left[\sum_{t=1}^{T} \mathbf{1}\left\{ A^{\phi}_3(t)^c\right\}\right]    
\end{aligned}
\end{equation*}

Combining these, we get

\begin{equation*}\begin{aligned}
&\mathbb{E}\left[\sum_{t=1}^{T} \mathbf{1}\left\{E^{\phi}_1(t)\right\}\right]\\
&\le \underaccent{\text{I}}{\underbrace{\mathbb{E}\left[\mathbf{1}\left\{\exists \, j_1 \in \mathcal{K}, \, \phi_1 \in \mathcal{M}: \, \left(I_t= \phi_1(j_1)\right)\cap\left(\bar{\nu}^{j_1}_{\phi_1(j_1)}(t)-\mu^{j_1}_{\phi_1(j_1)}>\Delta_{\min}\right)\right\}\right]}}\\
&\quad  +\underaccent{\text{II}}{\underbrace{\mathbb{E}\left[\sum_{t=1}^{T} \mathbf{1}\left\{I_t=\phi_1(j_1),\, \bar{\nu}^{j_1}_{\phi_1(j_1)}(t)- \mu^{j_1}_{\phi_1(j_1)} >\Delta_{\min}/2\right\}\right]}} + \underaccent{\text{III}}{\underbrace{\mathbb{E}\left[\sum_{t=1}^{T} \mathbf{1}\left\{I_t=\phi_1(j_1),\, \mu^{j_1}_{\phi_1(j_1)}-\underline{\nu}^{j_1}_{\phi_1(j_1)}(t)>\Delta_{\min}/2\right\}\right]}}\\
&\quad + \underaccent{\text{IV}}{\underbrace{\mathbb{E}\left[\sum_{t=1}^{T} \mathbf{1}\left\{ A^{\phi}_1(t)^c\right\}\right]}}+\underaccent{\text{V}}{\underbrace{\mathbb{E}\left[\sum_{t=1}^{T} \mathbf{1}\left\{ A^{\phi}_2(t)^c\right\}\right]}} + \underaccent{\text{VI}}{\underbrace{\mathbb{E}\left[\sum_{t=1}^{T} \mathbf{1}\left\{ A^{\phi}_3(t)^c\right\}\right]}}.
\end{aligned}\end{equation*}
Using the argument at the beginning, we have term IV less than 
\begin{equation*}\begin{aligned}
&\mathbb{E}\left[\sum_{t=1}^{T} \mathbf{1}\left\{\exists \, \phi_2 \in \mathcal{M}: \ \min_{j \in \mathcal{K}} \bar{\nu}^j_{\phi_2(j)}(t) \le \min_{j \in \mathcal{K}} \mu^{j}_{\phi_2(j)}\right\}\right]\\
&\le \sum_{t=1}^{T} \mathbb{P}\left\{\exists \, \phi_2 \in \mathcal{M}: \ \min_{j \in \mathcal{K}} \bar{\nu}^j_{\phi_2(j)}(t) \le \min_{j \in \mathcal{K}} \mu^{j}_{\phi_2(j)}\right\} \\
&\le  N\, K \left(1+\sum_{t=2}^{\infty}\frac{1}{t^{\alpha-1}}\right) \\
& \le N\, K \left(1+\int_{1}^{\infty}\frac{1}{t^{\alpha-1}} dt\right)\\
&\le \frac{ N\, K (\alpha-1)}{\alpha-2}.
\end{aligned}\end{equation*}
Using a similar argument for V and VI, we have the same bound for each of these terms. For terms I-III, we use Corollary~\ref{main:coro:suffpull} to get
$$\mathbb{E}\left[\sum_{t=1}^{T} \mathbf{1}\left\{E^{\phi}_1(t)\right\}\right]\le \frac{3\, K\, N \left(\alpha-1\right)}{\alpha-2}+3\, K\, N\left( C\, \log T+ \frac{2}{T^{\hat{\Delta}^2/2-2}}\right).$$
Therefore, noting that 
$$R_T \le \Delta_{\max}\, \mathbb{E}\left[\sum_{t=1}^{T} \mathbf{1}\left\{E^{\phi}_1(t)\right\}\right]$$
we get our final bound as 
$$3\, \Delta_{\max}\, \left[\frac{\, K\, N \left(\alpha-1\right)}{\alpha-2}+\, K\, N\left( C\, \log T+ \frac{2}{T^{\hat{\Delta}^2/2-2}}\right)\right].$$

\hfill \Halmos 

We may obtain a coarser regret bound where there is no identifiability condition for the max-min solutions. Denote by $\Phi(\delta)=\{\phi \in \mathcal{M}\backslash \Phi^*: \, \max_{\phi \in \Phi^*} \min_{j \in \mathcal{K}} \mu^{j}_{\phi(j)}-\min_{j \in \mathcal{K}} \mu^{j}_{\phi(j)}\le \delta \}$ the set of all allocations which are at most $\delta$ away from the optimal allocation. 
\begin{proposition}\label{prop:MAB:nogap}
	Let $\left|\mu^{j_1}_{i_1}-\mu^{j_2}_{i_2}\right|\le \Delta_{max}$ for all $i_1,i_2 \in \mathcal{N}$, $j_1,j_2 \in \mathcal{K}$ and some $\Delta_{\max}>0$ hold. Then regret of Algorithm~\ref{algo:ucblcb} satisfies
	$$ R_T\le 6\,\Delta_{\max} \, N\, K\left[\frac{\alpha-1}{\alpha-2}+1\right] + 4.5^{2/3}\, \left( \Delta_{\max} \, N\, K \, \left(\sqrt{2\alpha}+2\right)^2 \sigma^2 \log T\right)^{1/3} T^{2/3}.$$
\end{proposition}
\noindent\noindent\textbf{Proof of Proposition~\ref{prop:MAB:nogap}.}
Define, for any $\delta>0$,

$$\Phi(\delta)=\left\{\phi: \, \mu_{\maxmin}-\min_{j \in \mathcal{K}}\mu^{j}_{\phi(j)}>\delta \right\}.$$ 

We know that the regret for Algorithm~\ref{algo:ucblcb} is given as

\begin{equation*}
\begin{aligned}
R_T&=\mathbb{E}\left[\sum_{t=1}^{T} \left(\mu_{\maxmin}-\mu^{j}_{\phi_t(j)}\right)\right]\\
&=\mathbb{E}\left[\sum_{t=1}^{T} \left(\mu_{\maxmin}-\mu^{j}_{\phi_t(j)}\right) \mathbf{1}\left\{\phi_t \in \Phi(\Delta_T)\right\}\right] +\mathbb{E}\left[ \sum_{t=1}^{T} \left(\mu_{\maxmin}-\mu^{j}_{\phi_t(j)}\right) \mathbf{1}\left\{\phi_t \in \Phi(\Delta_T)^c\right\}\right]\\
& \le \mathbb{E}\left[\sum_{t=1}^{T} \left(\mu_{\maxmin}-\mu^{j}_{\phi_t(j)}\right) \mathbf{1}\left\{\phi_t \in \Phi(\Delta_T)\right\}\right] +\Delta_T \, T.
\end{aligned}
\end{equation*}
Note that the first term allows us to use Theorem~\ref{thm:unit:demand} with gap $\Delta_T$. Therefore, we have 
\begin{equation*}
\begin{aligned}
R_T &\le 3\,\Delta_{\max} \, N\, K\left[ \frac{\left(\sqrt{2\alpha}+2\right)^2 \, \sigma^2}{\Delta^2_T}\,  \log T+2\,\frac{\alpha-1}{\alpha-2}+2\right] + \Delta_T\, T.
\end{aligned}
\end{equation*}
Optimizing over $\Delta_T$, we get the optimum value of $\Delta_T$ to be 
$$\Delta_T=\left(6\, \Delta_{\max}\, N\, K\, \left(\sqrt{2\alpha}+2\right)^2 \sigma^2 \, T^{-1}\, \log T \right)^{1/3}.$$
Substituting, the value, we get,
\begin{equation*}
\begin{aligned}
R_T\le 6\,\Delta_{\max} \, N\, K\left[\frac{\alpha-1}{\alpha-2}+1\right] + 4.5^{2/3}\, \left( \Delta_{\max} \, N\, K \, \left(\sqrt{2\alpha}+2\right)^2 \sigma^2 \log T\right)^{1/3} T^{2/3}
\end{aligned}
\end{equation*}
and hence we are done.
\hfill \Halmos

Note that Proposition~\ref{prop:MAB:nogap} exhibits a rate $O((\log T\, T^2)^{1/3})$ for the regret. This is different from the $O(\sqrt{T\, \log T})$ rate for the regret in the classical MAB setting. This is because the analysis we perform provides is a rate which is of the order $\Delta^{-2}_{\min}$ instead of $\Delta^{-1}_{\min}$. The key reason for this is that the regret analysis is extremely challenging if we decompose it in terms of the allocations. 

\noindent\noindent\textbf{Proof of Proposition~\ref{MAB:prop:gap:ind}.}\\
Define for all $i$
\[\mathcal{G}^K_i=\left\{\text{ the set of all subsets of size } K \text{ with } \mu_i \text{ as true minimum reward}\right\}.\]
Note that for any $i$ with $\mu_i>\mu_{(K)}$, we have $\mathcal{G}_i=\emptyset$.
Then regret can be written as 
\[R_T=\mathbb{E}\left[\sum_{t=1}^{T} \sum_{i \in G(0)} \Delta_i(K) \mathbf{1}\left\{G_t \in \mathcal{G}^K_i\right\}\right].\]

Note that
\begin{equation*}
\begin{aligned}
\mathbb{E}\left[\sum_{t=1}^{T} \sum_{i \in G(0)} \Delta_i(K) \mathbf{1}\left\{G_t \in \mathcal{G}^K_i\right\}\right]&\le  \underset{\text{(I)}}{\underbrace{\mathbb{E}\left[\sum_{t=1}^{T} \sum_{i \in G(0)} \Delta_i(K) \mathbf{1}\left\{G_t \in \mathcal{G}^K_i, \ I_t=i\right\}\right]}}\\
&\quad  + \underset{\text{(II)}}{\underbrace{\mathbb{E}\left[\sum_{t=1}^{T} \sum_{i \in G(0)}  \Delta_i(K) \mathbf{1}\left\{G_t \in \mathcal{G}^K_i, \ I_t\ne i\right\}\right]}}.
\end{aligned}
\end{equation*}
Now, for (I), we have
\begin{equation*}
\begin{aligned}
&\mathbb{E}\left[\sum_{t=1}^{T} \sum_{i \in G(0)}  \Delta_i(K) \mathbf{1}\left\{G_t \in \mathcal{G}^K_i, \ I_t=i\right\}\right]\\
&\le \mathbb{E}\left[\sum_{t=1}^{T} \sum_{i \in G(0)}  \Delta_i(K) \mathbf{1}\left\{\exists \ j \in G^*: \ I_t=i, \ \bar{\nu}_i(t) \ge \bar{\nu}_j(t), \ \mu_i < \mu_j\right\}\right]\\
&\le \mathbb{E}\left[\sum_{t=1}^{T} \sum_{i \in G(0)}  \sum_{j \in G^*} \Delta_i(K) \mathbf{1}\left\{\exists \ j \in G^*: \ I_t=i, \ \bar{\nu}_i(t) \ge \bar{\nu}_j(t), \ \mu_i < \mu_j\right\}\right]  
\end{aligned}
\end{equation*}
since we pull arm $i$, which not in the true optimal set and there exists some $j \in G^*$ which was not selected in the $G_t$. 
By \citep[Lemma C.4.]{cen2022regret} and noting that $\Delta(i,j)\ge \Delta_i(K)$, we have
\begin{equation*}
\begin{aligned}
\mathbb{E}\left[\sum_{t=1}^{T} \sum_{i \in G(0)} \Delta_i(K) \mathbf{1}\left\{G_t \in \mathcal{G}^K_i, \ I_t=i\right\}\right] &\le K\, \sum_{i \in G(0)} \left(\frac{8\,  \sigma^2\alpha \,\log T}{\Delta_i(K)}+\frac{\Delta_i(K)\, \alpha}{\alpha-2}\right).
\end{aligned}
\end{equation*}

For (II), we have
\begin{equation*}
\begin{aligned}
&\mathbb{E}\left[\sum_{t=1}^{T} \sum_{i \in G(0)} \Delta_i(K) \mathbf{1}\left\{G_t \in \mathcal{G}^K_i, \ I_t\ne i\right\}\right]\\ 
&\le \mathbb{E}\left[\sum_{t=1}^{T}\sum_{i \in G(0)}\sum_{j: \, \mu_j\ge \mu_{(K)}} \Delta_i(K) \mathbf{1}\left\{I_t=j, \ \underbar{$\nu$}_i(t)\ge \underbar{$\nu$}_j(t), \ \mu_i <\mu_j\right\}\right]\\
&+\mathbb{E}\left[\sum_{t=1}^{T}\sum_{i \in G(0)}\sum_{j: \, \mu_i<\mu_j< \mu_{(K)}} \sum_{l \in G^*} \Delta_i(K) \mathbf{1}\left\{I_t=j, \ \bar{\nu}_l(t)\le \bar{\nu}_j(t), \ \mu_l >\mu_j\right\}\right]        
\end{aligned}
\end{equation*}
where the first term on the right hand side follows from the event that some incorrect arm is chosen which belongs to the optimal set $G^*$ and the second term follows from the event that the arm selected lies between the $K$-th arm and the $i$-th arm. In this case, note that this arm was selected in the set $G_t$ instead of some arm $l$ in the set $G^*$. Therefore again, using \citep[Lemma C.4.]{cen2022regret}, we obtain

\begin{equation*}
\begin{aligned}
&\mathbb{E}\left[\sum_{t=1}^{T} \sum_{i \in G(0)} \Delta_i(K) \mathbf{1}\left\{G_t \in \mathcal{G}^K_i, \ I_t\ne i\right\}\right]\\ 
&\le \sum_{i \in G(0)}\sum_{j: \, \mu_j\ge \mu_{(K)}} \Delta_i(K)\, \frac{8\, \sigma^2 \, \alpha \log T}{\Delta^2(i,j)}+\frac{\Delta_i(K)\, \alpha}{\alpha-2}\\
&+\sum_{i \in G(0)}\sum_{j: \, \mu_i<\mu_j< \mu_{(K)}} \sum_{l \in G^*} \Delta_i(K)\, \frac{8\, \sigma^2 \, \alpha \log T}{\Delta^2(l,j)}+\frac{\Delta_i(K)\, \alpha}{\alpha-2}\\
& \le \sum_{i \in G(0)} \frac{8 \, K\, \sigma^2 \, \alpha \log T}{\Delta_i(K)}+\sum_{i \in G(0)}\frac{8\, (N-K) \, K\, \sigma^2 \, \alpha \log T}{\Delta_i(K)}+\sum_{i\in G(0)} \left(\frac{ K\,\Delta_i(K)\, \alpha}{\alpha-2}+\frac{ (N-K)\,K\, \Delta_i(K)\, \alpha}{\alpha-2}\right).
\end{aligned}
\end{equation*}
This implies that 

\begin{equation*}
\begin{aligned}
R_T & \le \sum_{i \in G(0)} \left(2\, K+ (N-K)\, K\right)\frac{8 \, \sigma^2\alpha \,\log T}{\Delta_i(K)} +\sum_{i \in G(0)} \left(\frac{ 2\, K\,\Delta_i(K)\, \alpha}{\alpha-2}+\frac{ (N-K)\,K\, \Delta_i(K)\, \alpha}{\alpha-2}\right)
\end{aligned}
\end{equation*}

Now, we establish the gap-independent regret bound.
Using this regret decomposition with respect to $\mathcal{G}^K_i$  and 
noting that $G(0)\cap G(\Delta_T)=G(\Delta_T)$, we have
\[R_T\le \mathbb{E}\left[\sum_{t=1}^{T} \sum_{i \in G(\Delta_T)} \Delta_i(K) \mathbf{1}\left\{G_t \in \mathcal{G}^K_i\right\}\right]+ \mathbb{E}\left[\sum_{t=1}^{T} \sum_{i \in G(0)\cap G(\Delta_T)^c} \Delta_i(K) \mathbf{1}\left\{G_t \in \mathcal{G}^K_i\right\}\right].\]

The second term is equivalent to 
\[\mathbb{E}\left[\sum_{t=1}^{T} \sum_{i \in G(0)\cap G(\Delta_T)^c} \Delta_i(K) \mathbf{1}\left\{G_t \in \mathcal{G}^K_i\right\}\right]=\sum_{i \in G(0)\cap G(\Delta_T)^c} \Delta_i(K) \mathbb{E}\left[\sum_{t=1}^{T} \mathbf{1}\left\{G_t \in \mathcal{G}^K_i\right\}\right]\]
and hence the second term is less than $ \Delta_T\, T$ since each $\Delta_i(K) \le \Delta_T$ and 
\[\sum_{i \in G(0)\cap G(\Delta_T)^c} \, \mathbb{E}\left[\sum_{t=1}^{T} \mathbf{1}\left\{G_t \in \mathcal{G}^K_i\right\}\right]\le T.\]
Using the fact that on $G(\Delta_T)$, $\Delta_i(K) > \Delta_T$ we know that the first term is less than

\begin{equation*}
\begin{aligned}
&\mathbb{E}\left[\sum_{t=1}^{T} \sum_{i \in G(0)} \Delta_i(K) \mathbf{1}\left\{G_t \in \mathcal{G}^K_i, \ I_t\ne i\right\}\right]\\ 
&\le \frac{8\, (N-K) \, K\,  \sigma^2\alpha \,\log T}{\Delta_T}+K\, \sum_{i \in G(0)}\frac{\Delta_i(K)\, \alpha}{\alpha-2}\\
& + \frac{8\, (N-K) \, K\, \sigma^2 \, \alpha \log T}{\Delta_T}+\frac{8\, (N-K)^2 \, K\, \sigma^2 \, \alpha \log T}{\Delta_T}+\sum_{i \in G(0)} \left(\frac{ K\,\Delta_i(K)\, \alpha}{\alpha-2}+\frac{ (N-K)\,K\, \Delta_i(K)\, \alpha}{\alpha-2}\right).
\end{aligned}
\end{equation*}

This implies that 
\begin{equation*}
\begin{aligned}
R_T & \le T\, \Delta_T +\left(2\,(N-K)\, K+ (N-K)^2\, K\right)\frac{8 \, \sigma^2\alpha \,\log T}{\Delta_T} +\sum_{i \in G(0)} \left(\frac{ 2\, K\,\Delta_i(K)\, \alpha}{\alpha-2}+\frac{ (N-K)\,K\, \Delta_i(K)\, \alpha}{\alpha-2}\right)
\end{aligned}
\end{equation*}

Choosing \[\Delta_T=\sqrt{\frac{\left((N-K+1)^2-1\right)\, K\, 8 \, \sigma^2\alpha \,\log T}{T}}\] 
gives us 

\begin{equation*}
\begin{aligned}
R_T & \le 2\, \sqrt{\left((N-K+1)^2-1\right)\, K\, 8 \, \sigma^2\alpha\, T \,\log T} \\
&\quad +\sum_{i \in G(0)} \left(\frac{ 2\, K\,\Delta_i(K)\, \alpha}{\alpha-2}+\frac{ (N-K)\,K\, \Delta_i(K)\, \alpha}{\alpha-2}\right).
\end{aligned}
\end{equation*}
Hence we are done.

\hfill \Halmos
\section{Proofs for Section~\ref{sec:CMAB}}
\subsection{Proofs for Max-Min Allocation}
We establish our main results by solving a more general problem. Consider a CMAB with $N$ base arms having true rewards $\mu_1,\mu_2,\cdots,\mu_N$. 
Denote \begin{equation}\begin{aligned}\label{comb_set}
\mathcal{A}^m\subseteq \left\{S: \ S \in 2^{\mathcal{N}} \ , |S|\le m\right\}
\end{aligned}\end{equation} 
as the set of arm combinations with capacity constraint that at most $m$ arms are included.
Note that in Sections~\ref{sec:MAB} and~\ref{sec:CMAB}, the parameter $m$ is $1$ and $N$, respectively. Further, note that in Section~\ref{sec:CMAB}, $\overset{K}{\underset{j=1}{\cup}} \mathcal{A}_j = \mathcal{A}^m$. In our setting, for an element in $\mathcal{A}^m$, the reward is given by a function defined as $r:\mathcal{K}\times \mathbb{R}^N\times \mathcal{A}^m \to \mathbb{R}$. 
Hence we have a separate reward function for each agent. Define $\bm{\mu}(S)=\sum_{i\in S} \mu_i$.
Our objective is to find
\[ \argmax_{\phi\in \bar{\mathcal{M}}}\min_{1\le j \le K} r^j(\bm{\mu};\phi(j))\]
where we recall that \[\mathcal{M}=\left\{\phi:\mathcal{K}\to \mathcal{A}^{m}, \, \phi(i) \ne \phi(j)\right\}\] is the set of all ordered sets of size $K$ from $2^{\mathcal{A}^m}$, which has each element as $[\phi(1),\phi(2),\cdots,\phi(K)]$ with  $\phi(i)$ being the set of goods/super-arm assigned to agent $i$. Note that the $\phi(i)$'s need not be disjoint which is the case when we have allocations.  
As in Section~\ref{sec:CMAB}, define the max-min as $\phi^*=\mathcal{O}_1(\bm{\mu},\mathcal{A}^m,K)$ and $j^*=\mathcal{O}_2(\bm{\mu},\phi^*,K)$. 
We subsequently present our main results which can be used to establish the results of Section~\ref{sec:CMAB}.
\begin{lemma}\label{lemma:subgaus}
	For a fixed allocation $\phi \in \mathcal{M}$, we have
	\begin{equation*}\begin{aligned}
	\mathbb{P}\left(\min_{1\le j \le K} r^{j}(\bar{\bm{\nu}}(t);\phi(j))\le \min_{1\le j \le K} r^{j}(\bm{\mu};\phi(j))\right)\le \frac{N}{t^{\alpha-1}}.
	\end{aligned}\end{equation*}
\end{lemma}
\noindent\textbf{Proof of Lemma~\ref{lemma:subgaus}.}
Note that the event 
\[\left\{\min_{1\le j \le K} r^{j}(\bar{\bm{\nu}}(t);\phi(j))\le \min_{1\le j \le K} r^{j}(\bm{\mu};\phi(j))\right\}
\]
implies that there exists some $i$ such that 
\[r^{i}(\bar{\bm{\nu}}(t);\phi(i))=\min_{1\le j \le K} r^{j}(\bar{\bm{\nu}}(t);\phi(j))\le \min_{1\le j \le K} r^{j}(\bm{\mu};\phi(j)) \le r^{i}(\bm{\mu};\phi(i))\]
which in turn implies 
$\exists \ i' \ \text{such that } \ \bar{\nu}_{i'}(t)\le \mu_{i'}$  by the monotonicity condition of Assumption~\ref{assm:reward:Lipschitz}. Therefore 
\begin{equation*}\begin{aligned}
\mathbb{P}\left(\min_{1\le j \le K} r^{j}(\bar{\bm{\nu}}(t);\phi(j))\le \min_{1\le j \le K} r^{j}(\bm{\mu};\phi(j))\right) & \le \mathbb{P}\left(\exists \ i' \ \text{such that } \ \bar{\nu}_i(t)\le \mu_{i'} \right)\\
&\le \sum_{i=1}^{N} \mathbb{P}\left(\bar{\nu}_i(t)\le \mu_{i'}\right) \le \frac{N}{t^{\alpha-1}},
\end{aligned}\end{equation*}
where the last line follows from Lemma~\ref{Lemma:cen:shah}.
\hfill \Halmos

Define for each time point $t$ and $\delta>0$,

\begin{equation}\begin{aligned}\label{mismatch:upper}
U^{(1)}_t(\delta)&=\left\{\exists \ \phi \in \mathcal{M}, \, i,j \in \mathcal{K}:\left(I_t=\phi(i)\right)\cap\left(r^{i}(\bar{\bm{\nu}}(t);\phi(i))\ge r^{j}(\bar{\bm{\nu}}(t);\phi(j))\right)\right.\\
&\left. \quad \quad \quad \cap\left(r^{i}(\bm{\mu};\phi(i))+\delta<r^{j}(\bm{\mu};\phi(j)\right)\right\} 
\end{aligned}\end{equation}
as the event which indicates that there is an allocation at time $t$, for which, the rewards calculated using the UCB values are not correctly ordered with respect to the true rewards; and similarly define 
\begin{equation}\begin{aligned}\label{mismatch:lower}
L^{(1)}_t(\delta)&=\left\{\exists \ \phi \in \mathcal{M}, \, i,j \in \mathcal{K}:\left(I_t=\phi(i)\right)\cap\left(r^{i}(\underline{\nu}(t);\phi(i))\le r^{j}(\underline{\nu}(t);\phi(j))\right)\right.\\
&\left. \quad \quad \quad \quad \cap\left(r^{i}(\bm{\mu};\phi(i))>r^{j}(\bm{\mu})+\delta\right)\right\} 
\end{aligned}\end{equation}
as the event which indicates that the rewards calculated using the LCB values of the arms are not correctly ordered. We further define 

\begin{equation*}\begin{aligned}
M_t&=\left\{\exists \, \phi_1 \in \mathcal{M}\backslash \Phi^*, \, \phi_2 \in \Phi^*, \, i \in {\mathcal{K}}:\right.\\
&\quad \quad\left. \left(I_t=\phi_1(i)\right)\cap\left(\min_{1\le j \le K} r^{j}(\bar{\bm{\nu}}(t);\phi_1(j))\ge \min_{1\le j\le K} r^{j}(\bar{\bm{\nu}}(t);\phi_2(j))\right)\right.\\
&\left. \quad \quad \quad \quad \cap\left(\min_{1\le j \le K} r^{j}(\bm{\mu};\phi_1(j))<\min_{1\le j \le K} r^{j}(\bm{\mu};\phi_2(j))\right)\right\}
\end{aligned}\end{equation*}
as the set of allocations with mismatched minimum assignments.

We are now ready to state our main results. Our first result shows that the existence of an allocation containing an incorrect ordering using UCB or LCB estimates on the reward  function may happen at most $O(\log T)$ times.
\begin{theorem}
	\label{order:prop}
	Let $I_t$ denote the super arm chosen at time $t$ by Algorithm~\ref{algo1}. Under Assumptions~\ref{assm:reward:Lipschitz}-~\ref{assm:gap},
	\begin{equation*}\begin{aligned} \mathbb{E}\left[\sum_{t=1}^{T}\mathbf{1}(U^{(1)}_t(\delta))\right]
	&\le N\left[ \frac{\left(\sqrt{2\alpha}+2\right)^2 c^2 \, m^2\, \sigma^2}{\delta^2}\,  \log T+\frac{\alpha-1}{\alpha-2}+2\right]
	\end{aligned}\end{equation*}
	and 
	\begin{equation*}\begin{aligned}
	\mathbb{E}\left[\sum_{t=1}^{T}\mathbf{1}(L^{(1)}_t(\delta))\right] &\le N\left[ \frac{\left(\sqrt{2\alpha}+2\right)^2 c^2 \, m^2\, \sigma^2}{\delta^2}\,  \log T+\frac{\alpha-1}{\alpha-2}+2\right].
	\end{aligned}\end{equation*}
\end{theorem}
\noindent\textbf{Proof of Theorem~\ref{order:prop}.}
The main idea of the proof is to reduce the problem to an MAB. To do this we show that on an event of high probability there exists some explored arm which has either not been sufficiently pulled or falls into a region of low probability when the event in question is true.
We start with observing that for any $1\le t\le T$, we have
\begin{equation*}\begin{aligned}
&U^{(1)}_t(\delta)
=\left(U^{(1)}_t(\delta)\cap\left\{\left(r^{j}(\bar{\bm{\nu}}(t);\phi(j))> r^{j}(\bm{\mu};\phi(j))\right)\cap \left(r^{i}(\bar{\bm{\nu}}(t);\phi(i))> r^{i}(\bm{\mu};\phi(i))\right)\right\}\right)\\
&\quad \quad \bigcup \left(U^{(1)}_t(\delta)\cap\left\{\left(r^{j}(\bar{\bm{\nu}}(t);\phi(j))\le  r^{j}(\bm{\mu};\phi(j))\right)\cup \left(r^{i}(\bar{\bm{\nu}}(t);\phi(i))\le  r^{i}(\bm{\mu};\phi(i))\right)\right\}\right).
\end{aligned}\end{equation*}
Now, on 
\begin{equation*}\begin{aligned}
&U^{(1)}_t(\delta)\cap\left\{\left(r^{j}(\bar{\bm{\nu}}(t);\phi(j))> r^{j}(\bm{\mu};\phi(j))\right)\cap \left(r^{i}(\bar{\bm{\nu}}(t);\phi(i))> r^{i}(\bm{\mu};\phi(i))\right)\right\},
\end{aligned}\end{equation*}
there exists $i' \in \phi(i)$, such that
\begin{equation*}\begin{aligned}
\left|\bar{\nu}_{i'}(t)-\mu_{i'}\right|&\overset{(1)}{\ge} \frac{1}{\left|\phi(i)\right|} \sum_{i' \in \phi(i)} \left|\bar{\nu}_{i'}(t)-\mu_{i'}\right|\\
&\overset{(2)}{\ge} \frac{1}{m}\sum_{i' \in \phi(i)} \left|\bar{\nu}_{i'}(t)-\mu_{i'}\right|\\
& \overset{(3)}{\ge} \frac{1}{c\, m} \left|r^{i}(\bar{\bm{\nu}}(t);\phi(i))-r^{i}(\bm{\mu};\phi(i))\right|\\
&\overset{(4)}{\ge} \frac{r^{i}(\bar{\bm{\nu}}(t);\phi(i))-r^{i}(\bm{\mu};\phi(i))}{c \, m}\\
&\overset{(5)}{\ge} \frac{r^{j}(\bar{\bm{\nu}}(t);\phi(j))-r^{i}(\bm{\mu}\phi(i))}{c\, m}\\
&\overset{(6)}{\ge} \frac{r^{j}(\bm{\mu};\phi(j))-r^{i}(\bm{\mu};\phi(i))}{c\, m}\\
&\overset{(7)}{\ge} \frac{\delta}{c\,m}.
\end{aligned}\end{equation*}
$(1)$ holds as there exits one value greater than the average. $(2)$ holds as $|\phi(i)|\le m$. $(3)$ holds due to Assumption~\ref{assm:reward:Lipschitz}. (4)-(6) hold due to the event considered. (7) holds due to Assumption~\ref{assm:gap}.
Note that 
\[
\left(r^{i}(\bar{\bm{\nu}}(t);\phi(i))\le  r^{i}(\bm{\mu};\phi(i))\right)\cup\left(r^{j}(\bar{\bm{\nu}}(t);\phi(j))\le  r^{j}(\bm{\mu};\phi(j))\right) \subseteq \left\{\exists \ i' \in \mathcal{N}: \bar{\nu}_{i'}(t)\le \mu_{i'}\right\}\]
by Assumption~\ref{assm:reward:Lipschitz}. 
Thus, we have 
\begin{equation*}\begin{aligned}
U^{(1)}_t(\delta) &\subseteq \left\{\exists \, i' \in \mathcal{N}: 	\left(\left|\bar{\nu}_{i'}(t)-\mu_{i'}\right|\ge \frac{\delta}{c \,m}\right)\cap\left(I_t=i'\right) \right\}\\
&\quad \bigcup\left\{\exists \ i' \in \mathcal{N}: \bar{\nu}_{i'}(t)\le \mu_{i'}\right\}.
\end{aligned}\end{equation*}
This implies 
\begin{equation*}\begin{aligned}
&\mathbb{E}\left[\sum_{t=1}^{T} \mathbf{1}\left\{U^{(1)}_t(\delta)\right\}\right]\\
& \le \mathbb{E}\left[\sum_{t=1}^{T}\mathbf{1}\left\{\exists \, i' \in \mathcal{N}: 	\left(\left|\bar{\nu}_{i'}(t)-\mu_{i'}\right|\ge \frac{\delta}{c \,m}\right)\cap\left(I_t=i'\right) \right\}\right]\\
&\quad +\mathbb{E}\left[\sum_{t=1}^{T}\mathbf{1}\left\{\exists \ i' \in \mathcal{N}: \bar{\nu}_{i'}(t)\le \mu_{i'}\right\}\right].
\end{aligned}\end{equation*}

For the first term, using Corollary~\ref{main:coro:suffpull}, we have
\begin{equation*}\begin{aligned}
\mathbb{E}\left[\sum_{t=1}^{T} \mathbf{1}\left\{\exists \, i' \in \mathcal{N}: 	\left(\left|\bar{\nu}_{i'}(t)-\mu_{i'}\right|\ge \frac{\delta}{c \,m}\right)\cap\left(I_t=i'\right) \right\}\right]
\le N\, C\, \log T+\frac{2\,N}{T^{\hat{\Delta}^2/2-2}}
\end{aligned}\end{equation*}
where $C=\left(\sqrt{2\alpha}+\hat{\Delta}\right)^2\, m^2\, c^2 \sigma^2\, \delta^{-2}$
with $\hat{\Delta}\ge 2$.
For the third term, using Lemma~\ref{lemma:subgaus}, we have 
\begin{equation*}\begin{aligned}
\mathbb{E}\left[\sum_{t=1}^{T}\mathbf{1}\left\{\exists \ i' \in \mathcal{N}: \bar{\nu}_{i'}(t)\le \mu_{i'}\right\}\right]&\le \sum_{i' \in \mathcal{N}}\sum_{t=1}^{T}\mathbb{P}\left(\bar{\nu}_{i'}(t)\le \mu_{i'}\right)\\
&\le N\, \left(1+\sum_{t=2}^{\infty}\frac{1}{t^{\alpha-1}}\right)\\
& \le N\, \frac{\alpha-1}{\alpha-2}.
\end{aligned}\end{equation*}
Combining all bounds and taking $\hat{\Delta}=2$, the first result follows.
The second result can be similarly derived.

We observe that for any $1\le t\le T$, we have
\begin{equation*}\begin{aligned}
L^{(1)}_t(\delta)
&=\left(L^{(1)}_t(\delta) \cap \left\{\left(r^{j}(\underline{\bm{\nu}}(t);\phi(j))< r^{j}(\bm{\mu};\phi(j))\right)\cap \left(r^{i}(\underline{\bm{\nu}}(t);\phi(i))< r^{i}(\bm{\mu};\phi(i))\right)\right\}\right)\\
&\quad \quad \bigcup \left( L^{(1)}_t(\delta) \cap \left\{ \left( \left(r^{j}(\underline{\bm{\nu}}(t);\phi(j))\ge  r^{j}(\bm{\mu};\phi(j))\right)\cup \left(r^{i}( \underline{\bm{\nu}}(t);\phi(i))\ge  r^{i}(\bm{\mu};\phi(i))\right)\right)\right\}\right).
\end{aligned}\end{equation*}
On 
\begin{equation*}\begin{aligned}
& L^{(1)}_t(\delta) \cap \left\{\left(r^{j}(\underline{\bm{\nu}}(t);\phi(j))< r^{j}(\bm{\mu};\phi(j))\right)\cap \left(r^{i}(\underline{\bm{\nu}}(t);\phi(i))< r^{i}(\bm{\mu};\phi(i))\right)\right\}
\end{aligned}\end{equation*}
there exists an $i' \in \phi(i)$ such that
\begin{equation*}\begin{aligned}
\left|\mu_{i'}-\underline{\nu}_{i'}(t)\right|&\overset{(1)}{\ge} \frac{1}{\left|\phi(i)\right|}\sum_{i'\in \phi(i)}\left|\mu_{i'}-\underline{\nu}_{i'}(t)\right|\\
&\overset{(2)}{\ge }\frac{1}{c\, m} \left|r^{i}(\bm{\mu};\phi(i))-r^{i}(\underline{\bm{\nu}}(t);\phi(i))\right|\\
&\overset{(3)}{\ge } \frac{1}{c\, m} \left(r^{i}(\bm{\mu};\phi(i))-r^{i}(\underline{\bm{\nu}}(t);\phi(i))\right)\\
&\overset{(4)}{\ge } \frac{1}{c\, m} \left(r^{i}(\bm{\mu};\phi(i))-r^{j}(\underline{\bm{\nu}}(t);\phi(j))\right)\\
&\overset{(5)}{\ge} \frac{1}{c\, m} \left(r^{i}(\bm{\mu};\phi(i))-r^{j}(\bm{\mu};\phi(j))\right)\\
&\overset{(6)}{\ge} \frac{\delta}{c\, m}.
\end{aligned}\end{equation*}
Again, $(1)$ occurs by property of the mean, (2) by Assumption~\ref{assm:reward:Lipschitz} and (3)-(5) are due to the set considered and (6) is due to Assumption~\ref{assm:gap}. Also, note that,
\begin{equation*}\begin{aligned}
\left(r^{j}(\underline{\bm{\nu}}(t);\phi(j))\ge  r^{j}(\bm{\mu};\phi(j))\right)\cup \left(r^{i}(\underline{\bm{\nu}}(t);\phi(i))\ge  r^{i}(\bm{\mu};\phi(i))\right)\subseteq \left\{\exists \, i' \in \mathcal{N}: \mu_{i'}\le  \underline{\nu}_{i'}(t)\right\}.
\end{aligned}\end{equation*}
Therefore 
\begin{equation*}\begin{aligned}
L^{(1)}_t(\delta) &\subseteq \left\{\exists \, i' \in \mathcal{N}: 	\left(\left| \underline{\nu}_{i'}(t)-\mu_{i'}\right|\ge \frac{\delta}{c\,m}\right)\cap\left(I_t=i'\right) \right\}\\
&\quad \bigcup\left\{\exists \ i' \in \mathcal{N}:  \underline{\nu}_{i'}(t)\ge \mu_{i'}\right\}.
\end{aligned}\end{equation*}
The rest of the proof is identical to the previous part. 
\hfill \Halmos


Using Theorem~\ref{order:prop}, we obtain our next result.
\begin{theorem}
	\label{thm:correct:allocation}
	Let $I_t$ denote the super arm chosen at time $t$ using Algorithm~\ref{algo1}. Then under Assumptions~\ref{assm:reward:Lipschitz}-\ref{assm:gap}, one has 
	\begin{equation*}\begin{aligned}
	\sum_{t=1}^{T} \mathbb{E}\left[\mathbf{1}(M_t)\right]
	&\le 2\,N\left[ \frac{\left(\sqrt{2\alpha}+2\right)^2 c^2 \, m^2\, \sigma^2}{\tilde{\Delta}^2_{\min}}\,  \log T+\frac{\alpha-1}{\alpha-2}+2\right] 
	\end{aligned}\end{equation*}
\end{theorem}

\noindent\textbf{Proof of Theorem~\ref{thm:correct:allocation}.}
Again, the idea is similar to Theorem~\ref{order:prop} where we reduce the problem to an MAB setting where we show that the explored arm is either not sufficiently pulled or the event falls in a region of low probability.
We start by defining some events which we shall use throughout the proof. Define 
\[A^{(2)}_{\phi}(t)=\left(r^{i}(\bar{\bm{\nu}}(t);\phi_1(i)) > r^{i}(\bm{\mu};\phi_1(i))\right)\]
as the event that the UCB estimate for the reward is higher than the true reward and  
\[E^{(2)}_{\phi}(t)=\left\{ i\in \arg\min_{1\le j\le K}r^{j}(\bm{\mu};\phi_1(j))\right\}.\]
Note that on the event

\begin{equation*}\begin{aligned}
M_t&=\left\{\exists \, \phi_1 \in \mathcal{M}\backslash \Phi^*, \, \phi_2 \in \Phi^*, \, i \in {\mathcal{K}}:\right.\\
&\quad \quad\left. \left(I_t=\phi(i)\right)\cap\left(\min_{1\le j \le K} r^{j}(\bar{\bm{\nu}}(t);\phi_1(j))\ge \min_{1\le j\le K} r^{j}(\bar{\bm{\nu}}(t);\phi_2(j))\right)\right.\\
&\left. \quad \quad \quad \quad \cap\left(\min_{1\le j \le K} r^{j}(\bm{\mu};\phi_1(j))<\min_{1\le j \le K} r^{j}(\bm{\mu};\phi_2(j))\right)\right\}
\end{aligned}\end{equation*}
we have 
\[r^{i}(\bar{\bm{\nu}}(t);\phi_1(i))\ge \min_{1\le j \le K} r^{j}(\bar{\bm{\nu}}(t);\phi_1(j))\ge \min_{1\le j\le K} r^{j}(\bar{\bm{\nu}}(t);\phi_2(j)).\]
Further,

\begin{equation*}
\begin{aligned}
&\mathbb{E}\left[\sum_{t=1}^{T}\mathbf{1}\left(M_t\right)\right]\\
&\le \mathbb{E}\left[\sum_{t=1}^{T}\mathbf{1}\left(M_t\cap A^{(2)}_{\phi}(t)^c \right)\right] + \mathbb{E}\left[\sum_{t=1}^{T}\mathbf{1}\left(M_t\cap A^{(2)}_{\phi}(t)\cap E^{(2)}_{\phi}(t) \right)\right]+ \mathbb{E}\left[\sum_{t=1}^{T}\mathbf{1}\left(M_t\cap A^{(2)}_{\phi}(t)\cap E^{(2)}_{\phi}(t)^c \right)\right] .     
\end{aligned}
\end{equation*}

On $M_t\cap A^{(2)}_{\phi}(t)\cap E^{(2)}_{\phi}(t)$, one has
\begin{equation*}\begin{aligned}
\left|\bar{\nu}_{i'}(t)-\mu_{i'}\right|&\ge \frac{1}{m}\sum_{j \in \phi_1(i)}\left|\bar{\nu}_{j}(t)-\mu_j\right|\\
& \ge \frac{1}{m \, c} \left|r^{i}(\bar{\bm{\nu}}(t);\phi_1(i))-r^{i}(\bm{\mu};\phi_1(i))\right|
\end{aligned}\end{equation*}
where the first inequality follows from the property of mean with $|\phi(i)|\le m$ and the second inequality follows from Assumption~\ref{assm:reward:Lipschitz}. 
Thus, 
\begin{equation*}\begin{aligned}
\left|\bar{\nu}_{i'}(t)-\mu_{i'}\right|&\ge\frac{1}{m \, c} \left(r^{i}(\bar{\bm{\nu}}(t);\phi_1(i))-r^{i}(\bm{\mu};\phi_1(i))\right)\\
&\ge \frac{1}{m \, c} \left(\min_{1\le j \le K} r^{j}(\bar{\bm{\nu}}(t);\phi_1(j))-r^{i}(\bm{\mu};\phi_1(i))\right)\\
& \ge \frac{1}{m \, c} \left(\min_{1\le j \le K} r^{j}(\bar{\bm{\nu}}(t);\phi_2(j))-r^{i}(\bm{\mu};\phi_1(i))\right)\\
&\ge \frac{1}{m \, c} \left(\min_{1\le j \le K} r^{j}(\bar{\bm{\nu}}(t);\phi_2(j))-\min_{1\le j \le K} r^{j}(\bm{\mu};\phi_1(j))\right)
&\ge \frac{\tilde{\Delta}_{\min}}{m\, c}.
\end{aligned}\end{equation*}
where the lines three and four follow by the set considered and the last inequality follows from Assumption~\ref{assm:gap}.

Note that the probability of the set 

\begin{equation*}
\begin{aligned}
&\mathbb{P}\left\{M_t \cap A^{(2)}_{\phi}(t)^c\right\}\\
&\le \mathbb{P}\left(r^{i}(\bar{\bm{\nu}}(t);\phi_1(i)) \le r^{i}(\bm{\mu};\phi_1(i))\right)\\
&\le \mathbb{P}\left(\exists \, i' \in \mathcal{N}: \, \bar{\nu}_{i'}(t)\le \mu_{i'}\right)\\
&\le \frac{N}{t^{\alpha-1}}  
\end{aligned}
\end{equation*}
where the last two lines follow from Assumption~\ref{assm:reward:Lipschitz} and Lemma~\ref{Lemma:cen:shah} respectively.
On the event, $M_t\cap A^{(2)}_{\phi}(t)\cap E^{(2)}_{\phi}(t)^c$, define, without loss of generality, 
$j'=\arg\min_{1\le j \le K} r^{j}(\bm \mu;\phi_1(j))$ ( again we may take any $j' \in \arg\min_{1\le j \le K} r^{j}(\bm \mu;\phi_1(j))$ and the proof does not change). Note that on $M_t\cap A^{(2)}_{\phi}(t)\cap E^{(2)}_{\phi}(t)^c$, one of the following events occur- either $r^{j'}(\bm{\mu};\phi_1(j'))-\min_{j \in \mathcal{K}} r^{j}(\bm{\mu};\phi_1(j))> \tilde{\Delta}_{\min}/2$ or  $\min_{j \in \mathcal{K}} r^{j}(\bm{\mu};\phi_2(j))-r^{j'}(\bm{\mu};\phi(j'))>\tilde{\Delta}_{\min}/2$. Therefore on the event 
\[M_t\cap A^{(2)}_{\phi}(t)\cap E^{(2)}_{\phi}(t)^c\cap \left(\min_{j \in \mathcal{K}} r^{j}(\bm{\mu};\phi_2(j))-r^{j'}(\bm{\mu};\phi(j'))>\tilde{\Delta}_{\min}/2\right)\]
we have 

\begin{equation*}\begin{aligned}
& \min_{j \in \mathcal{K}} r^{j}(\bm{\mu};\phi_2(j))> r^{j'}(\bm{\mu};\phi_1(j'))+\tilde{\Delta}_{\min}/2\\
& \min_{j \in \mathcal{K}} r^{j}( \bar{\bm{\nu}}(t);\phi_2(j))\le \min_{j \in \mathcal{K}} r^{j}( \bar{\bm{\nu}}(t);\phi_1(j))=r^{j'}(\bar{\bm{\nu}}(t);\phi_1(j')).
\end{aligned}\end{equation*}
Consider the event $A^{(3)}_{\phi}(t)=\{\min_{j \in \mathcal{K}} r^{j}( \bar{\bm{\nu}}(t);\phi_2(j)) \ge \min_{j \in \mathcal{K}} r^{j}(\bm{\mu};\phi_2(j))\}$
Therefore on the event
\[M_t\cap A^{(2)}_{\phi}(t)\cap E^{(2)}_{\phi}(t)^c\cap \left(\min_{j \in \mathcal{K}} r^{j}(\bm{\mu};\phi_2(j))-r^{j'}(\bm{\mu};\phi(j'))>\tilde{\Delta}_{\min}/2\right)\cap A^{(3)}_{\phi}(t),\]
we have 
\begin{equation*}
\begin{aligned}
&r^{j'}(\bar{\bm{\nu}}(t);\phi_1(j'))-r^{j'}(\bm{\mu};\phi_1(j'))\\
&\ge \min_{j \in \mathcal{K}} r^{j}( \bar{\bm{\nu}}(t);\phi_2(j))-r^{j'}(\bm{\mu};\phi_1(j'))\\
& \ge \min_{j \in \mathcal{K}} r^{j}(\bm{\mu};\phi_2(j))-r^{j'}(\bm{\mu};\phi_1(j')) \ge \tilde{\Delta}_{\min}/2
\end{aligned}
\end{equation*}

Therefore on the event in question, there exists some $i' \in \phi_1(j')$ such that 

\begin{equation*}
\begin{aligned}
|\bar{\nu}_{i'}(t)-\mu_{i'}|&\ge \frac{1}{m} \sum_{i \in \phi(j')} |\bar{\nu}_{i}(t)-\mu_{i}|\\
& \ge \frac{1}{mc} \left|r^{j'}(\bar{\bm{\nu}}(t);\phi_1(j'))-r^{j'}(\bm{\mu};\phi_1(j'))\right|\\
& \ge \frac{\tilde{\Delta}_{\min}}{2\, m\, c}. 
\end{aligned}
\end{equation*}
Note that the event $A^{(3)}_{\phi}(t)^c$ has probability at most $N/t^{\alpha-1}$. 
Finally, on the event 
\[M_t\cap A^{(2)}_{\phi}(t)\cap E^{(2)}_{\phi}(t)^c\cap \left(r^{j'}(\bm{\mu};\phi_1(j'))-\min_{j \in \mathcal{K}} r^{j}(\bm{\mu};\phi_1(j))> \tilde{\Delta}_{\min}/2\right),\] we have 
\begin{equation*}\begin{aligned}
& r^{i}(\bm{\mu};\phi_1(i))> r^{j'}(\bm{\mu};\phi_1(j')) + \tilde{\Delta}_{\min}/2\\
&r^{i}( \underline{\bm{\nu}}(t);\phi_1(i))\le r^{j'}(\underline{\bm{\nu}}(t);\phi_1(j')).
\end{aligned}\end{equation*}
Therefore the event $M_t\cap A^{(2)}_{\phi}(t)\cap E^{(2)}_{\phi}(t)^c$ implies that $L^{(1)}_t(\delta)$, defined in \eqref{mismatch:lower}, occurs for $\phi_1$.
Hence, combining the bounds, we have 
\begin{equation*}\begin{aligned}
\mathbb{E}\left[\sum_{t=1}^{T}\mathbf{1}(M_t)\right] &\le \mathbb{E}\left[\sum_{t=1}^{T}\mathbf{1}(L^{(1)}_t(\delta))\right] + 2\,  \sum_{t=1}^{T}\frac{N}{t^{\alpha-1}}\\
& \quad\quad +\mathbb{E}\left[\sum_{t=1}^{T}\mathbf{1}\left\{\left(\exists \, i' \in \mathcal{N}: \, \left|\bar{\nu}_{i'}(t)-\mu_{i'}\right|
\ge \frac{\tilde{\Delta}_{\min}}{m\, c}\right)\cap \left(I_t=i'\right) \right\}\right]\\
& \quad\quad +\mathbb{E}\left[\sum_{t=1}^{T}\mathbf{1}\left\{\left(\exists \, i' \in \mathcal{N}: \, \left|\bar{\nu}_{i'}(t)-\mu_{i'}\right|
\ge \frac{\tilde{\Delta}_{\min}}{2\,m\, c}\right)\cap \left(I_t=i'\right) \right\}\right] .
\end{aligned}\end{equation*}
Therefore the result follows using Theorem~\ref{order:prop} and Corollary~\ref{main:coro:suffpull} on the first, the third and the fourth term respectively.
\hfill \Halmos

\noindent\textbf{Proof of Theorem~\ref{coro:regret}.}
We note that

\begin{equation*}\begin{aligned}
R_t &\le \tilde{\Delta}_{\max}\mathbb{E}\left[ \sum_{t=1}^{T}\mathbf{1}\left\{\text{the correct allocation is not chosen at time }t\right\}\right] \le \tilde{\Delta}_{ \max}\, \mathbb{E}\left[\sum_{t=1}^{T}\mathbf{1}(M_t)\right]\
\end{aligned}\end{equation*}
and hence the result follows using Theorem~\ref{thm:correct:allocation}.
\hfill \Halmos 

We may also provide an instance independent regret bound in this setting
\begin{proposition}\label{prop:CMAB:inst:ind}
	Let for any $\phi \in \mathcal{M}$ and pair $i,j \in \mathcal{K}, \, i\ne j$, 
	\[\left|r^{j}(\bm{\mu};\phi(j))-r^{i}(\bm{\mu};\phi(i)) \right|\le \tilde{\Delta}_{\max}\] hold. Then, under Assumption~\ref{assm:reward:Lipschitz}, the regret for Algorithm~\ref{algo1} satisfies
	\[R_T\le 3\,\Delta_{\max} \, N\left[\frac{\alpha-1}{\alpha-2}+2\right] + 4.5^{2/3}\, \left( \Delta_{\max} \, N^3\, c^2 \, \left(\sqrt{2\alpha}+2\right)^2 \sigma^2 \log T\right)^{1/3} T^{2/3}\]
\end{proposition}
\noindent\textbf{Proof of Proposition~\ref{prop:CMAB:inst:ind}.}
The proof is identical to Proposition~\ref{prop:MAB:nogap}.
\hfill \Halmos
\subsection{ Replenishing Items, Same Rewards}\label{sec:best:K:CMAB}
In this section, we investigate a variant of our problem in which resources are continuously replenished, ensuring that each agent receives a unique set of resources. Here, we have $N$ items that are replenished after each agent's turn. During a turn, an agent is assigned a specific set of items, which are replenished before the next agent's turn begins. This process repeats until all agents have been served, marking the completion of one time instance. The system operates under two primary rules: ensuring no two agents receive the same set of items and disclosing only the base reward for one agent to the system, reflecting our commitment to active feedback.



The lack of constraints on the selection of super-arms is a direct consequence of the arms' resampling and the departure from considering partitions exclusively.
Therefore the problem reduces to finding

\begin{equation}\begin{aligned}\label{top:K:max:min}
\max_{\phi \in \mathcal{M}}\min_{j \in \mathcal{K}} r(\bm{\mu};\phi(j)).
\end{aligned}\end{equation}

This problem is essentially finding the top $K$ super-arms in the CMAB setting by revealing only one super-arm's rewards. Define any $K$-th best super-arm as $S^*$ and define the top $K$ super-arm set chosen at time $t$ as $\mathcal{S}_t$. In this case, the cumulative regret reduces to
\[R_T=\mathbb{E}\left[\sum_{t=1}^{T}\left(r(\bm{\mu};S^*)-\min_{S\in \mathcal{S}_t} r(\bm{\mu};S)\right)\right].\]
To solve this problem, we can use Algorithm~\ref{algo1} with a minor change. That is, the first oracle, when given a vector $\bm{\mu}=(\mu_1,\mu_2,\cdots,\mu_N)$ and a reward function $r$, solves the problem \eqref{top:K:max:min} and the second oracle is simply a sorting oracle. 
We can now establish the regret bound for Algorithm~\ref{algo1} in this setting.
\begin{corollary}\label{mainthm}
	Let Assumptions~\ref{assm:reward:Lipschitz}-\ref{assm:gap} hold. The regret bound for Algorithm~\ref{algo1} satisfies
	\begin{equation*}\begin{aligned}
	R_T&\le 3\,\tilde{\Delta}_{\max}\,N\left[ \frac{\left(\sqrt{2\alpha}+2\right)^2 c^2 \, N^2\, \sigma^2}{\tilde{\Delta}^2_{\min}}\,  \log T+\frac{\alpha-1}{\alpha-2}+2\right].
	\end{aligned}\end{equation*}
\end{corollary}
\noindent\textbf{Proof of Corollary~\ref{mainthm}.}
The proof follows immediately from Theorem~\ref{order:prop}.
\hfill \Halmos

An interesting case is when the reward is defined as

\begin{equation}\begin{aligned}
r(\bm{\mu};S)=\mathbb{E}\left[ f\left(\sum_{i \in S} X_i\right)\right]
\end{aligned}\end{equation} 
with $f(\cdot)$ being a known function. The main question of interest here is-what conditions on $f$ are sufficient to establish the desired regret bounds.

\begin{proposition}\label{prop:f:regret}
	Let the function $f(\cdot)$ is monotone and $L$-Lipschitz.. Then, for algorithm~\ref{algo1},
	\begin{equation*}\begin{aligned}
	R_T&= O(m^2 N\,\log T).
	\end{aligned}\end{equation*}
\end{proposition}

\noindent\textbf{Proof of Proposition~\ref{prop:f:regret}.}
With some notation abuse we denote $\sum_{i \in S} X_i=S^{T}\mathbf{X}$. Note that for any $\bm{\mu}$ and $\bm{\nu}$, we may take 
$\mathbf{\tilde{X}}(t)=\left(\mathbf{X}(t)-\bm{\mu}\right)+\bm{\nu}$ such that 
$r(\bm{\mu};S)=\mathbb{E}[f(S^{\mathsf{T}}\mathbf{X}]$ and $r(\bm{\nu};S)=\mathbb{E}[f(S^{\mathsf{T}}\mathbf{\tilde{X}}(t))]$.
Therefore, 
\begin{equation*}\begin{aligned}
\left|r(\bm{\mu};S)-r(\bm{\nu};S)\right|&=\left|\mathbb{E}[f(S^{\mathsf{T}}\mathbf{X}(t))]-\mathbb{E}[f(S^{\mathsf{T}}\mathbf{\tilde{X}}(t))]\right|\\
&\le \mathbb{E}\left|f(S^{\mathsf{T}}\mathbf{X}(t))-f(S^{\mathsf{T}}\mathbf{\tilde{X}}(t))\right|\\
&\le L\, \mathbb{E}\left|S^{\mathsf{T}}\mathbf{X}(t)-S^{\mathsf{T}}\mathbf{\tilde{X}}(t)|\right|\\
&=L\, \left|S^{\mathsf{T}}\left(\bm{\mu}-\bm{\nu}\right)\right|\\
&=L\, \left|\bm{\mu}(S)-\bm{\nu}(S)\right|\\
&\le L \sum_{i \in S} \left|\mu_i-\nu_i\right|.
\end{aligned}\end{equation*}
Also, if $\nu_i\ge \mu_i$, for all $i \in S$, then 
\begin{equation*}\begin{aligned}
S^{\mathsf{T}}\mathbf{\tilde{X}}(t)= S^{\mathsf{T}}\mathbf{X}(t)+S^{\mathsf{T}}\left(\mathbf{\mu}^{(2)}-\mathbf{\mu}^{(1)}\right)\ge S^{\mathsf{T}}\mathbf{X}(t).
\end{aligned}\end{equation*}
This implies $f(S^{\mathsf{T}}\mathbf{\tilde{X}}(t))\ge f(S^{\mathsf{T}} \mathbf{X}(t))$ by the monotone property of $f$. Therefore  
\begin{equation*}\begin{aligned}
r(\bm{\nu};S)-r(\bm{\mu};S)=\mathbb{E}\left[f(S^{\mathsf{T}}\mathbf{\tilde{X}}(t))-f(S^{\mathsf{T}}\mathbf{X}(t))\right]\ge 0.
\end{aligned}\end{equation*}
Hence Assumption~\ref{assm:reward:Lipschitz} is established and thus the proof follows.
\hfill \Halmos 

\subsection{Proofs for Minimal Envy Allocation}
In this setting we shall again work in the general regime where $\mathcal{M}=\{\phi: \, \phi: \mathcal{K}\rightarrow 2^{\mathcal{N}}, \ \phi(i)\ne \phi(j)\}$, $r^{j}: \mathbb{R}^{N}\times \mathcal{A}_j\rightarrow \mathbb{R}$ and $\overset{K}{\underset{j=1}{\cup}} \mathcal{A}_j=\mathcal{A}^{m}$. Thus we have a capacity constraint on each set with $|A|\le m$. We establish our results in this regime and then present our results with $m=N$.
\begin{lemma}\label{lemma:envy:conc}
	For any fixed allocation $\phi \in \mathcal{M}$, one has
	
	\begin{equation*}\begin{aligned}
	&\mathbb{P}\left(ev(\underline{\bm{\nu}}(t),\bar{\bm{\nu}}(t),\phi)>ev(\bm{\mu},\phi)\right)
	\le \frac{2\, K\,(K-1) \, N}{t^{\alpha-1}}\\
	&\text{and}\\
	& \mathbb{P}\left(ev(\bar{\bm{\nu}}(t),\underline{\bm{\nu}}(t),\phi)<ev(\bm{\mu},\phi)\right)\le  \frac{2\, K\,(K-1) \, N}{t^{\alpha-1}}.
	\end{aligned}\end{equation*}
\end{lemma}
\noindent\textbf{Proof of Lemma~\ref{lemma:envy:conc}.}
We start by fixing $i,j\in \mathcal{K}$. Note that in the event
\begin{equation*}\begin{aligned}
\left\{ev_{i \to j}(\underline{\bm{\nu}}(t),\bar{\bm{\nu}}(t),\phi)>ev_{i \to j}(\bm{\mu},\phi)\right\}
\end{aligned}\end{equation*}
one of the following must be true-
either 
\[r^i( \underline{\bm{\nu}}(t);\phi(j)) >r^i(\bm{\mu};\phi(j))\]
or 
\[r^i(\bar{\bm{\nu}}(t);\phi(i)) < r^i(\bm{\mu};\phi(i)).\]
This is immediate from the fact that on the event in question $ev_{i \to j}(\underline{\bm{\nu}}(t),\bar{\bm{\nu}}(t),\phi)>0$ since the inequality is strict and hence
\begin{equation*}\begin{aligned}  \left(\left(r^i( \underline{\bm{\nu}}(t);\phi(j))-r^i(\bar{\bm{\nu}}(t);\phi(i))\right)>\left(r^i(\bm{\mu};\phi(j))-r^i(\bm{\mu};\phi(i))\right)\right).
\end{aligned}\end{equation*}
Note that the statement holds true even if the right hand side is less than $0$.
This immediately implies 
\begin{equation*}\begin{aligned}
&\mathbb{P}\left(ev(\underline{\bm{\nu}}(t),\bar{\bm{\nu}}(t),\phi)>ev(\bm{\mu},\phi)\right)\\
&\le \mathbb{P}\left(ev(\underline{\bm{\nu}}(t),\bar{\bm{\nu}}(t),\phi)-ev(\bm{\mu},\phi)>0\right)\\
&\le \mathbb{P}\left(\max_{i,j \in \mathcal{K}}\left(\max \left(r^i(\underline{\bm{\nu}}(t);\phi(j))-r^i(\bar{\bm{\nu}}(t);\phi(i)),0\right)-\max \left(r^i(\bm{\mu};\phi(j))-r^i(\bm{\mu};\phi(i)),0\right)\right)>0\right)\\
&\le K\,(K-1)\, \mathbb{P}\left(\left(\max \left(r^i( \underline{\bm{\nu}}(t);\phi(j))-r^i(\bar{\bm{\nu}}(t);\phi(i)),0\right)-\max \left(r^i(\bm{\mu};\phi(j))-r^i(\bm{\mu};\phi(i)),0\right)\right)>0\right)\\
&\le K\,(K-1)\left[\mathbb{P}\left(r^i( \underline{\bm{\nu}}(t);\phi(j)) >r^i(\bm{\mu};\phi(j))\right)+\mathbb{P}\left(r^i(\bar{\bm{\nu}}(t);\phi(i)) < r^i(\bm{\mu};\phi(i))\right)\right]\\
&\le \frac{2\, K\,(K-1) \, N}{t^{\alpha-1}}
\end{aligned}\end{equation*}
where the third inequality follows from the definition of maximum and the fourth inequality follows from the fact that the event implies the existence of $i, j \in \mathcal{K}, \, i\ne j$ such that the event in question holds. The second last inequality follows from the discussion at the beginning of the proof and the final inequality follows from Lemma~\ref{Lemma:cen:shah}.
The proof for the other tail bound is exactly identical.
\hfill \Halmos

\begin{lemma}\label{trivial:lemma}
	For any $x_t,y_t \in \mathbb{R}^N$ with $t=1,2$, one has 
	
	\begin{equation*}\begin{aligned}
	&\max_{i,j \in \mathcal{K}}\max\left(r^i(x_1;\phi(j))-r^i(y_1;\phi(i)),0\right)-\max_{i,j \in \mathcal{K}}\max\left(r^i(x_2;\phi(j))-r^i(y_2;\phi(i)),0\right)\\
	& \le \left|r^{i^*}(x_1;\phi(j^*))-r^{i^*}(x_2;\phi(j^*))\right|+\left|r^{i^*}(y_1;\phi(i^*))-r^{i^*}(y_2;\phi(i^*))\right|
	\end{aligned}\end{equation*}
	for some $i^*,j^* \in \mathcal{K}$.
\end{lemma}
\noindent\textbf{Proof of Lemma~\ref{trivial:lemma}.}
The proof follows by observing that there exists $i^*,j^*$ such that

\begin{equation*}\begin{aligned}
&\max_{i,j \in \mathcal{K}}\max\left(r^i(x_1;\phi(j))-r^i(y_1;\phi(i)),0\right)-\max_{i,j \in \mathcal{K}}\max\left(r^i(x_2;\phi(j))-r^i(y_2;\phi(i)),0\right)\\
&\le \max\left(r^{i^*}(x_1;\phi(j^*))-r^{i^*}(y_1;\phi(i^*)),0\right)-\max\left(r^{i^*}(x_2;\phi(j^*))-r^{i^*}(y_2;\phi(i^*)),0\right).
\end{aligned}\end{equation*}
Now note that if $r^{i^*}(x_1;\phi(j^*))-r^{i^*}(y_1;\phi(i^*))\le 0$, the last term is less than equal to $0$. Hence the result trivially follows. If $r^{i^*}(x_1;\phi(j^*))-r^{i^*}(y_1;\phi(i^*))> 0$, then 

\begin{equation*}\begin{aligned}
&\max_{i,j \in \mathcal{K}}\max\left(r^i(x_1;\phi(j))-r^i(y_1;\phi(i)),0\right)-\max_{i,j \in \mathcal{K}}\max\left(r^i(x_2;\phi(j))-r^i(y_2;\phi(i)),0\right)\\
&\le \left(r^{i^*}(x_1;\phi(j^*))-r^{i^*}(y_1;\phi(i^*))\right)-\left(r^{i^*}(x_2;\phi(j^*))-r^{i^*}(y_2;\phi(i^*))\right)\\
&\le \left|r^{i^*}(x_1;\phi(j^*))-r^{i^*}(x_2;\phi(j^*))\right|+\left|r^{i^*}(y_1;\phi(i^*))-r^{i^*}(y_2;\phi(i^*))\right|.
\end{aligned}\end{equation*}
Thus we are done.    
\hfill \Halmos

Define 

\begin{equation}\begin{aligned}\label{envy:bad:event}
E_t(\delta)&=\left\{\exists \ i,j,i',j', \,\phi \not \in \mathcal{E}^* : \left(I_t=\phi(i)\cup \phi(j)\right)\cap \right.\\
& \left. \quad \left((i,j)\in \argmax_{i,j\in \mathcal{K}}\left(\max\left(r^i(\bar{\bm{\nu}}(t);\phi(j))-r^i(\underline{\bm{\nu}}(t);\phi(i)),0\right)\right)\right)\right.\\
& \left. \quad \cap \left((i',j')\not \in \argmax_{i',j'\in \mathcal{K}}\left(\max\left(r^{i'}(\bm{\mu};\phi(j'))-r^{i'}(\bm{\mu};\phi(i'),0\right)\right)\right) \right.  \\
&\left. \quad \cap \left(ev_{i \to j}(\bar{\bm{\nu}}(t),\underline{\bm{\nu}}(t),\phi)>ev_{i' \to j'}(\bar{\bm{\nu}}(t),\underline{\bm{\nu}}(t),\phi)\right)\right.\\
&\left. \quad \cap \left(ev_{i \to j}(\bm{\mu},\phi)+\delta<ev_{i' \to j'}(\bm{\mu},\phi)\right) \right\}
\end{aligned}\end{equation}
as the event that for some allocation $\phi \in \mathcal{M}$ the maximal envy is not chosen by the upper estimate at time $t$. We show that the total size of this event is controlled.
\begin{proposition}\label{prop:envy}
	Under Assumption~\ref{assm:reward:Lipschitz}, one has 
	
	\begin{equation*}\begin{aligned}
	\mathbb{E}\left[\sum_{t=1}^{T}\mathbf{1}(E_t(\delta))\right]
	&=O\left(\frac{c^2\, m^2\, N\log T}{\delta^2}\right).
	\end{aligned}\end{equation*}
\end{proposition}
\noindent\textbf{Proof of Proposition~\ref{prop:envy}.} We establish the result by reducing this problem to a MAB problem based on the arm explored. The remaining event can be reduced to that of low probability. Define the event 
\[A^{(3)}_{\phi}(t)=\left( ev_{i' \to j'}(\bar{\bm{\nu}}(t),\underline{\bm{\nu}}(t),\phi)
\ge  ev_{i' \to j'}(\bm{\mu},\phi)\right).\]
We note that 
\[\mathbb{E}\left[\sum_{t=1}^{T}\mathbf{1}(E_t(\delta))\right] \le \mathbb{E}\left[\sum_{t=1}^{T}\mathbf{1}(E_t(\delta)\cap A^{(3)}_{\phi}(t))\right] + \mathbb{E}\left[\sum_{t=1}^{T}\mathbf{1}(A^{(3)}_{\phi}(t)^c)\right].\]
Note that on the event $E_t(\delta)\cap A^{(3)}_{\phi}(t)$, 

\begin{equation*}\begin{aligned}
& ev_{i \to j}(\bar{\bm{\nu}}(t),\underline{\bm{\nu}}(t),\phi)-ev_{i \to j}(\bm{\mu},\phi)\\
&\overset{(1)}{\ge} ev_{i' \to j'}(\bar{\bm{\nu}}(t),\underline{\bm{\nu}}(t),\phi)-ev_{i \to j}(\bm{\mu},\phi)\\
&\overset{(2)}{\ge} ev_{i' \to j'}(\bm{\mu},\phi)-ev_{i \to j}(\bm{\mu},\phi).
\end{aligned}\end{equation*}
where $(1)$ and $(2)$ follow from the event in question. The last expression is greater than $\delta$. Therefore, it is easy to see that 
\begin{equation*}\begin{aligned}
& ev_{i \to j}(\bar{\bm{\nu}}(t),\underline{\bm{\nu}}(t),\phi)-ev_{i \to j}(\bm{\mu},\phi)\ge \delta,
\end{aligned}\end{equation*}
which implies 
\begin{equation*}\begin{aligned}
\left(r^{i}(\bar{\bm{\nu}}(t);\phi(j))-r^{i}(\underline{\bm{\nu}}(t);\phi(i))\right)-\left(r^{i}(\bm{\mu};\phi(j))-r^{i}(\bm{\mu};\phi(i))\right) \ge \delta.
\end{aligned}\end{equation*}
A brief explanation of this is as $\delta>0$, the first term must be positive and the second term being negative only increases the expression in value.
This in turn implies, 

\begin{equation*}\begin{aligned}
\left|r^{i}(\bar{\bm{\nu}}(t);\phi(j))-r^{i}(\bm{\mu};\phi(j))\right|+\left|r^{i}(\bm{\mu};\phi(i))-r^{i}(\underline{\bm{\nu}}(t);\phi(i))\right| &\ge \delta;\\ 
\text{which implies    }\quad \quad c \sum_{l\in \phi(j)}\left|\bar{\nu}_{l}(t)-\mu_l\right|+c\sum_{l \in \phi(i)} \left|\mu_l - \underline{\nu}_l(t)\right| &\ge   \delta
\end{aligned}\end{equation*}
using Assumption~\ref{assm:reward:Lipschitz}.
Therefore, by the property of average, there exists $l_1\in \phi(i) , \, l_2 \in \phi(j)$ such that
$|\phi(i)| \left|\bar{\nu}_{l_1}(t)-\mu_{l_1}\right|+|\phi(j)|\left|\mu_{l_2} - \underline{\nu}_{l_2}(t)\right| \ge   \frac{\delta}{c}$
which implies $\left|\bar{\nu}_{l_1}(t)-\mu_{l_1}\right|+\left|\mu_{l_2} - \underline{\nu}_{l_2}(t)\right| \ge   \frac{\delta}{c \, m}$
since $|\phi(i)|, \, |\phi(j)|\le m$.
Hence 

\begin{equation*}\begin{aligned}
\mathbb{E}\left[\sum_{t=1}^{T}\mathbf{1}(E_t(\delta)\cap A^{(3)}_{\phi}(t))\right]
\le \mathbb{E}\left[\sum_{t=1}^{T}\mathbf{1}\left\{\exists \ (i,j) \in \mathcal{N}\times\mathcal{N} : \left(I_t=(i,j)\right)\cap\left(        \left|\bar{\nu}_i(t)-\mu_{i}\right|+\left|\mu_{j} - \underline{\nu}_j(t)\right| \ge   \frac{\delta}{c \, m}\right)\right\}\right].
\end{aligned}\end{equation*}
This can be further bounded by 
\begin{equation*}\begin{aligned}  &\mathbb{E}\left[\sum_{t=1}^{T}\mathbf{1}\left\{\exists \ i \in \mathcal{N} : \left(I_t=i\right)\cap\left(        \left|\bar{\nu}_i(t)-\mu_{i}\right| \ge   \frac{\delta}{2\,c \, m}\right)\right\}\right]\\
&\quad +\mathbb{E}\left[\sum_{t=1}^{T}\mathbf{1}\left\{\exists \ j \in \mathcal{N} : \left(I_t=j\right)\cap\left(       \left|\mu_{j} - \underline{\nu}_j(t)\right| \ge   \frac{\delta}{2\,c \, m}\right)\right\}\right].
\end{aligned}\end{equation*}
Therefore, 
\begin{equation*}\begin{aligned}
&\mathbb{E}\left[\sum_{t=1}^{T} \mathbf{1}(E_t(\delta)) \right]\\
& \quad \le \mathbb{E}\left[\sum_{t=1}^{T}\mathbf{1}(E_t(\delta)\cap A^{(3)}_{\phi}(t))\right] + \mathbb{E}\left[\sum_{t=1}^{T}\mathbf{1}(A^{(3)}_{\phi}(t)^c)\right]\\
&\quad \le \underset{\text{(I)}}{\underbrace{\mathbb{E}\left[\sum_{t=1}^{T}\mathbf{1}\left\{\exists \ i \in \mathcal{N} : \left(I_t=i\right)\cap\left(        \left|\bar{\nu}_i(t)-\mu_{i}\right| \ge   \frac{\delta}{2\,c \, m}\right)\right\}\right]}}\\
&\quad +\underset{\text{(II)}}{\underbrace{\mathbb{E}\left[\sum_{t=1}^{T}\mathbf{1}\left\{\exists \ j \in \mathcal{N} : \left(I_t=j\right)\cap\left(       \left|\mu_{j} - \underline{\nu}_j(t)\right| \ge   \frac{\delta}{2\,c \, m}\right)\right\}\right]}}\\
& \quad \quad + \underset{\text{(III)}}{\underbrace{\sum_{t=1}^{T} \mathbb{P}\left(ev_{i' \to j'}(\bar{\bm{\nu}}(t),\underline{\bm{\nu}}(t),\phi)
		< ev_{i' \to j'}(\bm{\mu},\phi)\right)}}
\end{aligned}\end{equation*}
Note for (III), we have
\begin{equation*}
\begin{aligned}
& \sum_{t=1}^{T} \mathbb{P}\left(ev_{i' \to j'}(\bar{\bm{\nu}}(t),\underline{\bm{\nu}}(t),\phi)
< ev_{i' \to j'}(\bm{\mu},\phi)\right)\\ &= \sum_{t=1}^{T} \mathbb{P}\left(\max\left(r^{i'}(\bar{\bm{\nu}}(t), \phi(j'))-r^{i'}(\underline{\bm{\nu}}(t), \phi(i')),0\right)
< \max\left(r^{i'}(\bm{\mu}, \phi(j'))-r^{i'}(\bm{\mu}, \phi(i')),0\right)\right)\\
&\le \sum_{t=1}^{T} \mathbb{P}\left(r^{i'}(\bar{\bm{\nu}}(t), \phi(j'))<r^{i'}(\bm{\mu}, \phi(j'))\right)+\sum_{t=1}^{T} \mathbb{P}\left(r^{i'}(\underline{\bm{\nu}}(t), \phi(i'))>r^{i'}(\bm{\mu}, \phi(i'))\right)\\
&\le \sum_{t=1}^{T}\mathbb{P}\left(\exists \, i \in \mathcal{N}: \, \bar{\nu}_i(t)\le \mu_i\right)+\sum_{t=1}^{T}\mathbb{P}\left(\exists \, i \in \mathcal{N}: \, \underline{\nu}_i(t)\ge \mu_i\right)\\
&\le \frac{2\, N \, (\alpha-1)}{\alpha-2}.
\end{aligned}
\end{equation*}
Using Corollary~\ref{main:coro:suffpull} for (I) and (II), and Lemma~\ref{lemma:envy:conc} for the last term, we get the final bound as 
\begin{equation*}\begin{aligned}
\frac{\left(\sqrt{2\alpha}+2\right)^2\,8\, c^2\, m^2\, N \log T}{\delta^2}+4\, N +\frac{2\,N\, (\alpha-1)}{\alpha-2}
\end{aligned}\end{equation*}
Hence we are done. Note that when $m=N$, this bound is $O(N^3\log T \delta^{-2})$.
\hfill \Halmos 

\noindent\textbf{Proof of Theorem~\ref{envy:main:thm}.}The key idea in this proof is to leverage Proposition~\ref{prop:envy} and reduce the problem to the base arms except in a region of low probability. Define, for any $\phi^* \in \mathcal{E}^*$, the event 
$$A^{(4)}_{\phi}(t)=\left\{ev(\underline{\bm{\nu}}(t),\bar{\bm{\nu}}(t),\phi^*) \le ev(\bm{\mu},\phi^*)\right\}$$
and the event 

\begin{equation*}
\begin{aligned}
\text{Env}_t &=\left\{\exists \ i,j \in \mathcal{K}, \phi_1 \in \mathcal{M}\backslash \mathcal{E}^*, \, \phi_2 \in \mathcal{E}^*  : \left(I_t=\phi_1(i)\cup \phi_1(j))\right)\right.\\
&\left.\quad \quad \quad \quad \quad \quad \quad \quad \quad \quad \quad \cap \left(ev(\underline{\bm{\nu}}(t),\bar{\bm{\nu}}(t),\phi_1)\le ev(\underline{\bm{\nu}}(t),\bar{\bm{\nu}}(t),\phi_2)\right)\cap \left(ev(\bm{\mu},\phi_1)> ev(\bm{\mu},\phi_2)\right)\right\}
\end{aligned}
\end{equation*}
which is the event that the optimal envy allocation is not chosen and explored at time $t$.
Using Lemma~\ref{trivial:lemma}, note that,
\begin{equation*}\begin{aligned}   R_T&=\mathbb{E}\left[\sum_{t=1}^{T} \left(ev(\bm{\mu},\phi_t)-ev(\bm{\mu},\phi^*)\right)\right]\\
&\le \Delta_{e,\max}\,\mathbb{E}\left[\sum_{t=1}^{T} \mathbf{1}\left(\phi_t\ne \phi^*\right)\right]\\
&\le  \Delta_{e,\max}\,\mathbb{E}\left[\sum_{t=1}^{T} \mathbf{1}\left\{\text{Env}_t\right\}\right]\\
& \le \Delta_{e,\max}\,\mathbb{E}\left[\sum_{t=1}^{T} \mathbf{1}\left\{\text{Env}_t \cap A^{(4)}_{\phi}(t)\right\}\right]+ \Delta_{e,\max}\,\mathbb{E}\left[\sum_{t=1}^{T} \mathbf{1}\left\{A^{(4)}_{\phi}(t)^c\right\}\right].
\end{aligned}\end{equation*}
Invoking Lemma~\ref{lemma:envy:conc}, we know that 
\[\mathbb{E}\left[\sum_{t=1}^{T} \mathbf{1}\left\{A^{(4)}_{\phi}(t)^c\right\}\right]\le \sum_{t=1}^{T} \frac{2\,K(K-1)\, N}{t^{\alpha-1}}.\]
Further, on the event $\text{Env}_t \cap A^{(4)}_{\phi}(t)$, one has 
\begin{equation*}\begin{aligned}
& ev(\bm{\mu},\phi_1) - ev(\underline{\bm{\nu}}(t),\bar{\bm{\nu}}(t),\phi_2)\\
&\overset{(1)}{\ge}  ev(\bm{\mu},\phi_1) - ev(\underline{\bm{\nu}}(t),\bar{\bm{\nu}}(t),\phi_2)\\
& \overset{(2)}{\ge} ev(\bm{\mu},\phi_1) - ev(\bm{\mu},\phi_2)\\
& \overset{(3)}{\ge}  \Delta_{e,\min}>0
\end{aligned}\end{equation*}
where $(1)$ and $(2)$ hold because of the event in consideration and $(3)$ folds due to Assumption~\ref{envy:assm:gap}.  
Note that \[(i,j)=\argmax_{i_1,j_1 \in \mathcal{K}}\left(\max\left(r^{i_1}(\bar{\bm{\nu}}(t);\phi_1(i_1))-r^{i_1}(\underline{\bm{\nu}}(t);\phi_1(j_1)),0\right)\right)\] 
by property of the first oracle.
Define the event 
\begin{equation*}
\begin{aligned}
\text{Err}_t=\{(i,j)\in \argmax_{i_1,j_1 \in \mathcal{K}}\left(\max\left(r^{i_1}(\bm\mu; \phi_2(j_1))-r^{i_1}(\bm \mu;\phi_2(i_1)),0\right)\right)\}.
\end{aligned}
\end{equation*}
Further, without loss of generality, define 
$(i',j')=\argmax_{i_1,j_1 \in \mathcal{K}}\left(\max\left(r^{i_1}(\bm\mu; \phi_2(j_1))-r^{i_1}(\bm \mu;\phi_2(i_1)),0\right)\right)$. Note that it does not actually matter what pair we draw as the envy value is same.
Note that 
\[\mathbb{E}\left[\sum_{t=1}^{T} \mathbf{1}\left\{\text{Env}_t \cap A^{(4)}_{\phi}(t)\right\}\right]\le \mathbb{E}\left[\sum_{t=1}^{T} \mathbf{1}\left\{\text{Env}_t \cap A^{(4)}_{\phi}(t)\cap\text{Err}_t \right\}\right]+\mathbb{E}\left[\sum_{t=1}^{T} \mathbf{1}\left\{\text{Env}_t \cap A^{(4)}_{\phi}(t)\cap\text{Err}^c_t \right\}\right]. \]  
On $\text{Env}_t \cap A^{(4)}_{\phi}(t)\cap\text{Err}_t$, we have 
\begin{equation*}\begin{aligned}
&ev_{i\to j}(\bm{\mu},\phi_1) - ev_{i \to j}(\underline{\bm{\nu}}(t),\bar{\bm{\nu}}(t),\phi_1)\\
&=ev(\bm{\mu},\phi_1) - ev(\underline{\bm{\nu}}(t),\bar{\bm{\nu}}(t),\phi_1)\\
& \ge \Delta_{e,\min}>0.
\end{aligned}\end{equation*} 
Again, we see that 
\begin{equation*}\begin{aligned}
& ev_{i\to j}(\bm{\mu},\phi_1) - ev_{i \to j}(\underline{\bm{\nu}}(t),\bar{\bm{\nu}}(t),\phi_1)\ge \Delta_{e,\min}
\end{aligned}\end{equation*}
implies 
\begin{equation*}\begin{aligned}
\left(r^{i}(\bm{\mu};\phi_1(j))-r^{i}(\bm{\mu};\phi_1(i))\right)-\left(r^{i}(\underline{\bm{\nu}}(t);\phi_1(j))-r^{i}(\bar{\bm{\nu}}(t);\phi_1(i))\right) \ge \Delta_{e,\min}
\end{aligned}\end{equation*}
as $\Delta_{e,\min}>0$.
This in turn implies, 

\begin{equation*}\begin{aligned}
\left|r^{i}(\bar{\bm{\nu}}(t);\phi_1(i))-r^{i}(\bm{\mu};\phi_1(i))\right|+\left|r^{i}(\bm{\mu};\phi_1(j))-r^{i}(\underline{\bm{\nu}}(t);\phi_1(j))\right| &\ge \Delta_{e,\min};\\ 
\text{which implies    }\quad \quad c \sum_{l\in \phi_1(j)}\left|\bar{\nu}_{l}(t)-\mu_l\right|+c\sum_{l \in \phi_1(i)} \left|\mu_l - \underline{\nu}_l(t)\right| &\ge   \Delta_{e,\min}
\end{aligned}\end{equation*}
using Assumption~\ref{assm:reward:Lipschitz}.
Therefore, there exists $l_1\in \phi_1(i) , \, l_2 \in \phi_1(j)$ such that
\begin{equation*}\begin{aligned}
\left|\bar{\nu}_{l_1}(t)-\mu_{l_1}\right|+\left|\mu_{l_2} - \underline{\nu}_{l_2}(t)\right| &\ge   \frac{\Delta_{e,\min}}{c \, m}.
\end{aligned}\end{equation*} 
On $\text{Env}_t \cap A^{(4)}_{\phi}(t)\cap\text{Err}^c_t $ we have
one of the following events occur-
\begin{equation*}
\begin{aligned}
&ev_{i \rightarrow j }(\mu, \phi_1)-ev(\mu, \phi_2) \ge  \frac{\Delta_{e,\min}}{2}\\
& \text{or}\\
& ev(\mu, \phi_1)-ev_{i \rightarrow j}(\mu,\phi_1) \ge \frac{\Delta_{e,\min}}{2}
\end{aligned}
\end{equation*}
Therefore on 
\[\text{Env}_t \cap A^{(4)}_{\phi}(t)\cap\text{Err}^c_t \cap \left(ev_{i \rightarrow j }(\mu; \phi_1)-ev(\mu; \phi_2) \ge  \frac{\Delta_{e,\min}}{2}\right)\cap\left(ev(\underline{\bm{\nu}}(t),\bar{\bm{\nu}}(t), \phi_2)\le  ev(\mu, \phi_2)\right)\]
we have 
\begin{equation*}
\begin{aligned}
& ev_{i \rightarrow j}(\mu, \phi_1)-ev_{i \rightarrow j}(\underline{\bm{\nu}}(t),\bar{\bm{\nu}}(t), \phi_1)\\
& ev_{i \rightarrow j}(\mu, \phi_1)-ev(\underline{\bm{\nu}}(t),\bar{\bm{\nu}}(t), \phi_1)\\
&  ev_{i \rightarrow j}(\mu, \phi_1)-ev(\underline{\bm{\nu}}(t),\bar{\bm{\nu}}(t), \phi_2)\\
& ev_{i \rightarrow j}(\mu, \phi_1)-ev(\mu, \phi_2)\ge \frac{\Delta_{e,\min}}{2}.
\end{aligned}
\end{equation*}
Note that this further implies that there exists 
there exists $l_1\in \phi_1(i) , \, l_2 \in \phi_1(j)$ such that
\begin{equation*}\begin{aligned}
\left|\bar{\nu}_{l_1}(t)-\mu_{l_1}\right|+\left|\mu_{l_2} - \underline{\nu}_{l_2}(t)\right| &\ge   \frac{\Delta_{e,\min}}{2\,c \, m}.
\end{aligned}\end{equation*} 

Also note that using Lemma~\ref{lemma:envy:conc}, we know that 
\[\mathbb{P}\left(ev(\underline{\bm{\nu}}(t),\bar{\bm{\nu}}(t), \phi_2)>  ev(\mu, \phi_2)\right) \le \frac{2\, K(K-1)\, N}{t^{\alpha-1}}.\]
Therefore note that
\begin{equation*}
\begin{aligned}
&\mathbb{E}\left[\sum_{t=1}^{T} \, \mathbf{1}\left\{\text{Env}_t \cap A^{(4)}_{\phi}(t)\cap\text{Err}^c_t \cap \left(ev_{i \rightarrow j }(\mu; \phi_1)-ev(\mu; \phi_2) \ge  \frac{\Delta_{e,\min}}{2}\right)\right\}\right]\\
&\le \mathbb{E}\left[\sum_{t=1}^{T} \, \mathbf{1}\left\{\text{Env}_t \cap A^{(4)}_{\phi}(t)\cap\text{Err}^c_t \cap \left(ev_{i \rightarrow j }(\mu; \phi_1)-ev(\mu; \phi_2) \ge  \frac{\Delta_{e,\min}}{2}\right)\cap\left(ev(\underline{\bm{\nu}}(t),\bar{\bm{\nu}}(t), \phi_2)\le  ev(\mu, \phi_2)\right)\right\}\right]\\
&\quad +\sum_{t=1}^{T} \frac{2\, K(K-1)\, N}{t^{\alpha-1}}\\
&\le \mathbb{E}\left[\mathbf{1}\left\{\exists \, (l_1,l_2) \in \mathcal{N}\times \mathcal{N}: \, \left|\bar{\nu}_{l_1}(t)-\mu_{l_1}\right|+\left|\mu_{l_2} - \underline{\nu}_{l_2}(t)\right| \ge   \frac{\Delta_{e,\min}}{2\,c \, m}\right\}\right]+\frac{2\, K\, (K-1)\, N\, (\alpha-1)}{\alpha-2}.
\end{aligned}
\end{equation*}

Finally on the set 
\[\text{Env}_t \cap A^{(4)}_{\phi}(t)\cap\text{Err}^c_t \cap \left(ev(\mu, \phi_1)-ev_{i \rightarrow j}(\mu,\phi_1) \ge \frac{\Delta_{e,\min}}{2}\right)\]
we have 
\begin{equation*}\begin{aligned}
ev_{i\to j}(\bar{\bm{\nu}}(t),\underline{\bm{\nu}}(t),\phi_1) &\ge ev_{i'\to j'}(\bar{\bm{\nu}}(t),\underline{\bm{\nu}}(t),\phi_1)\\
\text{but} &\\
ev_{i\to j}(\bm \mu,\phi_1) + \frac{\Delta_{e,\min}}{2} &< ev_{i'\to j'}(\bm \mu,\phi_1)
\end{aligned}\end{equation*}
since the maximal envy is not selected. This is simply the event $E_t(\Delta_{e,\min}/2)$ as defined in \eqref{envy:bad:event}.
Therefore,
\begin{equation*}
\begin{aligned}
&\mathbb{E}\left[\sum_{t=1}^{T} \mathbf{1}\left\{\text{Env}_t \cap A^{(4)}_{\phi}(t)\cap\text{Err}^c_t \right\}\right]\\
&\le  \sum_{t=1}^{T}\mathbb{E}\left[\mathbf{1}(E_t(\Delta_{e,\min}/2))\right]  +\mathbb{E}\left[\mathbf{1}\left\{\exists \, (l_1,l_2) \in \mathcal{N}\times \mathcal{N}: \, \left|\bar{\nu}_{l_1}(t)-\mu_{l_1}\right|+\left|\mu_{l_2} - \underline{\nu}_{l_2}(t)\right| \ge   \frac{\Delta_{e,\min}}{2\, c \, m}\right\}\right]\\
&+\frac{K\, (K-1)\, N\, (\alpha-1)}{\alpha-2}.
\end{aligned}
\end{equation*}

which further implies 
\begin{equation*}\begin{aligned}
R_T &\le \Delta_{e,\max} \, \sum_{t=1}^{T}\mathbb{E}\left[\mathbf{1}(E_t)\right]\\
&\quad + \Delta_{e,\max} \, \mathbb{E}\left[\mathbf{1}\left\{\exists \, (l_1,l_2) \in \mathcal{N}\times \mathcal{N}: \, \left|\bar{\nu}_{l_1}(t)-\mu_{l_1}\right|+\left|\mu_{l_2} - \underline{\nu}_{l_2}(t)\right| \ge   \frac{\Delta_{e,\min}}{c \, m}\right\}\right]\\
&\quad \quad +\Delta_{e,\max} \,\mathbb{E}\left[\mathbf{1}\left\{\exists \, (l_1,l_2) \in \mathcal{N}\times \mathcal{N}: \, \left|\bar{\nu}_{l_1}(t)-\mu_{l_1}\right|+\left|\mu_{l_2} - \underline{\nu}_{l_2}(t)\right| \ge   \frac{\Delta_{e,\min}}{2\, c \, m}\right\}\right]\\
& \quad + \Delta_{e,\max}\, \frac{4\, K\, (K-1)\, N\, (\alpha-1)}{\alpha -2}.
\end{aligned}\end{equation*}
Thus using Proposition~\ref{prop:envy} for the first term and Corollary~\ref{main:coro:suffpull} for the second term, we obtain the final bound as 
\begin{equation*}\begin{aligned}
\frac{\left(\sqrt{2\alpha}+2\right)^2\,13\, \Delta_{e,\max}\, c^2\, m^2\, N \log T}{\Delta^2_{e,\min}}+6\, \Delta_{e,\max}\, N +\frac{6\, \Delta_{e,\max} \,K\,(K-1)\, N\,(\alpha-1)}{\alpha-2}.
\end{aligned}\end{equation*}
Hence the proof is concluded.
\hfill \Halmos 

Using Theorem~\ref{envy:main:thm}, we may also obtain an instance independent regret bound for Algorithm~\ref{algo:envy:cucblcb} based on whether the envy of the chosen allocation is close to the optimal envy or not.
\begin{proposition}\label{envy:gap:indep}
	Let for any $\phi_1,\phi_2$ and any pairs $(i,j) \ne (i',j') \in \mathcal{K}$, there exists $\Delta_{e,\max}>0$ such that
	\begin{equation*}
	\left|ev_{i\to j}(\bm{\mu},\phi_1)- ev_{i'\rightarrow j'}(\bm{\mu};\phi_2) \right| \le \Delta_{e,\max}
	\end{equation*}  
	hold. Then, under Assumption~\ref{assm:reward:Lipschitz}, the regret for Algorithm~\ref{algo:envy:cucblcb} satisfies 
	\begin{equation*}
	\begin{aligned}
	R_T=O(N\, T^{2/3} (\log T)^{1/2}).
	\end{aligned}
	\end{equation*}
	
\end{proposition} 
\textbf{Proof of Proposition~\ref{envy:gap:indep}.}
Define \[G(\delta)=\left\{\phi \, : \, ev(\bm{\mu},\phi)-\min_{\phi' \in \mathcal{E}^*} ev(\bm{\mu}, \phi')\ge \delta\right\}.\]
Therefore note that for the regret of Algorithm~\ref{algo:envy:cucblcb} as defined in \eqref{envy:regret}, we have
\begin{equation*}
\begin{aligned}
R_T=\mathbb{E}\left[\sum_{t=1}^{T} \left(ev(\bm{\mu},\phi_t)-ev(\bm{\mu},\phi^*)\right) \, \mathbf{1}\left\{\phi_t \in G(\Delta_T)^c\right\}\right] +\mathbb{E}\left[\sum_{t=1}^{T} \left(ev(\bm{\mu},\phi_t)-ev(\bm{\mu},\phi^*)\right) \, \mathbf{1}\left\{\phi_t \in G(\Delta_T)\right\}\right]
\end{aligned}
\end{equation*}
Note that the first term is bounded by $\Delta_T \, T$. Using Theorem~\ref{envy:main:thm}, we know that 
\begin{equation*}
\begin{aligned}
R_T& \le \Delta_T \, T+  \frac{\left(\sqrt{2\alpha}+2\right)^2\,13\, \Delta_{e,\max}\, c^2\, m^2\, N \log T}{\Delta^2_T}+6\, \Delta_{e,\max}\, N +\frac{6\, \Delta_{e,\max} \,K\,(K-1)\, N\,(\alpha-1)}{\alpha-2}\\
& \le \frac{3}{2}\left(\left(\sqrt{2\alpha}+2\right)^2\,13\, \Delta_{e,\max}\, c^2\, m^2\, N \log T\right)^{1/3}\, T^{2/3} +6\, \Delta_{e,\max}\, N +\frac{6\, \Delta_{e,\max} \,K\,(K-1)\, N\,(\alpha-1)}{\alpha-2}
\end{aligned}
\end{equation*}
where the last step follows via optimizing on $\Delta_T$. Hence we are done.
\hfill \Halmos 

\section{Proofs for Stable Allocations}\label{sec:proof-general}
In this section we prove the main results in Section~\ref{sec:stable}.
\begin{lemma}\label{g:L:lemma}
	Let Assumption~\ref{assm:reward:Lipschitz} hold. Then for any $L \subset \mathcal{K}$  and $\phi \in \mathcal{M}$ with $\phi' \in \mathcal{M}_{\phi,L}$, we have $g^{L}(\mathbf{x},\mathbf{y},\phi\to \phi')$ as monotone increasing in $\mathbf{x}$, monotone decreasing in $\mathbf{y}$ where $\mathbf{x}=(x_1,x_2,\cdots,x_N) \in \mathbb{R}^{N}$ and $\mathbf{y}=(y_1,y_2,\cdots,y_N) \in \mathbb{R}^{N}$. Further
	
	\begin{equation*}
	\begin{aligned}
	&\left| g^{L}(\mathbf x_1,\mathbf y_1,\phi \to \phi') -g^{L}(\mathbf x_2,\mathbf y_2,\phi \to \phi')\right|\\ 
	& \le \sum_{j \in L} \left(\left|r^{j}(\mathbf x_1, \phi(j))-r^{j}(\mathbf x_2, \phi(j))\right|+ \left|r^{j}(\mathbf y_1, \phi'(j))-r^{j}(\mathbf y_2, \phi'(j))\right|\right).
	\end{aligned}
	\end{equation*}
\end{lemma}
\textbf{Proof of Lemma~\ref{g:L:lemma}.}
By definition we know that
\[g^{L}(\mathbf x,\mathbf y,\phi \to \phi')=\max_{j \in L}r^{j}(\mathbf x, \phi(j))-r^{j}(\mathbf y, \phi'(j)).\]
Therefore for any $\mathbf{z}_x\ge \mathbf{x}$ (coordinate-wise) we have $r^{j}(\mathbf x, \phi(j)) \le r^{j}(\mathbf{z}_x, \phi(j))$ for each $j \in L$. Therefore is is increasing in the first coordinate since maximum is a monotone function. Similarly, for any $\mathbf{z}_y\ge \mathbf{y}$ (coordinate-wise), we have $r^{j}(\mathbf y, \phi(j)) \le r^{j}(\mathbf{z}_y, \phi(j))$ for each $j \in L$ which implies it is decreasing in the second coordinate. Finally, for any $\mathbf{x}_1, \mathbf{x}_2, \mathbf{y}_1, \mathbf{y}_2 \in \mathbb{R}^N$, one has 

\begin{equation*}
\begin{aligned}
&\left| g^{L}(\mathbf x_1,\mathbf y_1,\phi \to \phi') -g^{L}(\mathbf x_2,\mathbf y_2,\phi \to \phi')\right|\\ &=\left|\max_{j \in L}\left(r^{j}(\mathbf x_1, \phi(j))-r^{j}(\mathbf y_1, \phi'(j))\right)-\max_{j \in L}\left(r^{j}(\mathbf x_2, \phi(j))-r^{j}(\mathbf y_2, \phi'(j))\right)\right| \\
& \le \left|\max \left[\max_{j \in L}\left(r^{j}(\mathbf x_1, \phi(j))-r^{j}(\mathbf y_1, \phi'(j))-r^{j}(\mathbf x_2, \phi(j))+r^{j}(\mathbf y_2, \phi'(j))\right), \right. \right.\\
& \quad \quad \quad \left. \left. \max_{j \in L}\left(r^{j}(\mathbf x_2, \phi(j))-r^{j}(\mathbf y_2, \phi'(j))-r^{j}(\mathbf x_1, \phi(j))+r^{j}(\mathbf y_1, \phi'(j))\right)\right]\right|\\
& \le \max_{j \in L} \left(\left|r^{j}(\mathbf x_1, \phi(j))-r^{j}(\mathbf x_2, \phi(j))\right|+ \left|r^{j}(\mathbf y_1, \phi'(j))-r^{j}(\mathbf y_2, \phi'(j))\right|\right)\\
& \le \sum_{j \in L} \left(\left|r^{j}(\mathbf x_1, \phi(j))-r^{j}(\mathbf x_2, \phi(j))\right|+ \left|r^{j}(\mathbf y_1, \phi'(j))-r^{j}(\mathbf y_2, \phi'(j))\right|\right).
\end{aligned}
\end{equation*}
Hence we are done.\hfill \Halmos

\begin{lemma}\label{approx:lemma:C}
	Under Assumption~\ref{assm:feas:sep}, one has
	
	\begin{equation*}\begin{aligned}
	&\mathbb{E}_{H_a}\left[\sum_{t=1}^{T} \mathbf{1}\left\{\delta^{\epsilon}_t=0, \phi_t \not \in \mathbb{F}(\bm{\mu},\eta)\right\}\right]\\
	& \quad \quad\le 8\, N\, \kappa^2 \, m^2\left(\sqrt{2\alpha}+\hat{\Delta}\right)^2
	\, c^2\, \sigma^2 \, \left(\Delta^{-2}_{\mathcal{M}^*,q}+(\eta-\epsilon)^{-2}\right)\, \log T +\frac{8\, N}{T^{\hat{\Delta}^2/2-2}}+\frac{2\, N \, (\alpha-1)}{\alpha-2} .
	\end{aligned}\end{equation*}
\end{lemma}
\noindent\textbf{Proof.}
The main idea of the proof is to reduce the problem to an MAB setting except a region of low probability. The key is to divide the problem into multiple events which allow us in reducing each case to an MAB setting.

Note that the event $\left\{\delta^{\epsilon}_t=0, \phi_t \not \in \mathbb{F}(\bm{\mu},\eta)\right\}$ implies that for all $\phi \in \mathcal{M}$, there exists $L\subset K$ and $\phi'\in \mathcal{M}|_{\phi,L}$ such that $g^{L}(\underline{\bm{\nu}}(t);\bar{\bm \nu}(t); \phi \to \phi')<\epsilon$. In addition, we have one of the following hold based on the matching returned by Algorithm~\ref{algo:gen:problem}- for all matching in $\mathbb{F}(\bm\mu, \eta)$ there exists $L \subset \mathcal{K}, \, \phi' \in \mathcal{M}|_{\phi,L}$ such that $g^{L}(\bar{\bm \nu}(t), \underline{\bm{\nu}}(t);\phi \to \phi')<\eta$ or there exists a matching $\phi \in \mathbb{F}(\bm\mu,\eta)$ such that for all $(L,\phi')$ we have  $g^{L}(\bar{\bm \nu}(t), \underline{\bm{\nu}}(t);\phi \to \phi')\ge \eta$. That is 

\begin{equation*}
\begin{aligned}
&\left\{\delta^{\epsilon}_t=0, \phi_t \not \in \mathbb{F}(\bm{\mu},\eta)\right\}\\ &\subseteq \left\{\delta^{\epsilon}_t=0, \, \exists\,  \phi \in \mathbb{F}(\bm{\mu},\eta), \, (L,\phi') \text{ such that } g^L(\bar{\bm\nu}(t);\underline{\bm{\nu}}(t);\phi \to \phi')<\eta \right\}\\
& \bigcup \left\{\delta^{\epsilon}_t=0, \, \exists \,  \phi \in \mathcal{M}^*\backslash\mathbb{F}(\bm{\mu},\eta) \, \text{such that }\right.\\
&  \left. \qquad \qquad \qquad g^L(\bar{\bm\nu}(t);\underline{\bm{\nu}}(t);\phi \to \phi') \ge \eta \ \forall \, (L,\phi') \right\}.
\end{aligned}
\end{equation*}

Therefore, we have

\begin{equation*}
\begin{aligned}
&\mathbb{E}_{H_a}\left[\sum_{t=1}^{T} \mathbf{1}\left\{\delta^{\epsilon}_t=0, \phi_t \not \in \mathbb{F}(\bm{\mu},\eta)\right\}\right]\\
& \le \underset{\text{(I)}}{\underbrace{\mathbb{E}_{H_a}\left[\sum_{t=1}^{T} \mathbf{1}\left\{\delta^{\epsilon}_t=0, \, \exists\,  \phi \in \mathbb{F}(\bm{\mu},\eta), \, (L,\phi') \text{ such that } g^L(\bar{\bm\nu}(t);\underline{\bm{\nu}}(t);\phi \to \phi')<\eta \right\}\right]}}\\
&\quad +\underset{\text{(II)}}{\underbrace{\mathbb{E}_{H_a}\left[\sum_{t=1}^{T} \mathbf{1}\left\{\delta^{\epsilon}_t=0, \, \exists \,  \phi \in \mathcal{M}^*\backslash\mathbb{F}(\bm{\mu},\eta) \, \text{such that }
		g^L(\bar{\bm\nu}(t);\underline{\bm{\nu}}(t);\phi \to \phi') \ge \eta \ \forall \, (L,\phi') \right\}\right]}}.
\end{aligned}
\end{equation*}

We first consider (I) where $\phi \in \mathbb{F}(\bm{\mu},\eta)$ is not selected as there exists $(L,\phi')$ such that $g^L(\bar{\nu}(t);\underline{\bm{\nu}}(t);\phi\to\phi')<\eta$. Note that for any $\phi \in \mathbb{F}(\bm\mu,\eta)$, one has $ g^L(\bm \mu;\phi \to \phi') \ge \eta $ for all $(L,\phi')$. Hence (I) is bounded by

\begin{equation*}\begin{aligned}  &\mathbb{E}_{H_a}\left[\sum_{t=1}^{T} \mathbf{1}\left\{\delta^{\epsilon}_t=0, \,  \exists\,  \phi \in \mathbb{F}(\bm{\mu},\eta), \, (L,\phi') \text{ such that } g^L(\bar{\bm\nu}(t);\underline{\bm{\nu}}(t);\phi \to \phi')<\eta \right\}\right]\\
&\le \mathbb{E}\left[\sum_{t=1}^{T} \mathbf{1}\left\{ \exists \ \phi, \, (L,\phi') \,: g^L(\bar{\bm\nu}(t);\underline{\bm{\nu}}(t);\phi \to \phi')<\eta \, , g^L(\bm\mu;\phi \to \phi') \ge \eta \right\}\right]\\
& \overset{(1)}{\le }\mathbb{E}\left[\sum_{t=1}^{T} \mathbf{1}\left\{\exists \, i \in \mathcal{N} \, \text{such that } \mu_i>\bar{\nu}_i(t) \, \text{ or } \mu_i<\underline{\nu}_i(t) \right\}\right]\\
&\le \frac{2\,N\, (\alpha-1)}{\alpha-2}.
\end{aligned}\end{equation*}
Here $(1)$ holds by the definition of $g^{L}$ and the monotonicity of the reward function. An argument for this is as follows- the event in question is equivalent to $\max_{j \in L} r^j(\bar{\bm{\nu}}(t),\phi(j))-r^j(\underline{\bm{\nu}}(t),\phi'(j))\le \max_{j \in L} r^j(\bm{\mu},\phi(j)) -r^j(\bm{\mu},\phi'(j))$ by definition of $g^{L}$. This implies there exists a $j \in \mathcal{K}$ such that the maximum on the right side is realized and for that $j \in \mathcal{K}$, we have
$$ r^j(\bar{\bm{\nu}}(t),\phi(j))-r^j(\underline{\bm{\nu}}(t),\phi'(j))\le  r^j(\bm{\mu},\phi(j)) -r^j(\bm{\mu},\phi'(j)).$$ This further implies that either $r^j(\bar{\bm{\nu}}(t),\phi(j))\le  r^j(\bm{\mu},\phi(j))$ or $r^j(\underline{\bm{\nu}}(t),\phi'(j))> r^j(\bm{\mu},\phi'(j))$. Thus by the monotonicity of $r^{j}$, we get the inequality.
The last step follows from Lemma~\ref{Lemma:cen:shah} and an easy bounding which has been shown in previous proofs.

For (II), we note that as $\phi_t \in \mathcal{M}^*\backslash \ \mathbb{F}(\bm{\mu},\eta)$, there exists $(\bar{L}_t,\bar{\phi}_t')$ such that $g^{\bar{L}_t}(\bm \mu;\phi_t \to \bar{\phi}'_t) <\eta$ but $g^L(\bar{\bm\nu}(t);\underline{\bm{\nu}}(t);\phi_t \to \phi')\ge \eta $ for all $(L,\phi')$. Further note that the exploration is only the sets $\phi_t(j), \phi_t'(j)$ for all $j \in L_t$, for some $L_t$ such that $g^L(\underline{\bm{\nu}}(t);\bar{\bm\nu}(t);\phi \to \phi')<\epsilon$. In the case where for $(\bar{L}_t,\bar{\phi}'_t)\ne (L_t,\phi'_t)$, that is the exploration is not on the correct active sets, then $g^{L_t}(\underline{\bm{\nu}}(t);\bar{\bm\nu}(t);\phi_t \to \phi'_t)<\epsilon$, but $g^{L_t}(\bm \mu;\phi_t \to \phi'_t) \ge \eta$. Hence
for the event in (II), one has 

\begin{equation*}
\begin{aligned}
&\left\{\delta^{\epsilon}_t=0, \, \exists \,  \phi \in \mathcal{M}^*\backslash\mathbb{F}(\bm{\mu},\eta) \, \text{such that } I_t=(L,\phi,\phi'),\right.\\
&  \left. \qquad \qquad \qquad g^L(\bar{\bm\nu}(t);\underline{\bm{\nu}}(t);\phi \to \phi') \ge \eta \ \forall \, (L,\phi')\right\} \\
& \subseteq\left\{\exists \, \phi \in \mathcal{M}^*\backslash \mathbb{F}(\bm{\mu},\eta), \, (L,\phi') : I_t=(L,\phi, \phi')\right.\\
& \left. \qquad \qquad  g^L(\bar{\bm\nu}(t);\underline{\bm{\nu}}(t);\phi \to \phi') \ge \eta \, , \, g^L(\bm\mu;\phi \to \phi')< \eta\right\}\\
&\bigcup \left\{\exists \, \phi \in \mathcal{M}^*\backslash \mathbb{F}(\bm{\mu},\eta), \, (L,\phi') : I_t=(L,\phi, \phi')\right.\\
& \left. \qquad \qquad  g^L(\bm\mu;\phi_t \to \phi'_t) \ge \eta, \, g^L(\underline{\bm{\nu}}(t);\bar{\bm\nu}(t);\phi_t \to \phi'_t) <\epsilon \right\}.
\end{aligned}
\end{equation*}
Thus we have 

\begin{equation*}\begin{aligned}
&\mathbb{E}_{H_a}\left[\sum_{t=1}^{T} \mathbf{1}\left\{\delta^{\epsilon}_t=0, \, \exists \,  \phi \in \mathcal{M}^*\backslash\mathbb{F}(\bm{\mu},\eta) \, \text{such that } I_t=(L,\phi,\phi'),\right.\right.\\
& \left. \left. \qquad \qquad \qquad g^L(\bar{\bm\nu}(t);\underline{\bm{\nu}}(t);\phi \to \phi') \ge \eta \ \forall \, (L,\phi')\right\}\right]\\
& \le \underset{\text{(III)}}{\underbrace{\mathbb{E}_{H_a}\left[\sum_{t=1}^{T} \mathbf{1}\left\{\exists \, \phi \in \mathcal{M}^*\backslash \mathbb{F}(\bm{\mu},\eta), \, (L,\phi') : I_t=(L,\phi, \phi')\,   g^L(\bar{\bm\nu}(t);\underline{\bm{\nu}}(t);\phi \to \phi') \ge \eta \, , \, g^L(\bm\mu;\phi \to \phi')< \eta\right\}\right]}} \\
& +\underset{\text{(IV)}}{\underbrace{\mathbb{E}_{H_a}\left[\sum_{t=1}^{T} \mathbf{1}\left\{\exists \, \phi \in \mathcal{M}^*\backslash \mathbb{F}(\bm{\mu},\eta), \, (L,\phi') : I_t=(L,\phi, \phi') \, g^L(\bm\mu;\phi_t \to \phi'_t)-g^L(\underline{\bm{\nu}}(t);\bar{\bm\nu}(t);\phi_t \to \phi'_t) \ge \eta-\epsilon \right\}\right]}}.
\end{aligned}\end{equation*}

For (III), we have 
\begin{equation*}\begin{aligned}
& \mathbb{E}_{H_a}\left[\sum_{t=1}^{T} \mathbf{1}\left\{\exists \, \phi \in \mathcal{M}^*\backslash \mathbb{F}(\bm{\mu},\eta), \, (L,\phi') : I_t=(L,\phi, \phi')\right.\right.\\
& \left.\left. \qquad \qquad  g^L(\bar{\bm\nu}(t);\underline{\bm{\nu}}(t);\phi \to \phi') \ge \eta \, , \, g^L(\bm\mu;\phi \to \phi')< \eta\right\}\right]\\
& \le \mathbb{E}_{H_a}\left[\sum_{t=1}^{T} \mathbf{1}\left\{\exists \, (L,\phi') \, , \, \phi \in\mathcal{M}^*\backslash\mathbb{F}(\bm \mu, \eta) \,: I_t=(L,\phi, \phi')\right. \right. \\
& \left. \left. \qquad \qquad \, g^L(\bar{\bm\nu}(t);\underline{\bm{\nu}}(t);\phi \to \phi')- g^L(\bm\mu;\phi \to \phi')\ge \eta-g^L(\bm\mu;\phi \to \phi')\right. \right.\\
& \left. \left. \qquad \qquad \text{and } g^L(\bm\mu;\phi \to \phi')< \eta\right\}\right]\\
& \le \mathbb{E}_{H_a}\left[\sum_{t=1}^{T} \mathbf{1}\left\{\exists \, (L,\phi') \, , \, \phi \in\mathcal{M}^*\backslash\mathbb{F}(\bm \mu, \eta) \, :I_t=(L,\phi, \phi')\right. \right. \\
& \left. \left. \qquad \qquad g^L(\bar{\bm\nu}(t);\underline{\bm{\nu}}(t);\phi \to \phi')- g^L(\bm\mu;\phi \to \phi')\right. \right. \\
& \left. \left. \qquad \qquad \ge\inf_{\mathcal{B}_{\phi}} \left(\eta-  g^L(\bm\mu;\phi \to \phi')\right) \right\}\right].
\end{aligned}\end{equation*}
From Assumption~\ref{assm:feas:sep}, we know that $\Delta_{\mathcal{M}^*,q}\le\inf_{\mathcal{B}_{\phi}} \left(\eta-  g^L(\bm\mu;\phi \to \phi')\right)$ which is positive due. Therefore, using Lemma~\ref{g:L:lemma}, one has (III) less than

\begin{equation*}
\begin{aligned}
& \mathbb{E}_{H_a}\left[\sum_{t=1}^{T} \mathbf{1}\left\{\exists \, (L,\phi') \, , \, \phi \in\mathcal{M}^*\backslash\mathbb{F}(\bm \mu, \eta) : I_t=(L,\phi, \phi') \right. \right.\\
&\left. \left. \qquad \qquad \qquad  \left(\sum_{j \in L} \left|r^{j}(\bar{\bm{\nu}}(t);\phi(j))-r^{j}(\bm{\mu};\phi(j))\right| + \sum_{j \in L} \left|r^{j}(\underline{\bm{\nu}}(t);\phi'(j))-r^{j}(\bm{\mu};\phi'(j))\right|\right)  \ge \Delta_{\mathcal{M}^*,q} \right\}\right].
\end{aligned}
\end{equation*}
Using triangle inequality and Assumption~\ref{assm:reward:Lipschitz}, this in turn implies that (III) is less than 

\begin{equation*}
\begin{aligned}
&\mathbb{E}_{H_a}\left[\sum_{t=1}^{T} \mathbf{1}\left\{\exists \, (L,\phi') \, : I_t=(L,\phi)   \, \left(\sum_{j \in L} \sum_{i \in \phi(j)} \left|\bar{\nu}_{i}(t)-\mu_i\right|\right)  \ge \frac{\Delta_{\mathcal{M}^*,q}}{2\, c} \right\}\right]\\
& \qquad + \mathbb{E}_{H_a}\left[\sum_{t=1}^{T} \mathbf{1}\left\{\exists \, (L,\phi') \, : I_t=(L, \phi')   \, \left(\sum_{j \in L} \sum_{i \in \phi'(j)} \left|\underline{\nu}_{i}(t)-\mu_i\right|\right)  \ge \frac{\Delta_{\mathcal{M}^*,q}}{2\, c} \right\}\right].
\end{aligned}
\end{equation*}
Using the property that there exists a value greater than the average and that $|L| \le \kappa$, one has
\begin{equation*}
\begin{aligned}
& \le \mathbb{E}_{H_a}\left[\sum_{t=1}^{T} \mathbf{1}\left\{\exists \, i \in \mathcal{N} \, : I_t=i,  \, \left|\bar{\nu}_{i}(t)-\mu_i\right|  \ge \frac{\Delta_{\mathcal{M}^*,q}}{ 2\, c\, \kappa \, m} \right\}\right]\\
& \qquad + \mathbb{E}_{H_a}\left[\sum_{t=1}^{T} \mathbf{1}\left\{\exists \, i \in \mathcal{N} \, : I_t=i,  \, \left|\underline{\nu}_{i}(t)-\mu_i\right|  \ge \frac{\Delta_{\mathcal{M}^*,q}}{ 2\, c\, \kappa \, m} \right\}\right].
\end{aligned}\end{equation*}
From an identical argument, we have (IV) less than
\begin{equation*}\begin{aligned}
&\mathbb{E}_{H_a}\left[\sum_{t=1}^{T} \mathbf{1}\left\{\exists \, i \in \mathcal{N} \, : I_t=i,  \, \left|\bar{\nu}_{j}(t)-\mu_j\right|  \ge \frac{\left(\eta-\epsilon\right)}{ 2\, c\, \kappa \, m} \right\}\right]\\
& \qquad + \mathbb{E}_{H_a}\left[\sum_{t=1}^{T} \mathbf{1}\left\{\exists \, i \in \mathcal{N} \, : I_t=i,  \, \left|\underline{\nu}_{i}(t)-\mu_i\right|  \ge \frac{\left(\eta-\epsilon\right)}{ 2\, c\, \kappa \, m} \right\}\right]
\end{aligned}\end{equation*}
Using Corollary~\ref{main:coro:suffpull} for each term in the final bounds for (III) and (IV), we get
\begin{equation*}\begin{aligned}
&\mathbb{E}_{H_a}\left[\sum_{t=1}^{T} \mathbf{1}\left\{\delta^{\epsilon}_t=0, \, \exists \,  \phi \in\mathcal{M}^*\backslash\mathbb{F}(\bm \mu, \eta) \, \text{such that } g^L(\bar{\bm\nu}(t);\underline{\bm{\nu}}(t);\phi \to \phi') \ge \eta \, \forall \, (L,\phi')\right\}\right]\\
& \le 8\, N\, \kappa^2 \, m^2\left(\sqrt{2\alpha}+\hat{\Delta}\right)^2
\, c^2\, \sigma^2 \, \left(\Delta^{-2}_{\mathcal{M}^*,q}+(\eta-\epsilon)^{-2}\right)\, \log T +\frac{8\, N}{T^{\hat{\Delta}^2/2-2}}.
\end{aligned}\end{equation*}
Combining, the terms, the proof is completed.
\hfill \Halmos

\noindent\textbf{Proof of Proposition~\ref{prop:null:hyp}.}
Note that when $\delta^{\epsilon}_t=1$, then at time $t$ there exists $\phi \in \mathcal{M}$ such that $g^{L}(\underline{\bm{\nu}}(t);\bar{\bm \nu}(t);  \phi\to  \phi' )\ge \epsilon$ for all $L,\phi'$. Also note that since $H_0$ is true then for all $\phi$ there exists $L , \phi$ such that $g^{L}(\bm \mu; \phi \to \phi')<0$. Hence, by the definition of $R^{H_0}_T$, we have
\begin{equation*}\begin{aligned}
R^{H_0}_T &\le \mathbb{E}\left[\sum_{t=1}^{T} \mathbf{1}\left\{\exists \, \phi, \, (L_,\phi') , \, \phi \in \mathcal{M}^* : \right.\right.\\
& \quad \quad \quad \left. \left. g^L(\bar{\bm\nu}(t);\underline{\bm{\nu}}(t);\phi \to \phi') \ge \epsilon>0> g^L(\bm\mu);\phi \to \phi')\right\}\right]\\
& \le \mathbb{E}\left[\sum_{t=1}^{T} \mathbf{1}\left\{\exists \, j \in \mathcal{K}, \, \phi, \phi' : r^{j}(\underline{\bm{\nu}}(t);\phi(j))- r^{j}(\bar{\bm{\nu}}(t);\phi'(j)) \ge r^{j}(\bm{\mu};\phi(j))- r^{j}(\bm{\mu};\phi'(j))\right\}\right].
\end{aligned}\end{equation*}
Hence, we can argue that 
\begin{equation*}\begin{aligned}
R^{H_0}_T &\le \mathbb{E}\left[\sum_{t=1}^{T} \mathbf{1}\left\{\exists \, i \in \mathcal{N}: \underline{\nu}_i(t) \ge \mu_i \ \text{or }\, \bar{\nu}_i(t) \le \mu_i\right\}\right]       \le \frac{2\,N\, \left(\alpha-1\right) }{\alpha-2}.
\end{aligned}\end{equation*}
Therefore the proof is completed.\hfill \Halmos 

\noindent\textbf{Proof of Theorem~\ref{thm:main:feas}.}
To establish Theorem~\ref{thm:main:feas}, we divide the problem into two parts based on whether the output belongs to the feasible set or not. This further allows us to reduce the problem to an MAB setting.

Hence we shall have
\begin{equation*}\begin{aligned}
R^{H_a}_T &=\mathbb{E}_{H_a}\left[\sum_{t=1}^{T} \mathbf{1}\left\{\delta^{\epsilon}_t=0\right\}\right]\\
&\le \underset{\text{(I)}}{\underbrace{\mathbb{E}_{H_a}\left[\sum_{t=1}^{T} \mathbf{1}\left\{ \delta^{\epsilon}_t(\phi_t)=0, \phi_t\in \mathbb{F}(\bm \mu, \eta)\right\}\right]}}\\
& \quad \quad +\underset{\text{(II)}}{\underbrace{\mathbb{E}_{H_a}\left[\sum_{t=1}^{T} \mathbf{1}\left\{ \delta^{\epsilon}_t(\phi_t)=0, \phi_t \not \in \mathbb{F}(\bm \mu, \eta)\right\}\right]}}.
\end{aligned}\end{equation*}
From Lemma~\ref{approx:lemma:C} we know that for (II), we have
\begin{equation*}\begin{aligned}
&\mathbb{E}\left[\sum_{t=1}^{T} \mathbf{1}\left\{ \delta^{\epsilon}_t(\phi_t)=0, \phi_t \not \in \mathbb{F}(\bm \mu, \eta)\right\}\right] \\
& \le  8\, N\, \kappa^2 \, m^2\left(\sqrt{2\alpha}+\hat{\Delta}\right)^2
\, c^2\, \sigma^2 \, \left(\Delta^{-2}_{\mathcal{M}^*,q}+(\eta-\epsilon)^{-2}\right)\, \log T +\frac{8\, N}{T^{\hat{\Delta}^2/2-2}}+\frac{2\,N\, \alpha}{\alpha-2}.
\end{aligned}\end{equation*}
For (I), we know that there exists $(L,\phi'_t)$ such that $g^L(\underline{\bm{\nu}}(t);\bar{\bm{\nu}}(t);\phi_t\to  \phi'_t)<\epsilon$ since the null hypothesis is declared true. However, $\phi_t \in \mathbb{F}(\bm \mu, \eta)$.\\
Therefore $ g^L(\bm\mu;\phi_t \to \phi'_t)\ge \eta$. Thus,
\begin{equation*}\begin{aligned}
&\mathbb{E}_{H_a}\left[\sum_{t=1}^{T} \mathbf{1}\left\{ \delta^{\epsilon}_t(\phi_t)=0, \phi_t\in \mathbb{F}(\bm \mu, \eta)\right\}\right]\\
&\le \mathbb{E}_{H_a}\left[\sum_{t=1}^{T} \mathbf{1}\left\{ \exists \, (L,\phi'): \, I_t=(L,\phi,\phi'), \, g^L(\underline{\bm{\nu}}(t);\bar{\bm{\nu}}(t);\phi\to \phi')<\epsilon , \, g^L(\bm\mu;\phi \to \phi')\ge \eta\right\}\right]\\
&\le \mathbb{E}_{H_a}\left[\sum_{t=1}^{T} \mathbf{1}\left\{ \exists \, (L,\phi'): \, I_t=(L,\phi,\phi'),  \ g^L(\bm\mu;\phi \to \phi')-g^L(\underline{\bm{\nu}}(t);\bar{\bm{\nu}}(t);\phi\to \phi')\ge \eta-\epsilon\right\}\right].
\end{aligned}\end{equation*}
Replicating the same argument as in Lemma~\ref{approx:lemma:C}, we get (I) to be bounded by 
\begin{equation*}
\begin{aligned}
& \mathbb{E}_{H_a}\left[\sum_{t=1}^{T} \mathbf{1}\left\{\exists \, (L,\phi') \,  : I_t=(L,\phi, \phi') \right. \right.\\
&\left. \left. \qquad \qquad \qquad  \left(\sum_{j \in L} \left|r^{j}(\bar{\bm{\nu}}(t);\phi'(j))-r^{j}(\bm{\mu};\phi'(j))\right| + \sum_{j \in L} \left|r^{j}(\underline{\bm{\nu}}(t);\phi(j))-r^{j}(\bm{\mu};\phi(j))\right|\right)  \ge \eta-\epsilon \right\}\right].
\end{aligned}
\end{equation*}

This, as in Lemma~\ref{approx:lemma:C}, further gives   
\begin{equation*}\begin{aligned}
&\mathbb{E}_{H_a}\left[\sum_{t=1}^{T} \mathbf{1}\left\{ \delta^{\epsilon}_t(\phi_t)=0, \phi_t\in \mathbb{F}(\bm \mu, \eta)\right\}\right]\\
&\le \mathbb{E}_{H_a}\left[\sum_{t=1}^{T} \mathbf{1}\left\{\exists \, (L,\phi') \, : I_t=(L,\phi,\phi'),   \, \left(\sum_{j \in L} \sum_{i \in \phi'(j)} \left|\bar{\nu}_{i}(t)-\mu_i\right|\right)  \ge \frac{(\eta-\epsilon)}{2\, c} \right\}\right]\\
& \qquad + \mathbb{E}_{H_a}\left[\sum_{t=1}^{T} \mathbf{1}\left\{\exists \, (L,\phi') \, : I_t=(L, ,\phi,\phi'), \, \left(\sum_{j \in L} \sum_{i \in \phi(j)} \left|\underline{\nu}_{i}(t)-\mu_i\right|\right)  \ge \frac{(\eta-\epsilon)}{2\, c} \right\}\right]\\
&\le 8\, N\,\left(\sqrt{2\alpha}+\hat{\Delta}\right)^2\, c^2\, \kappa^2 \, m^2 \, \sigma^2 \, \left(\eta-\epsilon\right)^{-2} \, \log T +\frac{8\, N}{T^{\hat{\Delta}^2/2-2}}
\end{aligned}\end{equation*}
where the final step follows from Corollary~\ref{main:coro:suffpull}.
Hence combining the bounds for terms (I) and (II) we get the final bound as 
\begin{equation*}
\begin{aligned}
8\, N\, \kappa^2 \, m^2\left(\sqrt{2\alpha}+\hat{\Delta}\right)^2
\, c^2\, \sigma^2 \, \left(\Delta^{-2}_{\mathcal{M}^*,q}+2\, (\eta-\epsilon)^{-2}\right)\, \log T +\frac{16\, N}{T^{\hat{\Delta}^2/2-2}}+\frac{2\,N\, \alpha}{\alpha-2}       
\end{aligned}
\end{equation*}
and hence we are done.
\hfill \Halmos 

\noindent\textbf{Proof of Proposition~\ref{Prop:feasibility}.}
To address this result we divide the problem based on the decision to accept or reject the null hypothesis which allows us to reduce the problem to an MAB setting in order to apply Corollary~\ref{main:coro:suffpull}. We note that 
\begin{equation*}\begin{aligned}
R_T &= \mathbb{E}_{H_a}\left[\sum_{t=1}^{T}\mathbf{1}\left\{\phi_t \not \in \mathbb{F}(\bm \mu, \eta)\right\}\right]\\
&=\mathbb{E}_{H_a}\left[\sum_{t=1}^{T}\mathbf{1}\left\{\phi_t \not \in \mathbb{F}(\bm \mu, \eta), \, \delta^{\epsilon}_t=1\right\}\right] +\mathbb{E}_{H_a}\left[\sum_{t=1}^{T} \mathbf{1}\left\{\phi_t \not \in \mathbb{F}(\bm \mu, \eta), \, \delta^{\epsilon}_t=0\right\}\right].
\end{aligned}\end{equation*}
Now from Lemma~\ref{approx:lemma:C}, for the second term, we have the bound as

\[8\, N\, \kappa^2 \, m^2\left(\sqrt{2\alpha}+\hat{\Delta}\right)^2
\, c^2\, \sigma^2 \, \left(\Delta^{-2}_{\mathcal{M}^*,q}+(\eta-\epsilon)^{-2}\right)\, \log T +\frac{8\, N}{T^{\hat{\Delta}^2/2-2}}+\frac{2\, N \, (\alpha-1)}{\alpha-2}.\]
For the first term, we note that if $\phi_t \not \in \mathbb{F}(\bm \mu, \eta)$, which implies that there exists $(L,\phi')$ such that $g^{L}(\bm\mu;\phi_t\to \phi')<\eta-\Delta_{\mathcal{M}^*,q}$ while $g^L(\underline{\bm{\nu}}(t);\bar{\bm{\nu}}(t);\phi_t\to \phi')\ge \epsilon$.\\ Now as $\epsilon$ is chosen carefully, this implies $g^L(\underline{\bm{\nu}}(t);\bar{\bm{\nu}}(t);\phi_t\to \phi') \ge g^{L}(\bm\mu;\phi_t \to \phi')$ Hence
\begin{equation*}\begin{aligned}
&\mathbb{E}_{H_a}\left[\sum_{t=1}^{T}\mathbf{1}\left\{\phi_t \not \in \mathbb{F}(\bm \mu, \eta), \, \delta^{\epsilon}_t=1\right\}\right]\\
&\le \mathbb{E}_{H_a}\left[\sum_{t=1}^{T}\mathbf{1}\left\{\exists \, (L,\phi'): \, g^L(\underline{\bm{\nu}}(t);\bar{\bm{\nu}}(t);\phi_t\to  \phi')\ge g^{L}(\bm\mu;\phi_t\to \phi')\right\}\right]\\
&\le\mathbb{E}_{H_a}\left[\sum_{t=1}^{T}\mathbf{1}\left\{\exists \, i \in \mathcal{N}: \, \underline{\nu}_i(t)\ge \mu_i \ \text{ or 
}\mu_i \ge \bar{\nu}_i(t)\right\}\right]\\
&\le \frac{2\,N\, \left(\alpha-1\right)}{\alpha-2}.
\end{aligned}\end{equation*}
Combining the bounds the result follows. \hfill \Halmos

\end{document}